\documentclass{article}

\usepackage[preprint, nonatbib]{neurips_2025}

\usepackage[utf8]{inputenc} 
\usepackage[T1]{fontenc}    
\usepackage{hyperref}       
\usepackage{url}            
\usepackage{booktabs}       
\usepackage{amsfonts}       
\usepackage{nicefrac}       
\usepackage{microtype}      
\usepackage{xcolor}         
\usepackage{graphicx}
\usepackage{amsmath} 
\usepackage{algorithm}
\usepackage{algpseudocode}
\usepackage{array}
\usepackage{multirow}
\usepackage{subcaption}
\usepackage{comment}

\newcommand{\solution}{{LLMPred}}

\title{Univariate to Multivariate: LLMs as Zero-Shot Predictors for Time-Series Forecasting}

\author{%
  Chamara Madarasingha\thanks{corresponding author--chamara.kattadige@rmit.edu.au, chamarakcm@gmail.com} \\
  School of Computing Technologies\\
  RMIT University, Australia \\
  \And
  Nasrin Sohrabi \\
  School of Information Technologies\\
  Deakin University, Australia \\
  \AND
  Zahir Tari \\
  School of Computing Technologies\\
  RMIT University, Australia \\
}

\begin{document}

\maketitle

\begin{abstract}
Time-series prediction or forecasting is critical across many real-world dynamic systems, and recent studies have proposed using Large Language Models (LLMs) for this task due to their strong generalization capabilities and ability to perform well without extensive pre-training.
However, their effectiveness in handling complex, noisy, and multivariate time-series data remains underexplored. To address this, we propose \solution{} which enhances LLM-based time-series prediction by converting time-series sequences into text and feeding them to LLMs for zero-shot prediction along with two main data pre-processing techniques. First, we apply time-series sequence decomposition to facilitate accurate prediction on complex and noisy univariate sequences. Second, we extend this univariate prediction capability to multivariate data using a lightweight prompt-processing strategy. Extensive experiments with smaller LLMs such as Llama 2 7B, Llama 3.2 3B, GPT-4o-mini, and DeepSeek 7B demonstrate that \solution{} achieves competitive or superior performance compared to state-of-the-art baselines. Additionally, a thorough ablation study highlights the importance of the key components proposed in \solution{}. \url{https://github.com/llmpred/llmpred-code})
\end{abstract}

\section{Introduction}

Time-series forecasting plays a critical role in various applications such as financial and stock market prediction~\cite{fischer2018deep}, weather forecasting~\cite{rasp2020weatherbench}, and transportation and traffic flow prediction~\cite{lv2014traffic} etc. Unlike text, audio, or video sequences—which typically exhibit consistent value ranges and contextual patterns—time-series data often involve a wide range of feature values and multiple channels, making them more challenging to model~\cite{gruver2023large}. Although recent advances in deep learning (DL) have significantly improved time-series prediction and forecasting performance, these models typically require large volumes of diverse data and expert knowledge~\cite{sivaroopan2024netdiffus}. However, collecting sufficient and high-quality real-time data remains a significant challenge, and inadequate data can not only degrade model performance but also hinder scalability and generalizability\cite{madarasingha2022videotrain++}.

To overcome these challenges, Large Lanugage Modles (LLM) have been identified as a promising solution due to several reasons~\cite{gruver2023large, liu2024autotimes, jin2023time}.
First, LLMs exhibit strong generalization capabilities, as they are trained on massive and diverse datasets spanning text, audio, and numerical sequences. This broad exposure enables them to effectively handle applications involving numerical data, such as numerical reasoning and mathematical problem-solving~\cite{zhu2025language, vacareanu2024words, feng2024numerical, rahman2025large}. Second, since LLMs are pre-trained on large amount of data, they can be adapted to new tasks with minimal or no fine-tuning~\cite{gruver2023large}, significantly reducing the time and effort required for training while maintaining high predictive accuracy. Third, their pre-training on extensive datasets makes LLMs naturally well-suited for scenarios with limited available data, offering a robust foundation even in data-scarce settings.


While LLMs hold promise for time-series forecasting, their performance on complex, noisy data remains underexplored. Noise—such as fluctuations, outliers, or level shifts—can obscure trends and seasonal patterns, making accurate forecasting difficult. Unlike natural language, which follows more predictable structures, time-series data often exhibit abrupt changes, posing challenges for LLMs. Additionally, most existing LLM-based methods focus on univariate generation~\cite{gruver2023large, liu2024autotimes, jin2023time}, limiting their ability to model cross-variable dependencies crucial for multivariate forecasting.

To this end, in this paper, we further explore the feasibility of LLMs for time-series generation by introducing a novel framework, \solution{}, with two main contributions. First, we propose a decomposition strategy that separates time-series sequences into low-frequency trends and high-frequency fluctuations to enhance univariate prediction. Leveraging the inherent zero-shot capabilities of LLMs on text data, we convert time-series sequences into text prompts for prediction and apply post-processing steps to further refine the generated outputs, improving both fidelity and temporal consistency. Second, we extend univariate generation to multivariate generation by introducing a lightweight prompt-processing strategy that enables LLMs to model multivariate distributions and cross-channel dependencies more effectively.

In addition, we conduct a critical analysis to highlight the inherent challenges LLMs face in multivariate prediction. As \solution{} integrates multiple components into a unified framework, we perform an extensive ablation study to evaluate the contribution and necessity of each module. Our experiments focus on relatively smaller LLMs (e.g.,Llama 2 7B, Llama 3.2 3B) showcasing that \solution{} not only adapts well to resource-constrained models but also broadens the practical usability of LLM-based forecasting systems. Experimental results demonstrate that \solution{} outperforms state-of-the-art (SOTA) benchmarks by achieving a 26.8\% reduction in MSE for univariate generation and closely matches benchmark performance while relying solely on the zero-shot prediction capabilities of LLMs. \solution{} improves forecasting performance by 17.4\% when moving from univariate to multivariate tasks, and its ablation study confirms the importance of all components used in model.

\section{Related work}
\paragraph{LLMs for time-series prediction}

LLMs have recently been explored for time-series forecasting by representing time-series data as sequences, similar to text~\cite{gruver2023large}. After tokenization, LLMs process numerical sequences using transformer architectures, which are well-suited for capturing sequential patterns~\cite{chen2024sdformer, wu2020adversarial, sommers2024survey}. Existing methods generally follow two approaches. The first approach converts numerical sequences into text and leverages pre-trained LLMs in a zero-shot manner. 
For instance, PromptCast~\cite{xue2023promptcast} first converts input and target numerical sequences into prompt-style sentences and runs forecasting as a text generation task. In~\cite{gruver2023large}, Gruver et al.~further show that effective tokenization, by adding whitespace between the digits and then converting to text, improves the zero-shot prediction performance of LLMs. 
In the second approach, LLMs are fine-tuned using embedded representations of the numerical values. In~\cite{chang2023llm4ts}, the authors first align the segmented time-series patches with the input space of LLMs, followed by fine-tuning the model parameters. Building on this idea, Jin et al.~\cite{jin2023time} present a patch reprogramming method to better match the LLM’s original embedding distribution. Liu et al.~\cite{liu2024autotimes} extend these techniques to autoregressive forecasting by segmenting time-series data and predicting future segments based on already predicted segments.

\paragraph{Univariate and Multivariate time-series prediction}
Univariate and multivariate time series refer to sequences with a single feature and multiple features (or channels), respectively. To the best of our knowledge, existing LLM-based time-series predictors focus mainly on univariate sequence prediction~\cite{gruver2023large, liu2024autotimes, jin2023time}. This is because LLMs are inherently trained on text sequences that align more closely with univariate structures. Multivariate forecasting has been widely explored in prior work, particularly using transformer-based models~\cite{han2024softs, wu2021autoformer, zhou2022fedformer, zhou2021informer, yi2023fouriergnn, zheng2024multivariate}. Since transformers are the foundation of LLMs, we focus mainly on transformer-based approaches for multivariate prediction. Models such as Autoformer~\cite{wu2021autoformer}, Informer~\cite{zhou2021informer}, and FEDformer~\cite{zhou2022fedformer}, which we further describe in~\ref{append:subsec-benchmarks}, support multivariate inputs by capturing inter-variable dependencies.

\paragraph{Decomposing time-series for efficient prediction}
Decomposing time series into multiple components based on time- or frequency-related features (e.g., trend and seasonal patterns or frequency components) has been explored in the literature~\cite{wu2021autoformer, zhou2022fedformer, zhou2021informer, yi2023fouriergnn, wang2022learning}. Such decomposition reduces the complexity of raw time-series sequences and facilitates the learning of underlying patterns, leading to improved forecasting performance. For instance, Autoformer~\cite{wu2021autoformer} introduces a decomposition architecture that integrates trend-seasonal decomposition into each model layer, enabling progressive modeling of complex temporal dynamics. Similarly, FEDformer~\cite{zhou2022fedformer} employs a dedicated component in its model to decompose the seasonal and trend patterns of time-series data.

\textbf{In \solution}, we draw key insights from prior literature to address two major limitations in current LLM-based time-series forecasting. First, we recognize that models like GPT-4o-mini struggle to capture complex temporal patterns in raw sequences; thus, we introduce an effective time-series decomposition strategy to separate low- and high-frequency components, simplifying the learning task. Second, while existing LLM-based approaches predominantly focus on univariate prediction, we demonstrate that this univariate prediction can be extended to multivariate prediction through carefully designed prompt engineering techniques that align with multivariate structures.

\section{ \solution{} Methodology}\label{sec:method}
In multivariate time-series forecasting, a dataset consists of multiple variables (a.k.a. channels or features) at each time step. Given historical values $X \in \mathbb{R}^{L\times C}$, where $L$ is the number of samples and $C$ is th number of channels, the goal of \solution{} is to predict future values $Y \in \mathbb{R}^{H \times C}$, where $H > 0$ and is the prediction length. In developing \solution{}, we first focus on forecasting univariate data and then extend the approach to multivariate time-series generation. A given feature/channel is indicated by $X^c$. Next we present the details of individual components of \solution{}.

\subsection{Decomposition into frequency components}

\begin{figure}[h]
    \centering
    \includegraphics[width=1.0\linewidth]{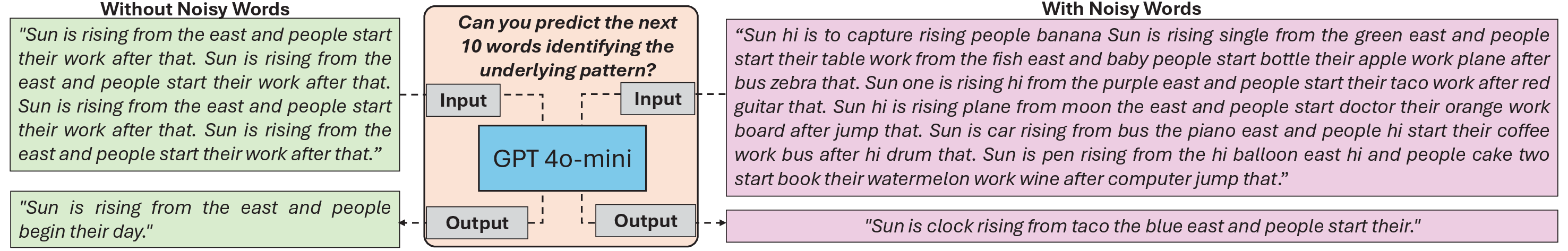}
    \caption{A simple example: The left prompt shows a repetitive structure, allowing GPT-4o-mini to predict the next 10 words by recognizing the pattern. When noise is added (right), the model struggles to capture the key signal, and the output also becomes noisy.}
    \label{fig:sample_text_analogy}
\end{figure}

LLMs are inherently designed to capture contextual meaning from text-based inputs. When applied to numerical data that has been transformed into a textual distribution, a significant challenge arises in identifying meaningful contextual patterns. Unlike natural text sequences, numerical time-series data exhibit unique characteristics, particularly the presence of noise, which is often represented as high-frequency components. These high-frequency fluctuations introduce randomness into the time-series distribution, making it difficult to extract underlying seasonal patterns or trends. A similar phenomenon can be observed in text prediction, where the addition of noisy words disrupts the model’s ability to predict subsequent words accurately, as illustrated in Fig.~\ref{fig:sample_text_analogy}.

\begin{figure}[h]
    \centering
    \includegraphics[width=\linewidth]{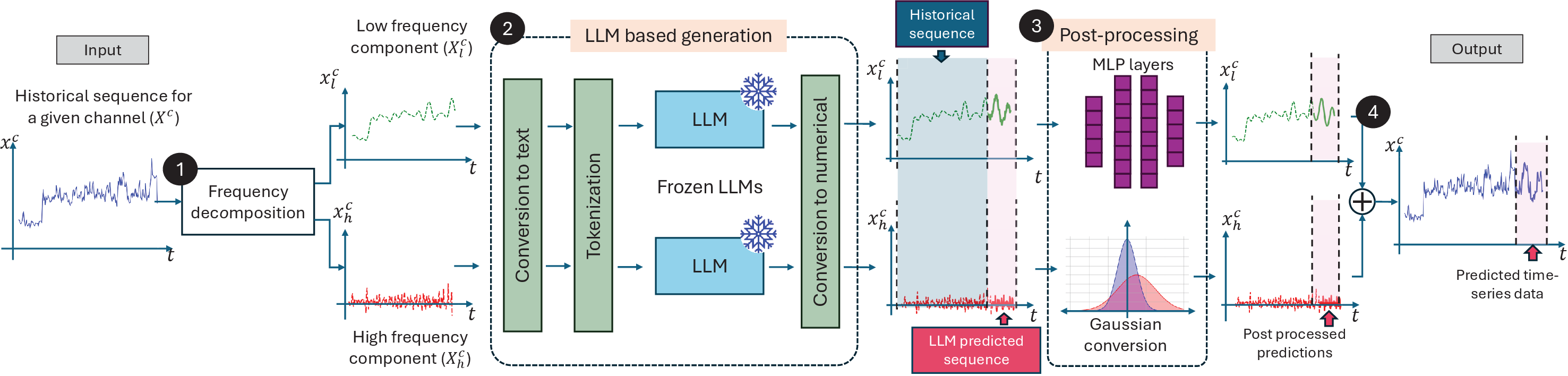}
    \caption{Overview of \solution{} Univariate Generation: 1)~A given time-series sequence ($X^c$) is first decomposed into high ($X^c_h$)- and low ($X^c_l$) -frequency components. 2)~Each component is then processed separately using pre-trained LLMs after converting the numeric to text and applying tokenization. The LLM outputs are converted back into numerical sequences. 3) Post-processing steps are applied: predicted $X^c_l$ is refined using a simple MLP network to improve pattern fidelity, while a Gaussian transformation is applied to predicted $X^c_h$ to align its value range with historical data. 4)~Final step combines the two components together to obtain the complete prediction.}
    \label{fig:individual-gen}
\end{figure}

Motivated by this observation, our first contribution in \solution{} is the decomposition of a given time-series distribution into two distinct components: low-frequency ($X^c_l$) and high-frequency ($X^c_h$) parts (see Fig.~\ref{fig:individual-gen}). $X^c_l$ generally preserve the global structure of the time series while remaining robust to high-frequency noise artifacts. Given that the fundamental architecture of LLMs—the Transformer—is inherently more effective at capturing low-frequency variations \cite{piao2024fredformer}, this decomposition facilitates more accurate predictions by enabling the model to focus on the stable, structured components of the data. In the meantime, $X^c_h$ is now separated from long term trends, where the LLM can understand noisy fluctuations without affected by low frequency trends.

\paragraph{Similarity metric based frequency decomposition}
We follow a systematic approach for frequency decomposition based on metrics: {Cosine similarity} and {Mean Squared Error (MSE)}, as detailed in Algorithm~\ref{alg:cutoff} (Appendix~\ref{append-subsec:find-cut-off}). A selected range of cut-off frequencies ($F$) from 2.5 to 15.0\,Hz is applied to each feature. This range captures the dominant low-frequency components, as observed in the power-spectral diagrams of the datasets used in \solution{}, enabling effective low- and high-frequency decomposition. For a given time-series sequence, $X^c$, we derive $X_l^c$ and $X_h^c$ components  
using the following approach. 

For a given $f\in F$, we first compute the MSE between $X^c$ and $X_l^{c,f}$ (denoted $m_{\text{mse}}$) to assess alignment in amplitude trends. Here, $X_l^{c,f}$ is the low frequency component at $f$. Similarly, we measure Cosine similarity (denoted $m_{\text{cos}}$) between $X^c$ and $X_h^{c,f}$ (higher frequency component at $f$) , a magnitude-invariant metric, to evaluate how well the high-frequency component captures the finer fluctuations. 
\begin{equation}\label{eq:cut-freq}
    m = \alpha * m_{mse} + (1-\alpha) * 1/m_{cos}
\end{equation}
Following Eq.~\ref{eq:cut-freq}, we compute the weighted sum of $m_{\text{mse}}$ and $1/m_{\text{cos}}$ to balance the contributions of $X_l^{c,f}$ and $X_h^{c,f}$ and the $f$ value which minimizes combined metric, $m$, is selected as cut-off frequency denoted by $f_{cut}$. The $X_l^{c,f}$ and $X_h^{c,f}$ at $f_{cut}$ is selected as optimum high and low frequency components, $X_l^{c}$ and $X_h^{c}$ respectively. A higher value of $m_{\text{cos}}$ indicates better similarity between $X^c$ and $X_h^c$. However, since we aim to minimize $m$ in Eq.~\ref{eq:cut-freq}, we instead use $1/m_{\text{cos}}$.The weighting factor $\alpha$ can be set arbitrarily.

\subsection{LLM based generation}\label{subsec:llm_gen}
 
\paragraph{Conversion to text:}
This is the first step in LLM-based data generation, where we convert numerical sequences to textual format before tokenization. To avoid large value ranges and to preserve the context length of the LLM, we apply max normalization (i.e., taking the absolute maximum value), which scales the sequence to the range $[-1, 1]$ and the values are with maximum of two decimal points. From a text-based perspective, the negative sign $(-)$ requires a dedicated token. Therefore, we apply a linear transformation $0.5x + c$ to shift values into a positive range. For example, setting $c=0.5$ maps $[-1,1]$ to $[0,1]$, while setting $c=1.5$ shifts it to $[1,2]$ respectively. To further guide the LLM, we prepend the historical sequence with an instructional prompt, as detailed in Appendix~\ref{append-subsec:intro prompt}.


\paragraph{Tokenization:}
Efficient tokenization is crucial for converting numerical inputs into an LLM-understandable textual format. At the same time, tokenization should provide a meaningful representation of numerical sequences while preserving their essential characteristics. We follow the approach proposed by~\cite{gruver2023large} for Llama and {Deepseek} LLMs. In \solution{}, we retain up to a maximum of two decimal places.  
LLMs such as Llama treat text values as separate characters by default (e.g., ``3.75''$\rightarrow$[``3'', ``.'', ``7'', ``5'']), whereas GPT models like GPT-4o-mini use Byte Pair Encoding (BPE), which tokenizes ``3.75'' as [``3'', ``.'', ``75'']. This approach allows the LLM to better preserve numerical values even after textual transformation.

\paragraph{Predictions through LLM:}  
In \solution{}, we use pre-trained LLMs in a zero-shot setting to predict numerical values without fine-tuning. The intuition behind this approach is to evaluate the ability of LLMs to predict purely numerical sequences, despite being primarily trained on large-scale textual data. On one hand, this also allows us to assess the direct applicability of LLMs for numerical sequence modeling without additional fine-tuning. On the other hand, we avoid the compute power and the time required for model training making \solution{} adaptable quickly for a given time-series dataset. Tokenized numerical sequences are provided as prompts, along with an instruction specifying the number of values to predict (Appendix~\ref{append-subsec:intro prompt}).

\paragraph{Conversion to numerical:}
This step converts textual sequences into numerical values before further optimization through post-processing steps (see $\S~\ref{subsec:post-process}$). During this process, we perform additional validation on the textual output to detect any incomplete numerical generations, which can occur in two main ways.  In univariate generation, a partial value may be produced (e.g., ``3.75''$\rightarrow$``3.''). In multivariate generation, incomplete feature values can occur when multiple features for a given time instance are arranged in a single record, which we further described in \S~\ref{subsec:multi gen}, (e.g., ``0.34, 1.54, 2.67''$\rightarrow$``0.34, 1.54, ``...''). To ensure data integrity, we exclude such cases during this validation step. Moreover, we apply the reverse process of the linear transformation we applied during \textit{Conversion to text} process and the resulting feature value range is now $[-1,1]$.

\subsection{Post-processing}~\label{subsec:post-process}
Since we do not fine-tune the LLM and rely on pre-trained models, some complex patterns may not be captured properly by the LLM. To mitigate this, we apply distinct post-processing methods for the predicted $X_l^c$ and $X_h^c$ to refine the outputs.

\paragraph{Simple MLP model for predicted $X_l^c$:}
We train a simple Multi-Layer Perceptron (MLP) to further refine $X_l^c$ predictions for three reasons. First, \solution{} targets short-term prediction, where an MLP suffices~\cite{lazcano2024back}. Second, the low-frequency component is less complex than high-frequency patterns, making it easier for the MLP to model~\cite{han2024softs}. Third, while sequential models like Recurrent Neural Networks (RNNs) capture time-series patterns better, they require more computational resources. The MLP, in contrast, allows efficient training and fast inference~\cite{yi2023frequencydomain}.

\paragraph{Gaussian transformation for predicted $X_h^c$:}
During the design of the high-frequency post-processing mechanism, we made three key observations. First, since historical $X_h^c$ is centralized around a constant value (0), the predicted output is close to 0. Second, applying an MLP or similar neural network is challenging due to the noisy and unpredictable nature of the data. Third, $X_h^c$ often follows a Gaussian distribution (see Appendix~\ref{subsubsec:gaussian plotsww}). Based on these, we apply a Gaussian transformation to the LLM high-frequency output. We compute the mean ($\mu$) and standard deviation ($\sigma$) for both historical ($\mu_h$, $\sigma_h$) and predicted ($\mu_p$, $\sigma_p$) sequences, and use the transformation $x_{\text{new}} = \left( \frac{x - \mu_p}{\sigma_p} \right) \cdot \sigma_h + \mu_h$ to adjust the predicted $X_h^c$ to match its historical $X^c_f$ distribution.

\subsection{Extended prediction of multivariate time-series generation}\label{subsec:multi gen}

\begin{figure}[h]
    \centering
    \includegraphics[width=\linewidth]{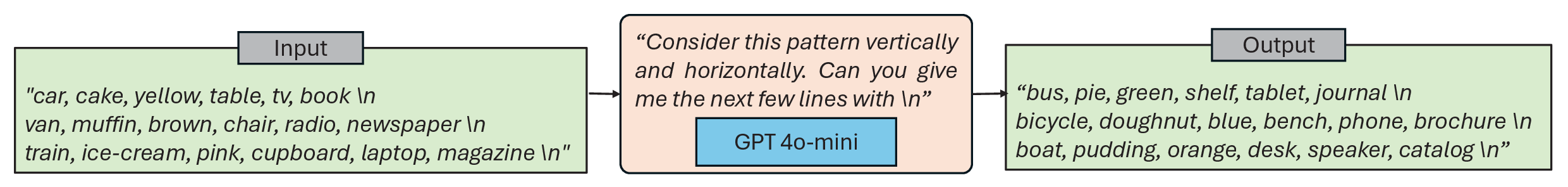}
    \caption{LLM demonstrates contextual understanding by recognizing categorical patterns across structured data and continue those patterns logically. For example, the column 1 in the input prompt contains the vehicles and the column 1 of the output prompt also contains vehicles.}
    \label{fig:llm-uni-to-multi}
\end{figure}

We extend univariate generation to the multivariate setting by leveraging the LLM’s ability to distinguish structural patterns in text. As illustrated in Fig.~\ref{fig:llm-uni-to-multi}, LLMs are capable of identifying the contextual meaning embedded in such structural cues. We adopt a similar methodology by transferring features into distinct ranges using a simple linear transformation function, as described in $\S$~\ref{subsec:llm_gen}. Specifically, assigning values such as $c = 0.5$, $1.5$, and $2.5$ transforms the feature distributions into the ranges $[0\text{--}1]$, $[1\text{--}2]$, and $[2\text{--}3]$, respectively. Inspired by~\cite{requeima2024llm}, we arrange the multiple features of a given time instance as a single row in the text prompt, separated by commas (e.g., ``0.32, 1.14, 2.45, 3.55, 4.56''), and append ``\textbackslash n'' to indicate the end of the time instance (see Appendix~\ref{append-subsec:sample-output-prediction-len} for sample value representations).

\section{Evaluation}\label{sec:eval}\label{subsec:evaluation}
\subsection{Evaluation setup}

In \solution{}, we assess the zero-shot forecasting capabilities of four widely-used LLM models: Llama~3.2-3B (denoted as Llama-3B)~\cite{meta2024llama3.2}, Llama~2-7B (Llama-7B)~\cite{touvron2023llama2}, Deepseek-basic-7B (Deepseek-7B)~\cite{deepseek2024llm}, and OpenAI GPT-4o-mini-8B (GPT-4o-mini)~\cite{openai2024gpt4omini}. These models were chosen for their relatively smaller parameter sizes compared to larger models like Llama~2-70B or GPT-4-Turbo. This because they require less computational power and are easier to integrate into systems at scale. By evaluating \solution{} with these smaller LLMs, we demonstrate its adaptability in low-resource environments. Further details of evaluation setup is in Appendix~\ref{append-sec:eval}.

We use a widely adopted time-series prediction dataset from~\cite{zhou2021informer}. We evaluate the dataset under two settings: \textit{i})~{univariate}, where the model is provided with individual features for prediction, and \textit{ii})~{multivariate}, where multiple features are simultaneously input into \solution{} for forecasting.We compare the performance of \solution{} against several transformer-based baselines, including FedFormer~\cite{zhou2022fedformer}, Autoformer~\cite{wu2021autoformer}, and Informer~\cite{zhou2021informer}. For LLM-based approaches, we consider {AutoTimes}~\cite{liu2024autotimes}, which makes predictions on time-series data by converting numerical values directly into embeddings, and {LLMTime}~\cite{gruver2023large}, which first transforms numerical sequences into text—a representation that closely aligns with the input mechanism of \solution{}. To ensure a fair comparison, we apply the  max-normalization to all benchmark models for the datasets independently during evaluation.

\subsection{Univariate analysis}\label{subsec:uni-analysis}

\begin{figure}[h!]
    \centering
    \begin{subfigure}[b]{0.40\textwidth}
        \centering
        \includegraphics[width=\textwidth]{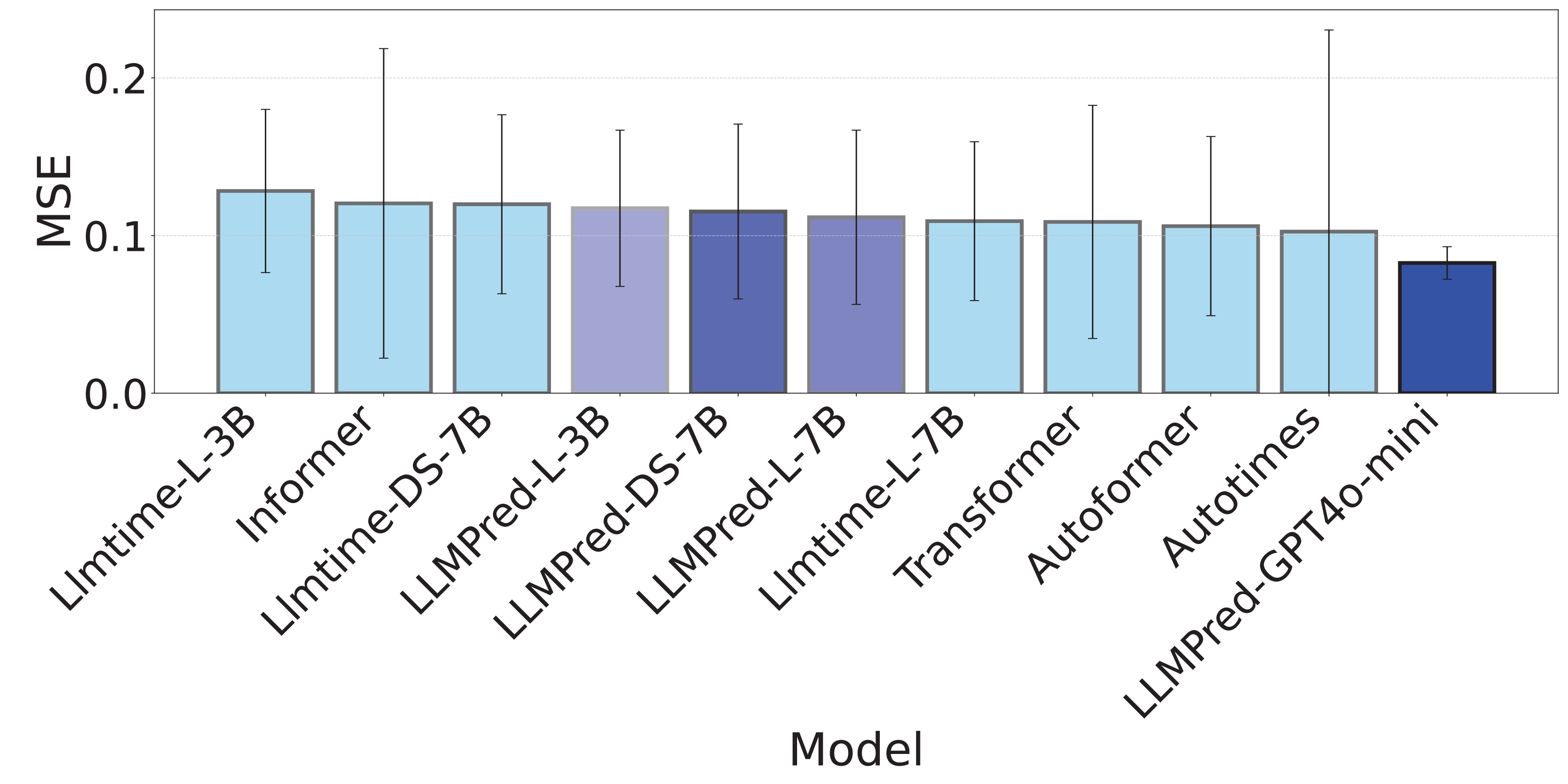}
        \caption{Univariate prediction}
        \label{fig:mse-uni}
    \end{subfigure}
    \begin{subfigure}[b]{0.265\textwidth}
        \centering
        \includegraphics[width=\textwidth]{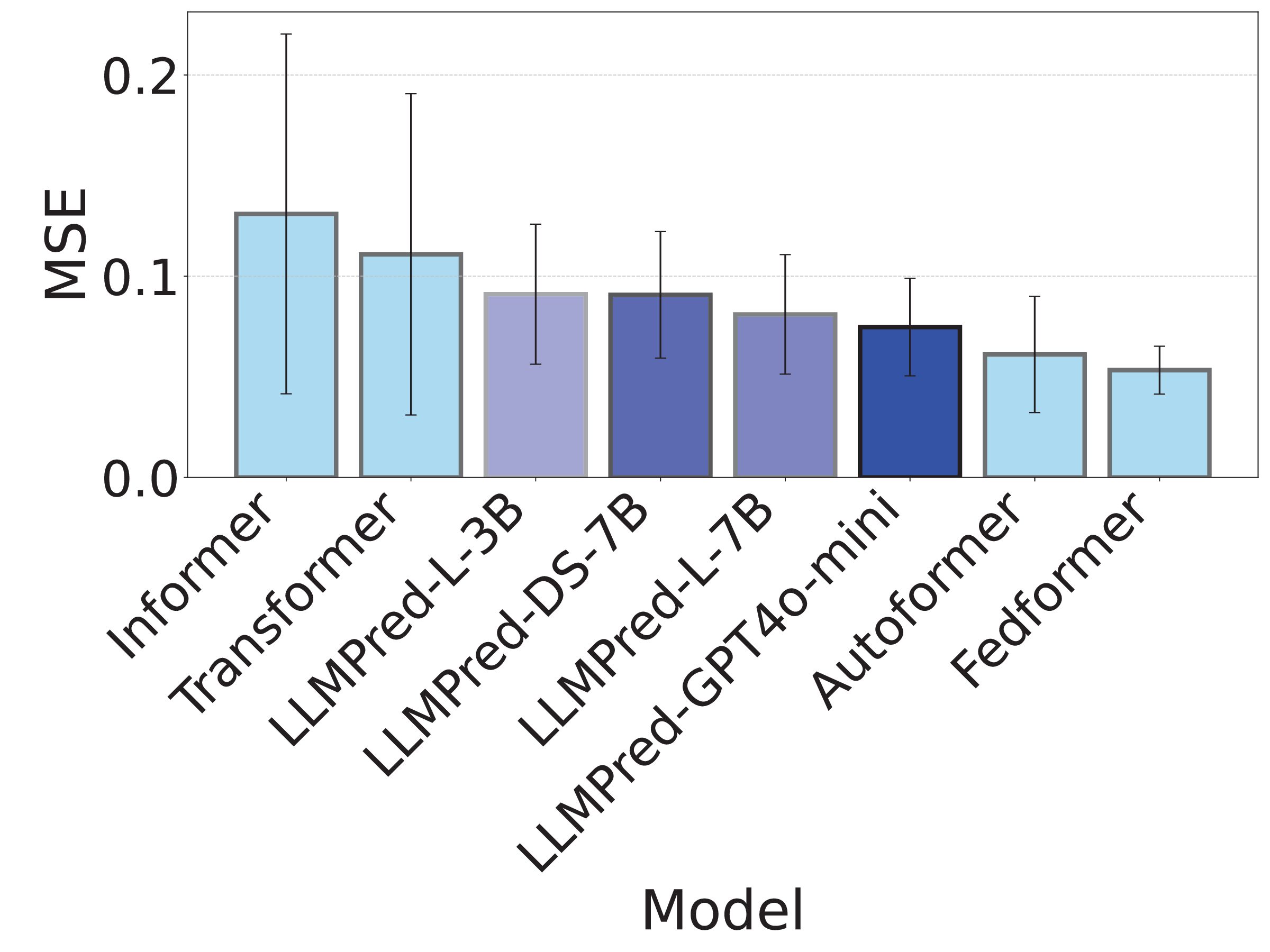}
        \caption{Multivariate prediction}
        \label{fig:mse-multi}
    \end{subfigure}
    \begin{subfigure}[b]{0.265\textwidth}
        \centering
        \includegraphics[width=\textwidth]{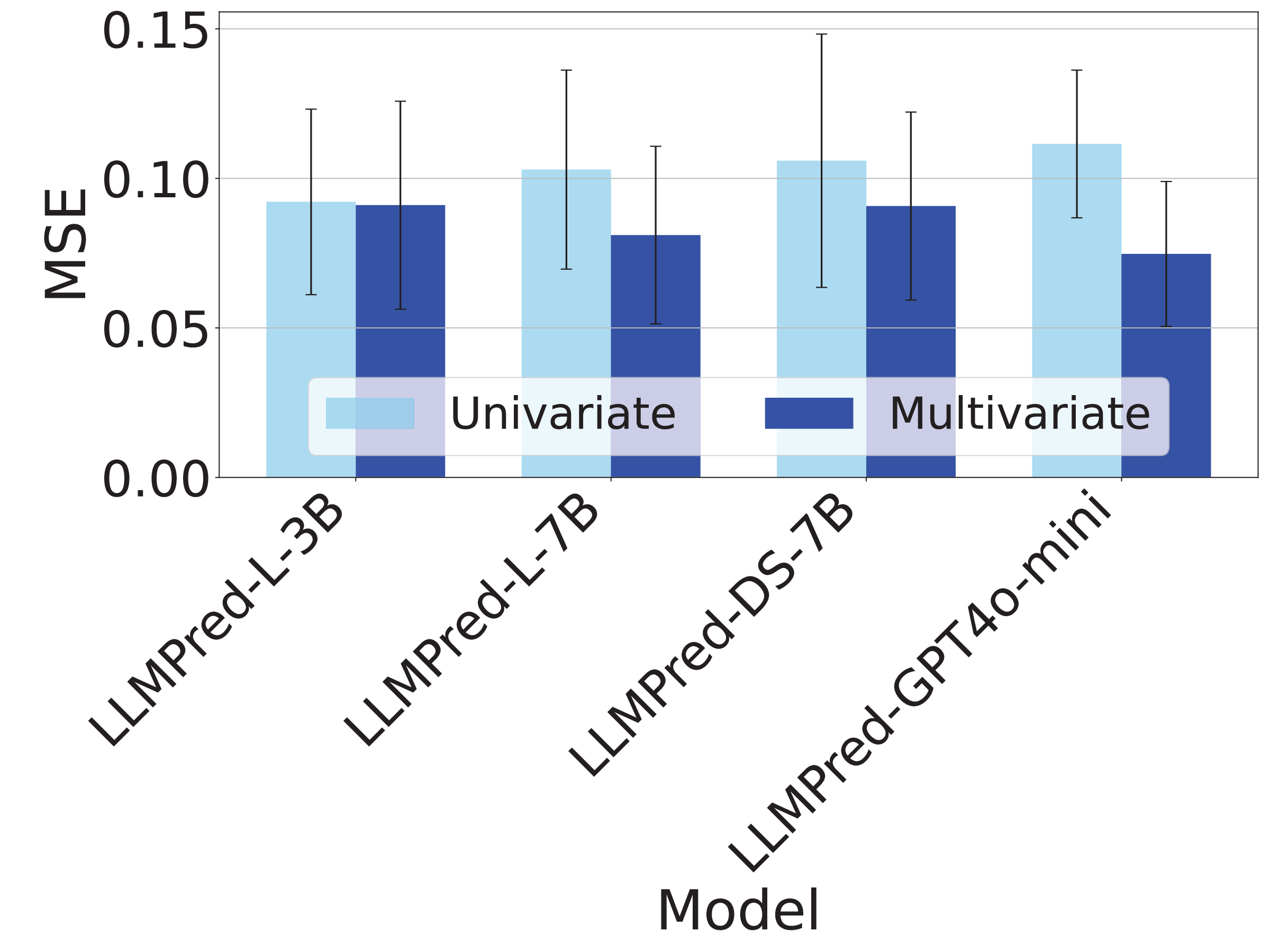}
        \caption{Univariate vs Multivariate}
        \label{fig:mse-uni vs multi}
    \end{subfigure}
    \caption{Fig.~\ref{fig:mse-uni} and Fig.~\ref{fig:mse-multi} show MSE values measured across different datasets for univariate and multivariate generation respectively taking both prediction lengths 48 and 96 samples together. Fig.~\ref{fig:mse-uni vs multi} compares the MSE value between univariate and multivariate generation of \solution{}. Here L: Llama and DS: Deepseek.}
    \label{fig:var-diff-alpha}
\end{figure}

Fig.~\ref{fig:mse-uni} compares \solution{} with baseline models in terms of MSE. \solution{}-GPT4o-mini consistently outperforms all other methods, achieving the lowest average MSE of $0.08 \pm 0.01$, representing a 26.8\% reduction compared to the benchmarks, with minimal variance across datasets. Other variants of \solution{}, such as those based on Llama and Deepseek, show relatively lower performance compared to {LLMTime} (Llama-7B), {Autoformer}, and {AutoTimes}. Nevertheless, all \solution{} models outperform the majority of {LLMTime} variants, which similarly treat numerical sequences as text. Additional comparisons, including MAE and varying prediction lengths, are provided in Appendix~\ref{append-subsec:uni analysis}.

\subsection{Multivariate analysis}\label{subsec:multi}

Fig.~\ref{fig:mse-multi} shows the comparison with the benchmarks in multivariate forecasting. Here, we compare only with methods that support multivariate prediction. In this setting, Autoformer and FEDformer achieve the lowest MSE values, while \solution{}-GPT4o-mini records the third-best MSE distribution with (0.07$\pm$0.02). Autoformer~\cite{wu2021autoformer} and FEDformer~\cite{zhou2022fedformer} perform better in this setting due to their built-in decomposition mechanisms and frequency-based modeling, fine-tuned for multivariate scenarios. However,  \solution{}, with its simpler and zero-shot prediction, achieves comparable results, showing only a 0.007–0.010 MSE difference relative to these benchmarks. On the other hand, \solution{} variants demonstrate more consistent performance across datasets, with a lower average standard deviation of 0.030, compared to the benchmark models, which exhibit a higher average standard deviation of approximately 0.041. 
This shows improved robustness of \solution{} across  datasets.

Despite the relatively higher MSE compared to Autoformer and FEDformer, \solution{} achieves the advantage of multivariate forecasting which is the ability to capture inter-feature dependencies. As shown in Fig.~\ref{fig:mse-uni vs multi}, the multivariate prediction of \solution{} outperforms its univariate version by 17.4\% on average, indicating that \solution{} benefits from multivariate inputs. Further evaluations are in Appendix~\ref{append:subsec:multi}.

\subsection{Critical evaluation of multivariate prediction}\label{subsec:critical-eval}
\subsubsection{Impact of context length}\label{subsubsec:impact context len}
\begin{figure}[h!]
    \centering
    \begin{subfigure}[b]{0.49\textwidth}
        \centering
        \includegraphics[width=\textwidth]{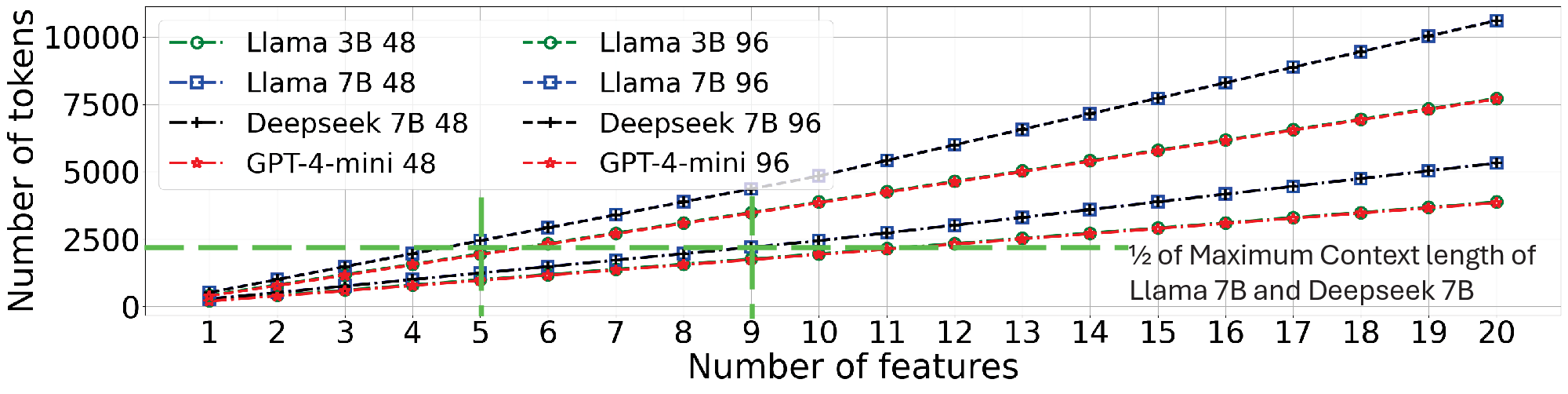}
        \caption{Context length variation}
        \label{fig:context-length}
    \end{subfigure}
     \begin{subfigure}[b]{0.49\textwidth}
        \centering
        \includegraphics[width=\textwidth]{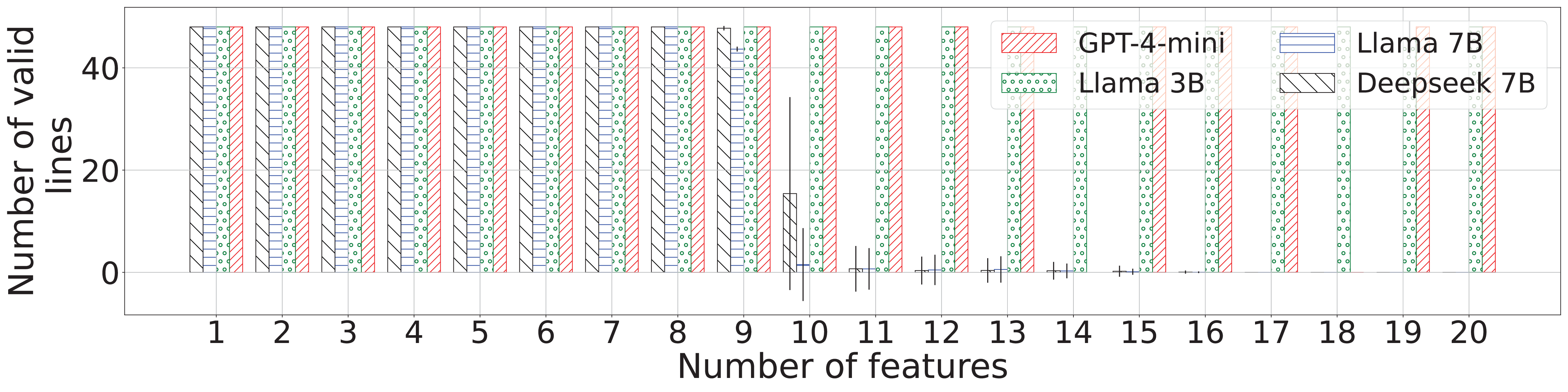}
        \caption{Number of rows with exact number of features}
        \label{fig:valid-lines}
    \end{subfigure}
    \begin{subfigure}[b]{0.49\textwidth}
        \centering
        \includegraphics[width=\textwidth]{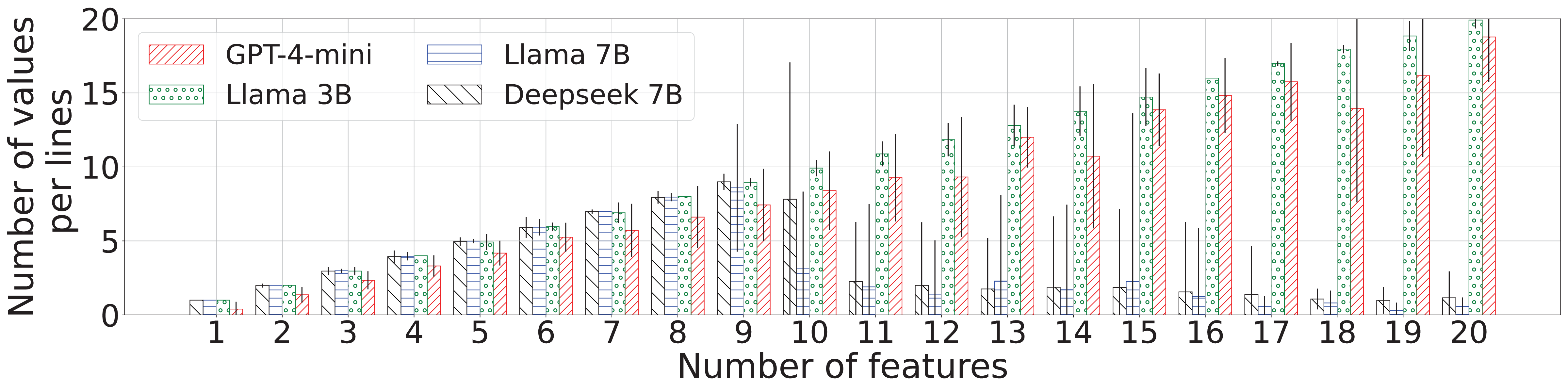}
        \caption{Number of features per line}
        \label{fig:valid-features-per-line}
    \end{subfigure}
    \begin{subfigure}[b]{0.24\textwidth}
        \centering
        \includegraphics[width=\textwidth]{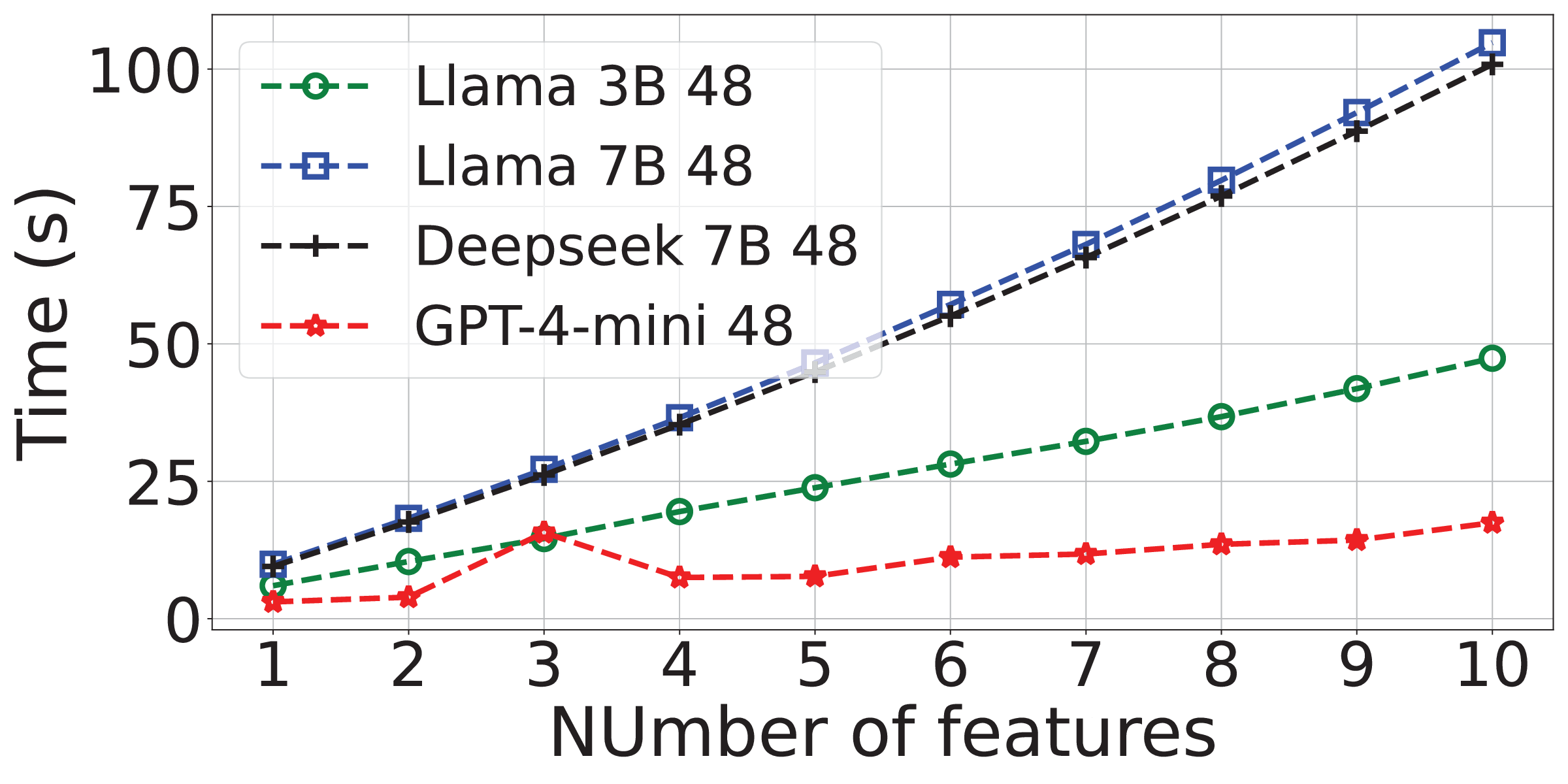}
        \caption{Time for forecast - 48}
        \label{fig:time-48}
    \end{subfigure}
    \begin{subfigure}[b]{0.24\textwidth}
        \centering
        \includegraphics[width=\textwidth]{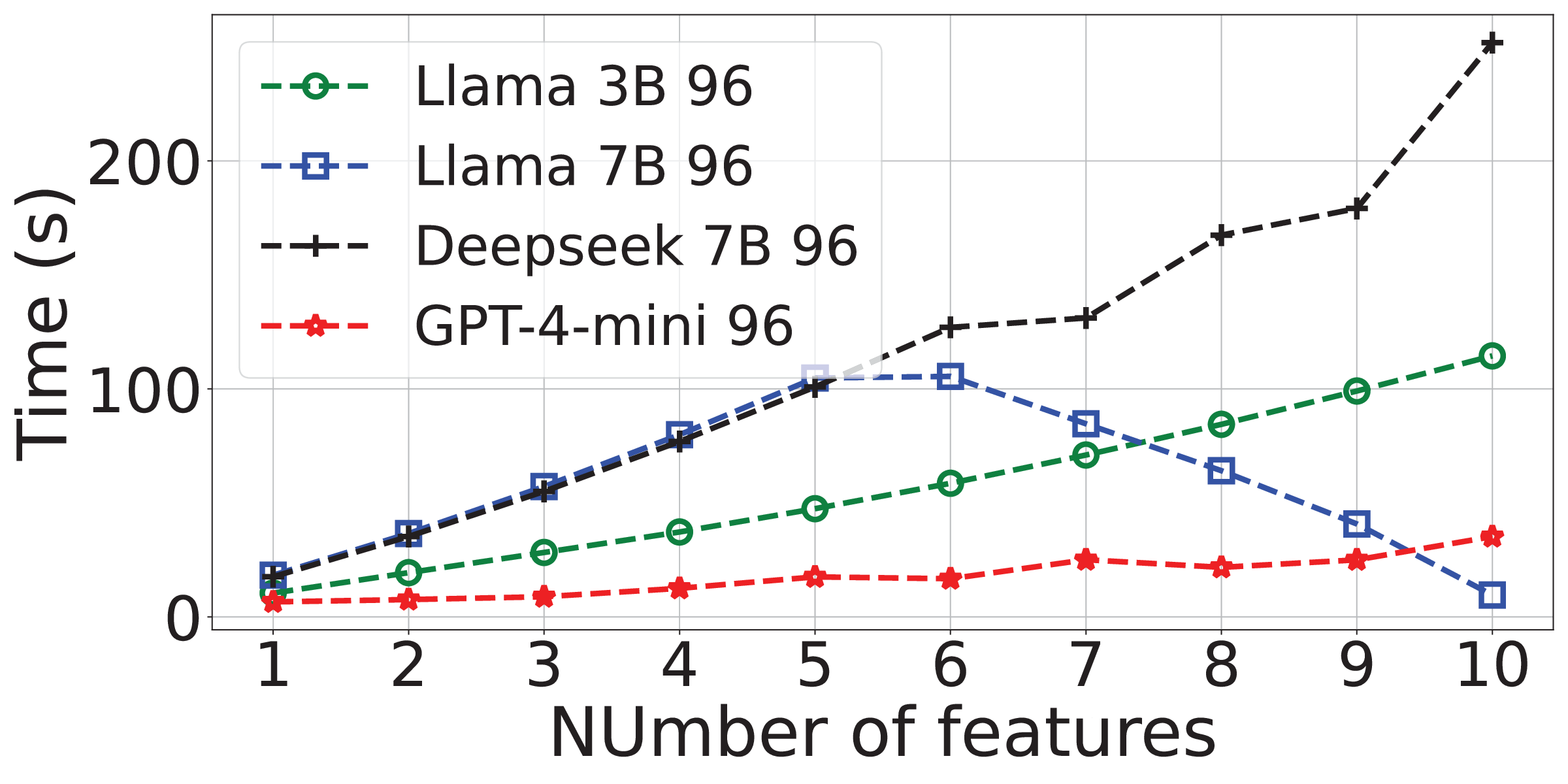}
        \caption{Time for forecast - 96}
        \label{fig:time-96}
    \end{subfigure}
    \caption{Impact of number of features in multivariate prediction measuring multiple parameters.}
    \label{fig:num-feat-multi}
\end{figure}

 Fig.~\ref{fig:context-length} shows the number of tokens in the input prompt as the number of features increases. Since each feature corresponds to a sequence of values, and each value has a fixed number of digits in text form, all curves in Fig.~\ref{fig:context-length} increase linearly. For both prediction lengths, 48 and 96, the model pairs GPT-4o-mini \& Llama 3B and Llama 7B \& Deepseek 7B show almost the same context lengths. The maximum context length for Llama 7B and Deepseek 7B is 4096 tokens, with half of the length (2048 tokens) typically allocated for output as the LLM prediction length is equal to the historical sequence length. We observe that in multivariate generation for 96 and 48 prediction lengths, the models reach the maximum allowable context length for output prediction with only 5 and 9 features, respectively.  If the number of tokens exceeds this limit (combining the input and output tokens), the model often produces repetitive or noisy outputs. Therefore, adding more features constrains the LLMs' prediction abilities, which we discuss further in the next section ($\S $\ref{subsubsec:why not longer prediction length}).

\subsubsection{Why not longer prediction lengths?}\label{subsubsec:why not longer prediction length}
Fig.~\ref{fig:valid-lines} and Fig.~\ref{fig:valid-features-per-line} show the number of valid lines (i.e., valid line: one row/line of output prediction with expected number of feature values separated by comma) and the number of features per line, respectively, for the prediction length scenario of 48. As discussed previously (see $\S$~\ref{subsubsec:impact context len}), at 9 features, both Llama 7B and Deepseek 7B reach their maximum context length, leading to fluctuations in the number of valid lines and the number of features per line in the output prompts. Therefore, to ensure a fair comparison between all models and considering both prediction lengths, we select 6 features for multivariate generation. 
Additionally, we limit the prediction length to a maximum of 96 to avoid reaching the maximum context length even when using a smaller number of features.

\subsubsection{Time taken for LLM prediction}\label{subsubsec:time-taken for prediction}
Fig.~\ref{fig:time-48} and Fig.~\ref{fig:time-96} illustrate the prediction time during multivariate generation for 48 and 96 prediction lengths, respectively, averaged over four sample predictions for each number of features. In both cases, GPT-4o-mini achieves the lowest prediction time. For all models except Llama 7B in the 96-prediction scenario, the prediction time increases linearly with the number of features, reaching up to 100 seconds and over 200 seconds when forecasting with 10 features. 
In the 96-prediction case, Llama 7B's prediction time drops after 5 features, but accuracy drops as context length limit exceeds.

\subsection{Ablation study}

\subsubsection{Cut-off frequency ($f_{cut}$) vs $\alpha$}\label{subsubsec:cut-off vs alpha}

\begin{figure}[h!]
    \centering
      \begin{subfigure}[b]{0.24\textwidth}
        \centering
        \includegraphics[width=\textwidth]{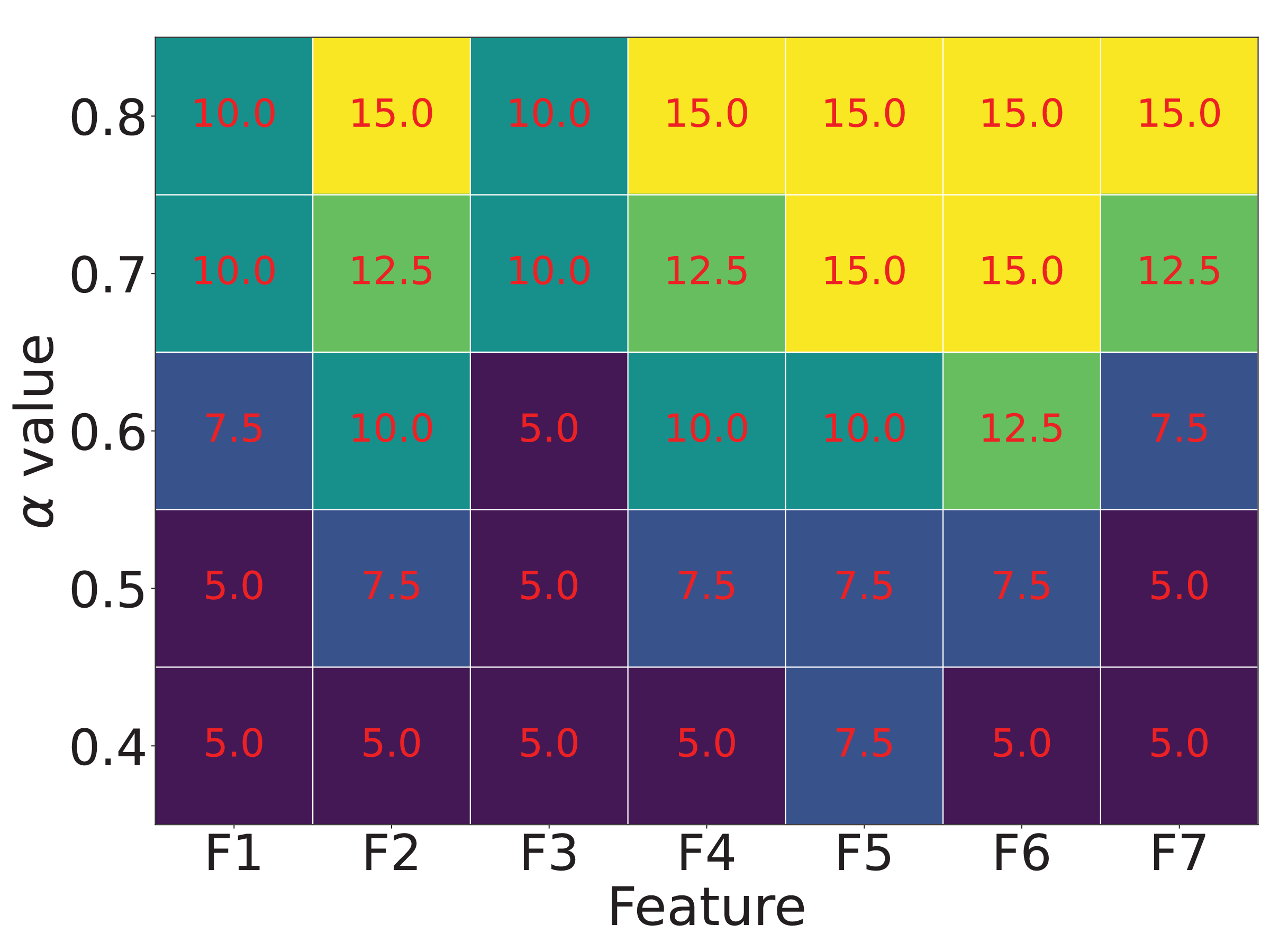}
        \caption{ETTh1}
        \label{fig:diff-alpha-0.5-etth1}
    \end{subfigure}
     \begin{subfigure}[b]{0.24\textwidth}
        \centering
        \includegraphics[width=\textwidth]{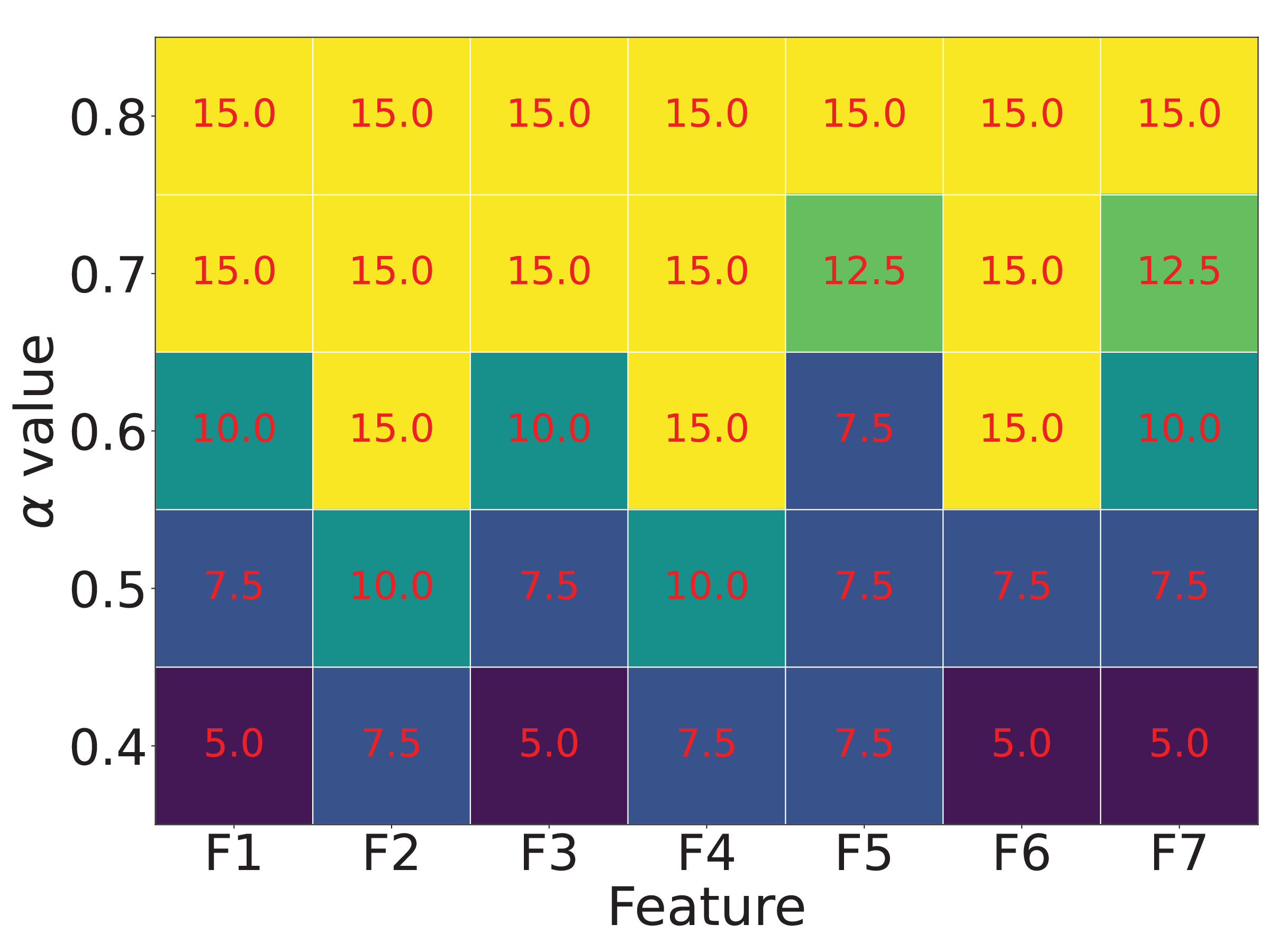}
        \caption{ETTm1}
        \label{fig:diff-alpha-0.5-ettm1}
    \end{subfigure}
    \begin{subfigure}[b]{0.24\textwidth}
        \centering
        \includegraphics[width=\textwidth]{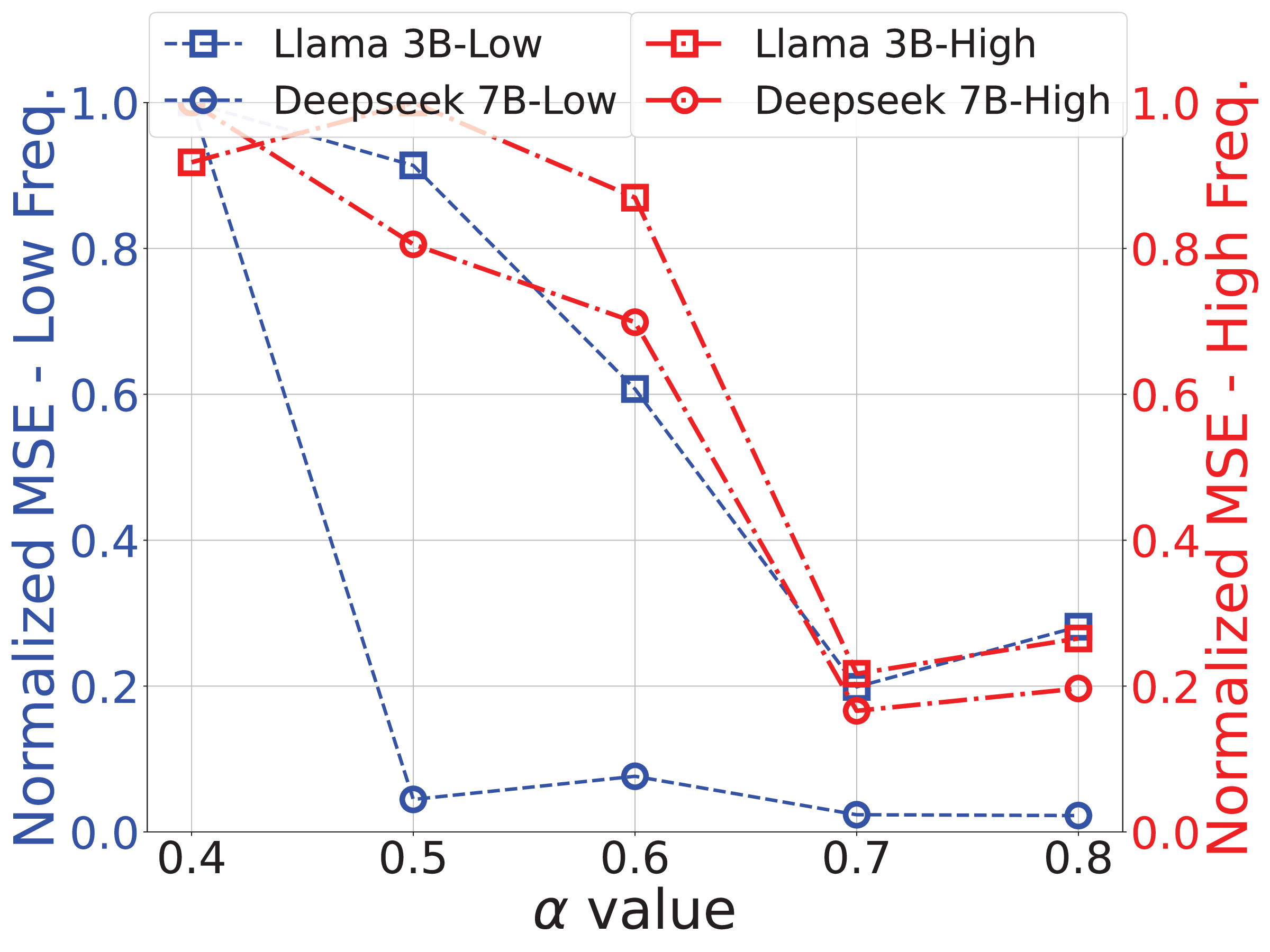}
        \caption{ETTh1}
        \label{fig:mse-alpha-etth1}
    \end{subfigure}
    \begin{subfigure}[b]{0.24\textwidth}
        \centering
        \includegraphics[width=\textwidth]{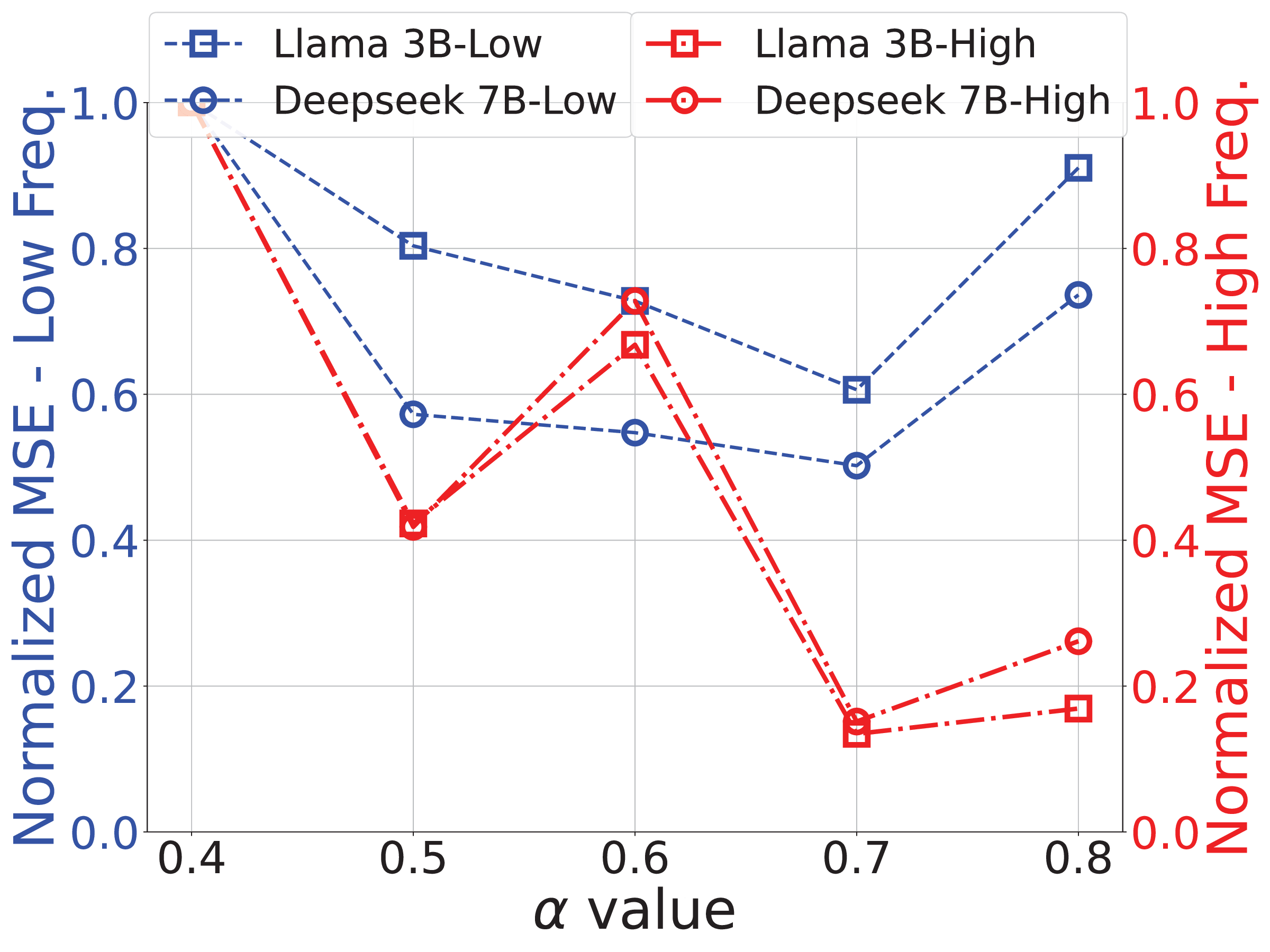}
        \caption{Weather}
        \label{fig:mse-alpha-weather}
    \end{subfigure} 
    \caption{Impact of $\alpha$ on LLM generation. Fig.~\ref{fig:diff-alpha-0.5-etth1} and Fig.~\ref{fig:diff-alpha-0.5-ettm1} - Variation of $f_{cut}$ for different features when changing $\alpha$. Higher the $f_{cut}$, brighter the cell. Fig.~\ref{fig:mse-alpha-etth1} and Fig.~\ref{fig:mse-alpha-weather} -
    Normalized MSE  measure for both low and high frequency components are given for selected two datasets and two models.}
    \label{fig:impact diff alpha}
\end{figure}

\paragraph{$f_{cut}$ distribution for different $\alpha$: } We examine the distribution of cut-off frequencies across different $\alpha$ values for various features in the datasets. 
As shown in Fig.~\ref{fig:diff-alpha-0.5-etth1} and Fig.~\ref{fig:diff-alpha-0.5-ettm1}, the resulting cut-off frequencies span from 5.0 to 15.0 Hz. We observe that lower $\alpha$ values yield lower cut-off frequencies, while higher $\alpha$ values result in higher cut-offs. This trend occurs because a lower $\alpha$ assigns more weight to the high-frequency component $X^c_h$, which can preserve more of the original signal's characteristics when $f_{cut}$ is kept low. Conversely, a higher $\alpha$ emphasizes the low-frequency component $X^c_l$, requiring a higher $f_{cut}$ to retain the dominant characteristics of $X^c$ within $X^c_l$. We present $f_{cut}$ vs $\alpha$ distributions for other datasets in Appendix~\ref{append-subsubsec:cut-off-dist}.

\paragraph{Impact of $\alpha$ on LLM generation:}
We evaluate the influence of the weighting factor $\alpha$ on LLM predictions using two models—Llama-3B and Deepseek-7B—across two benchmark time-series datasets: ETTh1 and Weather. To investigate the frequency-domain effects, we derive input features for various cut-off frequencies corresponding to different $\alpha$ values and feed them independently to the LLMs, thereby eliminating inter-feature interference observed in multivariate inputs.
To assess prediction fidelity, we compute normalized MSE values separately for $X_l^c$ and $X_h^c$, comparing the predicted and ground truth sequences prior to any post-processing.
The results in Fig.~\ref{fig:mse-alpha-etth1} and Fig.~\ref{fig:mse-alpha-weather} demonstrate that as the $\alpha$ value increases from 0.4 to 0.8, the MSE for both low- and high-frequency components decreases progressively. Notably, $\alpha = 0.7$ consistently produces lower MSE values for both components across all experimental settings. Therefore, we select $\alpha = 0.7$ for subsequent data generation tasks. At this setting, the corresponding cut-off frequency $f_{cut}$ lies within the range of 10.0–15.0 Hz (see Appendix~\ref{append-subsubsec:cut-off-dist}). According to Eq.~\ref{eq:cut-freq}, this choice prioritizes the retention of low-frequency trends (via the $m_{\text{mse}}$ term) while reducing the influence of high-frequency noise.

\begin{table}[htbp]
\centering
\caption{Impact of frequency decomposition comparing high, low, combined frequency components, and having original values (i.e., no decomposition) as the input to the LLM. Values are max normalized across datasets. \textbf{Highlighted} values are by comparing combined and original scenarios. }
\renewcommand{\arraystretch}{1.2}
\setlength{\tabcolsep}{4pt}
\scriptsize
\label{table:impact freq decom}
\begin{tabular}{|l|c|c|c|c|c|c|c|c|c|c|c|c|}
\hline
\multirow{2}{*}{\textbf{Dataset}} & \multicolumn{4}{c|}{\textbf{Llama-3B}} & \multicolumn{4}{c|}{\textbf{Llama2-7B}} & \multicolumn{4}{c|}{\textbf{Deepseek-7B}} \\
\cline{2-13}
 & Low & High & Comb. & Orig. & Low & High & Comb. & Orig. & Low & High & Comb. & Orig. \\
\hline
ETTh1 & 0.822 & 0.064 & \textbf{0.877} & 1.000 & 0.919 & 0.070 & 0.980 & \textbf{0.891} & 0.813 & 0.073 & \textbf{0.847} & 0.883 \\
\hline
ETTh2 & 0.815 & 0.141 & 0.942 & \textbf{0.743} & 0.799 & 0.183 & 1.000 & \textbf{0.730 }& 0.796 & 0.173 & 0.929 & \textbf{0.756} \\
\hline
ETTm1 & 0.622 & 0.018 & \textbf{0.616} & 1.000 & 0.541 & 0.022 & \textbf{0.557} & 0.926 & 0.488 & 0.022 & \textbf{0.493} & 0.890 \\
\hline
ETTm2 & 0.688 & 0.074 &\textbf{ 0.733} & 1.000 & 0.650 & 0.090 & \textbf{0.744} & 0.922 & 0.683 & 0.091 & \textbf{0.777 }& 0.910 \\
\hline
Electricity & 0.858 & 0.088 & \textbf{0.956} & 1.000 & 0.883 & 0.079 & 0.944 & \textbf{0.752} & 0.654 & 0.088 & \textbf{0.736}& 0.736 \\
\hline
Weather & 0.866 & 0.006 & \textbf{0.831} & 0.995 & 0.896 & 0.006 & \textbf{0.898} & 0.900 & 0.815 & 0.006 & \textbf{0.767 }& 1.000 \\

\hline
\end{tabular}
\end{table}

\subsubsection{Impact of frequency division on LLM generation}
We evaluate the effect of frequency decomposition on LLM outputs by measuring the normalized MSE between model predictions with and without frequency decomposition. Three models are considered in this analysis: Llama-3B, Llama-7B, and Deepseek-7B. The MSE is calculated between the predicted and ground-truth values for different sequence types: $X_l^c$, $X_h^c$, the reconstructed time series combining $X_l^c$ and $X_h^c$, and the prediction using the original time series ($X^c$) without decomposition. Table~\ref{table:impact freq decom} reports the normalized MSE results across all models for each dataset, enabling clear comparison between the models as well. Comparing the combined scenario with the original, we observe that the combined version achieves lower MSE in most cases, leading to an average reduction of 14.4\% across all datasets. Model-wise, Llama-3B exhibits the highest average MSE reduction, approximately 19.3\%. We further observe that separating $X_l^c$ and $X_h^c$ yields lower MSE values than their combined version. This occurs because superimposing $X_h^c$ onto $X_l^c$ introduces additional variation, causing the combined signal to deviate more from the ground truth compared to the smoother low-frequency input alone.

\subsubsection{Impact of Post-Processing Steps.}\label{subsubsec:impact post-processing}
We assess the effectiveness of \solution{}'s post-processing for low frequency component ($X_l^c$ and $c\in \{1,2,3\}$) by comparing the MSE between ground truth and predicted values before and after post-processing. Table~\ref{table:post-process low mse} reports MSE averaged across multiple test samples, along with their standard deviations (sd). Overall, the post-processing using MLP yields a 22.7\% reduction in MSE, with the GPT-4o-mini model achieving the highest improvement of 32.9\% across all six datasets. Notably, post-processing improves the stability of LLM predictions, as reflected by an 82.0\% reduction in the sd of the MSE values. Further evaluations are in Appendix~\ref{append-subsubsec:impact-post-process}.

Since the high-frequency component ($X_h^c$) follows a Gaussian distribution, we evaluate the effectiveness of the Gaussian transformation using the Kolmogorov–Smirnov (KS) test to measure the similarity between the predicted and ground truth distributions. Table~\ref{table:post process high ks} reports the KS statistics before and after applying the post-processing step. A lower KS statistic indicates higher similarity. We compare both the original LLM outputs and the post-processed outputs with the corresponding ground truth distributions. On average, the post-processing step reduces the KS statistic by 22.7\%, indicating improved alignment with the target distribution. Among all models, Llama-3B demonstrates the most significant improvement with a 36.2\% reduction in the KS statistic. In terms of stability, the Gaussian transformation also reduces the sd of the KS values across datasets by 15.8\% to 78.0\%, further supporting its robustness. 

\begin{table}[htbp]
\centering
\caption{Comparison of model performance before and after the proposed method fir \textbf{Low} frequency component. Values are rounded to two decimal positions due to the space limitation.}
\renewcommand{\arraystretch}{1.2}
\setlength{\tabcolsep}{5pt}
\label{table:post-process low mse}
\scriptsize
\begin{tabular}{|l|c|c|c|c|c|c|c|c|}
\hline
\multirow{2}{*}{\textbf{Dataset}} & \multicolumn{2}{c|}{\textbf{Llama-3B}} & \multicolumn{2}{c|}{\textbf{Llama2-7B}} & \multicolumn{2}{c|}{\textbf{Deepseek-7B}} & \multicolumn{2}{c|}{\textbf{GPT-4o-mini}} \\
\cline{2-9}
 & Before & After & Before & After & Before & After & Before & After \\
\hline
ETTh1 & 0.21 (0.28) & 0.12 (0.06) & 0.28 (0.39) & 0.15 (0.08) & 0.26 (1.00) & 0.14 (0.06) & 0.19 (0.28) & 0.17 (0.09) \\
\hline
ETTh2 & 0.10 (0.20) & 0.06 (0.02) & 0.11 (0.23) & 0.07 (0.01) & 0.10 (0.28) & 0.07 (0.02) & 0.18 (0.43) & 0.14 (0.08) \\
\hline
ETTm1 & 0.22 (0.36) & 0.15 (0.10) & 1.00 (1.00) & 0.19 (0.10) & 0.20 (0.29) & 0.17 (0.09) & 0.24 (0.37) & 0.15 (0.08) \\
\hline
ETTm2 & 0.07 (0.12) & 0.06 (0.02) & 0.08 (0.12) & 0.09 (0.02) & 0.07 (0.12) & 0.07 (0.01) & 0.17 (0.30) & 0.18 (0.06) \\
\hline
Electricity & 0.07 (0.12) & 0.04 (0.01) & 0.09 (0.18) & 0.05 (0.02) & 0.06 (0.09) & 0.04 (0.01) & 0.21 (0.48) & 0.05 (0.01) \\
\hline
Weather & 0.06 (0.13) & 0.09 (0.07) & 0.10 (0.30) & 0.09 (0.04) & 0.05 (0.12) & 0.11 (0.07) & 0.34 (0.68) & 0.18 (0.13) \\
\hline
\end{tabular}
\end{table}
\begin{table}[htbp]
\centering
\caption{KS value for high frequency component. Mean and standard deviation (sd) values are shown.}
\renewcommand{\arraystretch}{1.2}
\setlength{\tabcolsep}{5pt}
\label{table:post process high ks}
\scriptsize
\begin{tabular}{|l|c|c|c|c|c|c|c|c|}
\hline
\multirow{2}{*}{\textbf{Dataset}} & \multicolumn{2}{c|}{\textbf{Llama-3B}} & \multicolumn{2}{c|}{\textbf{Llama2-7B}} & \multicolumn{2}{c|}{\textbf{Deepseek-7B}} & \multicolumn{2}{c|}{\textbf{GPT-4o-mini}} \\
\cline{2-9}
 & Before & After & Before & After & Before & After & Before & After \\
\hline
ETTh1 & 0.07 (0.05) & 0.07 (0.03) & 0.09 (0.03) & 0.05 (0.01) & 0.07 (0.03) & 0.05 (0.02) & 0.08 (0.01) & 0.05 (0.01) \\
\hline
ETTh2 & 0.16 (0.01) & 0.09 (0.02) & 0.07 (0.01) & 0.07 (0.01) & 0.12 (0.04) & 0.07 (0.02) & 0.07 (0.02) & 0.07 (0.01) \\
\hline
ETTm1 & 0.18 (0.06) & 0.08 (0.02) & 0.08 (0.02) & 0.07 (0.02) & 0.07 (0.03) & 0.09 (0.03) & 0.07 (0.01) & 0.06 (0.01) \\
\hline
ETTm2 & 0.18 (0.06) & 0.09 (0.05) & 0.11 (0.07) & 0.08 (0.03) & 0.13 (0.01) & 0.08 (0.05) & 0.09 (0.04) & 0.07 (0.03) \\
\hline
Electricity & 0.18 (0.01) & 0.08 (0.02) & 0.09 (0.05) & 0.10 (0.05) & 0.09 (0.02) & 0.08 (0.02) & 0.08 (0.02) & 0.08 (0.05) \\
\hline
Weather & 0.14 (0.07) & 0.11 (0.03) & 0.10 (0.05) & 0.08 (0.02) & 0.09 (0.03) & 0.08 (0.01) & 0.10 (0.03) & 0.07 (0.01) \\
\hline
\end{tabular}
\end{table}

\section{Limitations}\label{sec:limitations}
\solution{} currently limits numerical values to two decimal places to improve tokenization efficiency and reduce token count. While effective, this may impair the model’s ability to capture subtle variations in sensitive domains. As an alternative, future work will explore numerically-aware tokenization that preserves full values in single tokens, maintaining both precision and semantic coherence~\cite{gruver2023large}.
Second, \solution{} uses a predefined frequency range (2.5–15.0 Hz) for decomposing time-series data into low- and high-frequency components. While this range effectively captures trends and fluctuations in a wide range of datasets, it may limit adaptability to time-series exhibiting slower or faster periodicities outside this range. Future versions of \solution{} could incorporate a data-driven mechanism for dynamically selecting the cut-off frequency based on the time-series' spectral properties.
Finally, our post-processing for high-frequency components assumes that time-series sequences follow a Gaussian distribution due to noisy nature, which holds in many but not all cases. Certain time-series may exhibit skewed or heavy-tailed distributions due to irregular or abrupt fluctuations. To improve generalization, we aim to enhance our approach by detecting and adapting to the underlying statistical properties of the high-frequency data, moving beyond a fixed Gaussian assumption. We further discuss limitations \solution{} through examples in Appendix~\ref{append:subsubsec:limitations through samples}.

\section{Conclusion}

In this paper, we presented \solution{}, a large language model (LLM)-based approach for multivariate time series forecasting by converting numerical sequences directly into textual format and feeding them into LLMs. To enhance forecasting performance, we proposed decomposing univariate time series into low-frequency and high-frequency components, enabling the model to capture both trend patterns and rapid fluctuations, and thereby helping LLMs understand the underlying temporal structures effectively. Furthermore, we extended univariate time series decomposition to multivariate time series generation by providing a critical analysis of multivariate time series modeling. Through extensive experiments comparing \solution{} with state-of-the-art methods, we demonstrate that it achieves competitive—or even superior—performance, despite relying on relatively small LLMs. We also conducted a comprehensive ablation study, highlighting the importance of each component and enhancement applied in \solution{}. In our immediate future work, we aim to integrate dynamic frequency decomposition, distribution-aware high-frequency modeling, and improved robustness in the generation of time series data.

\bibliographystyle{plain}  
\bibliography{20_ref} 
\appendix
\section{Further details on \solution{} methodology}

Here, we provide further details of the \solution{} methodology, including key algorithms, instructional prompts used to guide LLM for forecasting sequences and post-processing steps. 

\subsection{Finding cut-off frequency}\label{append-subsec:find-cut-off}
Algorithm~\ref{alg:cutoff} describes the procedure for determining the cut-off frequency ($f_{cut}$) for each feature, given a feature distribution $X^c$ and a specified $\alpha$ value. The algorithm is applied iteratively across a range of $\alpha$ values---[0.4, 0.5, 0.6, 0.7, 0.8]---for each feature. Based on experimental evaluation, we select $\alpha = 0.7$ as the optimal value. We adopt the Butterworth filter with a 100~Hz sampling rate due to its flat passband, which minimizes signal distortion while effectively separating low- and high-frequency components. Its smooth roll-off enables controlled noise suppression, preserving essential signal characteristics~\cite{hu2024intelligent,bruna2016selection,basu2020comparative}.

\begin{algorithm}
    \caption{Finding cut-off frequency}\label{alg:cutoff}
    \begin{algorithmic}[1]
        \Require Input: $X^c$ a feature vector with historical values
        \Ensure Output: $X_l^c$ and $X_h^c$ low and high frequency components

        \State $L_m \gets [\,]$ \Comment{Initialize an array to keep $m$ values below}
        \State $L_l \gets [\,]$ \Comment{Initialize an array to keep low frequency components}
        \State $L_h \gets [\,]$ \Comment{Initialize an array to keep high frequency components}
        \For{$f$ in $F$}\Comment{loop through different frequency values}
            \State $X^{c,f}_l \gets$ \text{Butterworth}$(X^c, f)$ \Comment{extract low frequency component for $f$}
            \State $X^{c,f}_h \gets$ \text{Butterworth}$(X^c,f)$ \Comment{extract high frequency component for $f$}
            \State $m_{mse} \gets$ \text{MSE}$(X^c,X^{c,f}_l)$ \Comment{get MSE value between $X$ and $X^{c,f}_l$}
            \State $m_{cos} \gets$ \text{CosineSim}$(X^c,X^{c,f}_h)$ \Comment{get Cosine similarity value between $X$ and $X^{c,f}_h$}
        
            \State $m = \alpha*m_{mse} + (1-\alpha)*1/m_{cos}$ \Comment{get weighted summation of $m_{mse}$ and $m_{cos}$}
            \State $L_m.\text{append}(m)$ \Comment{Add $m$ to the list, $L_m$}
            \State $L_l.\text{append}(X^{c,f}_l)$ \Comment{Add $X^c_l$ to the list, $L_l$}
            \State $L_h.\text{append}(X^{c,f}_h)$ \Comment{Add $X^c_h$ to the list, $L_h$}
            
            \EndFor
            
        \State $i \gets \text{argmin}(L_m)$ \Comment{get index of the minimum $m$ values}
        \State $f_{cut} \gets F[i]$ \Comment{get the best $f$ denoted by $f_{cut}$ for the minimum $m$}
        \State $X_l^c \gets L_l[i]$ \Comment{get  related low frequency component at $f_{cut}$}
        \State $X_h^c \gets L_h[i]$ \Comment{get related high frequency component at $f_{cut}$}
    \end{algorithmic}
\end{algorithm}

\subsection{Introductory prompt used in LLM prediction}\label{append-subsec:intro prompt}

We add an introductory prompt at the beginning of the numerical sequences to help guide the LLM in performing the task. The introductory prompt highlights the following conditions:

\begin{itemize}
    \item Instruct the LLM to consider the distribution provided.
    \item Specify the number of tokens or lines to predict. Note that we already set the \texttt{max\_tokens} parameter in the LLM generator.
    \item For multivariate generation, each feature is transformed into a specific value range (see $\S~\ref{subsec:multi gen}$). Since the integer part of each feature remains the same during this transformation, we explicitly ask the LLM to keep the integer part of each predicted feature the same as in the input historical data.
    \item Ensure that the predictions are numerical only.
\end{itemize}

A sample introductory prompt that covers these conditions is:

\begin{quote}
\textit{``Consider the distribution. Predict the next few lines. INTEGER component of the value SHOULD be SAME as the train data. ONLY provide numerical values.''}
\end{quote}

\subsection{Information of post-processing}\label{append-subsec-post processing}

\subsubsection{MLP architecture used}

\begin{figure}[h]
    \centering
    \includegraphics[width=0.8\textwidth]{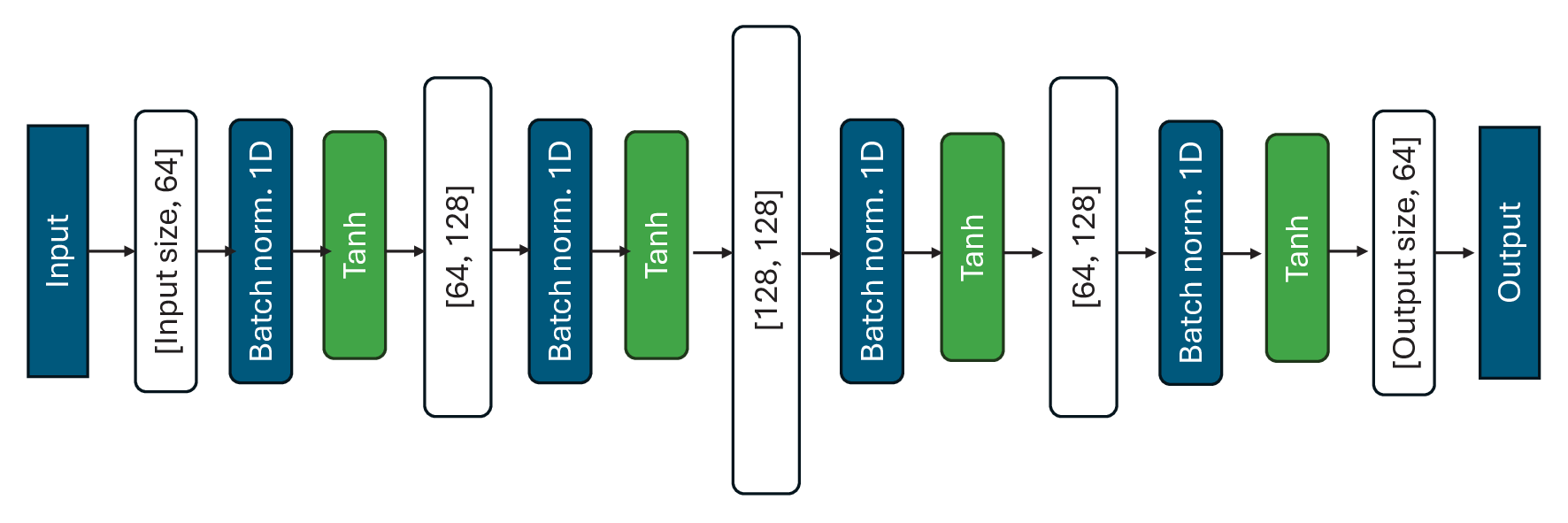}
    \caption{MLP architecture used in the post-processing steps for $X^c_l$ of the LLM output. Each hidden layer except the last layer is followed by a 1D batch normalization and Tanh activation function.}
    \label{fig:mlp}
\end{figure}

Fig.~\ref{fig:mlp} illustrates the proposed MLP architecture, which consists of five hidden layers with \texttt{tanh} activation to effectively handle values within the range $[-1, 1]$. To stabilize training and accelerate convergence, we apply a batch normalization layer. We set the batch size to 32, learning rate to 0.0001 with Adam optimizer, and the number of epochs to 32 for the Llama and Deepseek models, and 128 for the GPT-4o-mini. For the Llama and Deepseek models, we observed overfitting as the number of epochs increased, so we kept the epoch count relatively low. During MLP training we apply a 70/30\% train/test splits.

\subsubsection{Gaussian distribution of high frequency component values}\label{subsubsec:gaussian plotsww}
\begin{figure}[h!]
    \centering
    \begin{subfigure}[b]{0.24\textwidth}
        \centering
        \includegraphics[width=0.95\textwidth]{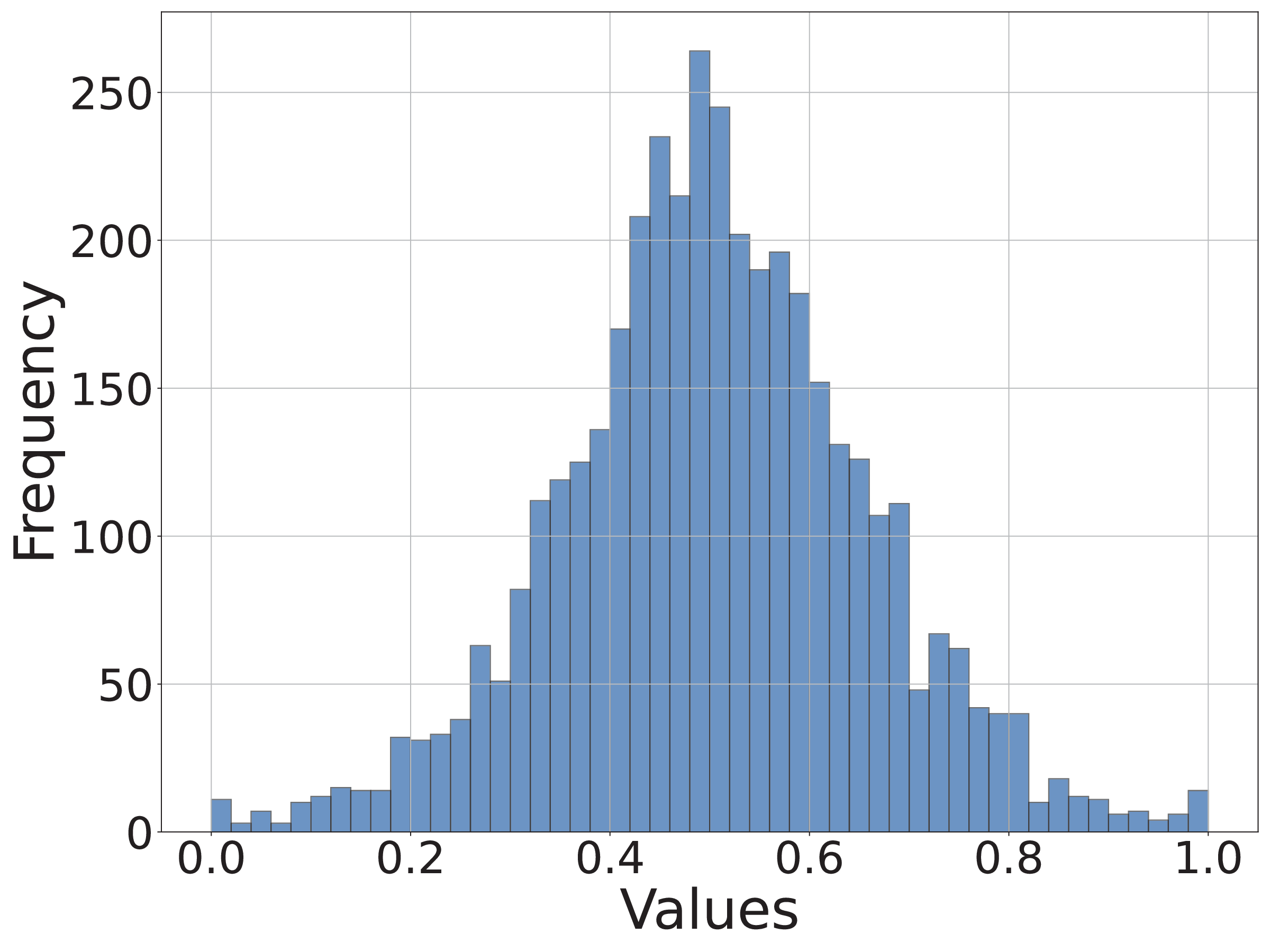}
        \caption{ETTh1}
    \end{subfigure}
    \begin{subfigure}[b]{0.24\textwidth}
        \centering
        \includegraphics[width=0.95\textwidth]{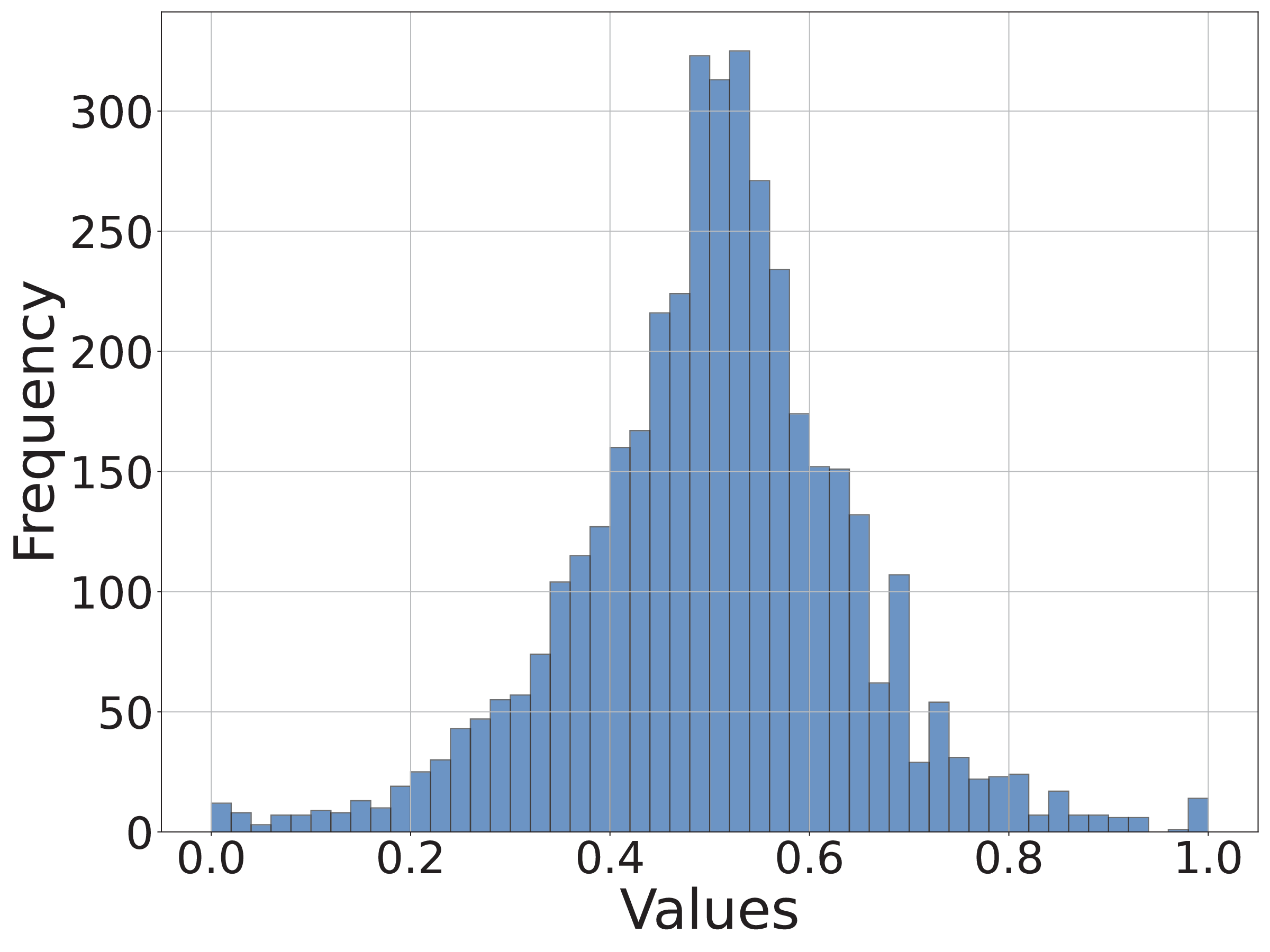}
        \caption{ETTm1}
    \end{subfigure}
    \begin{subfigure}[b]{0.24\textwidth}
        \centering
        \includegraphics[width=0.95\textwidth]{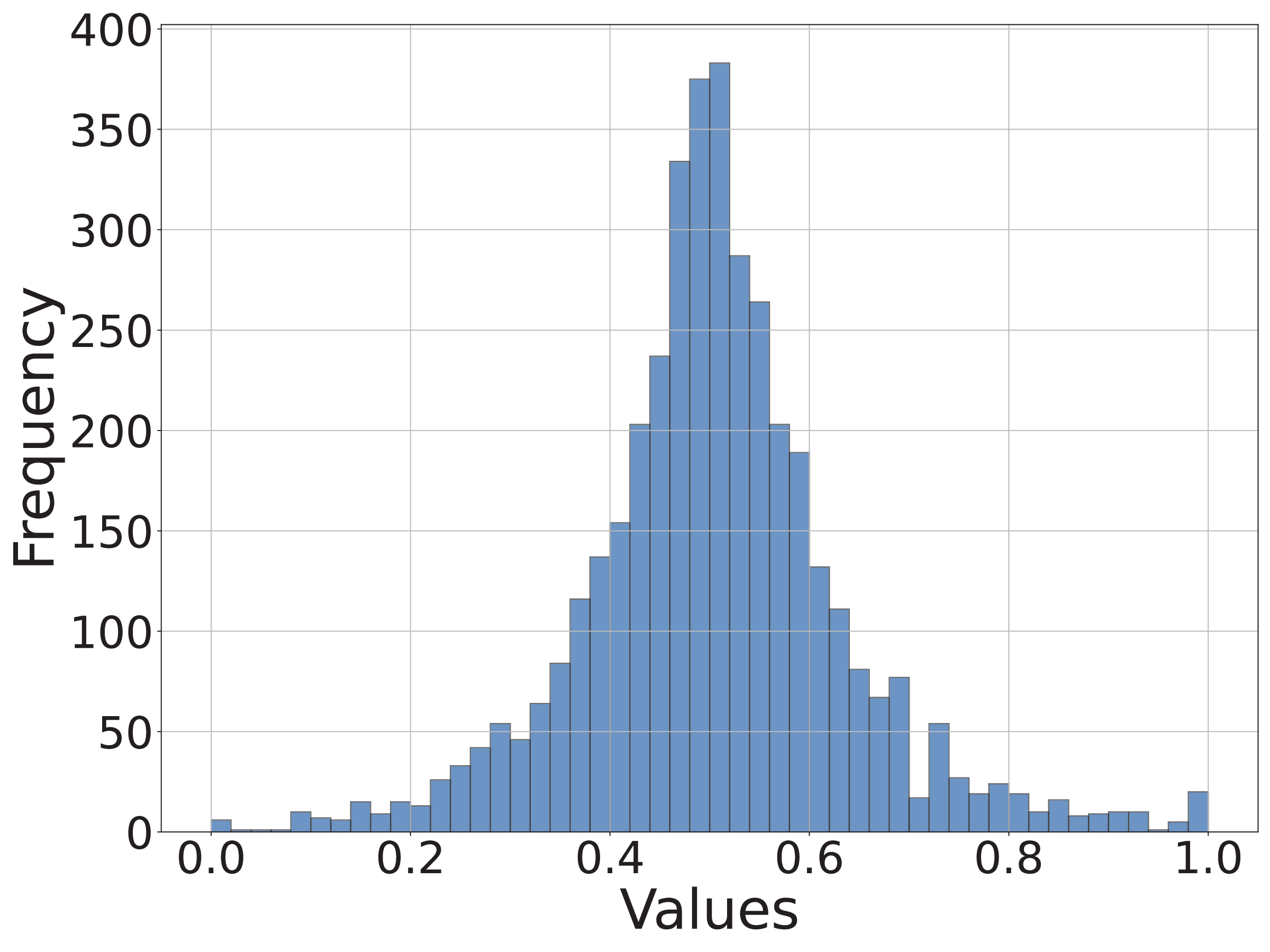}
        \caption{Weather}
    \end{subfigure}
    \begin{subfigure}[b]{0.24\textwidth}
        \centering
        \includegraphics[width=0.95\textwidth]{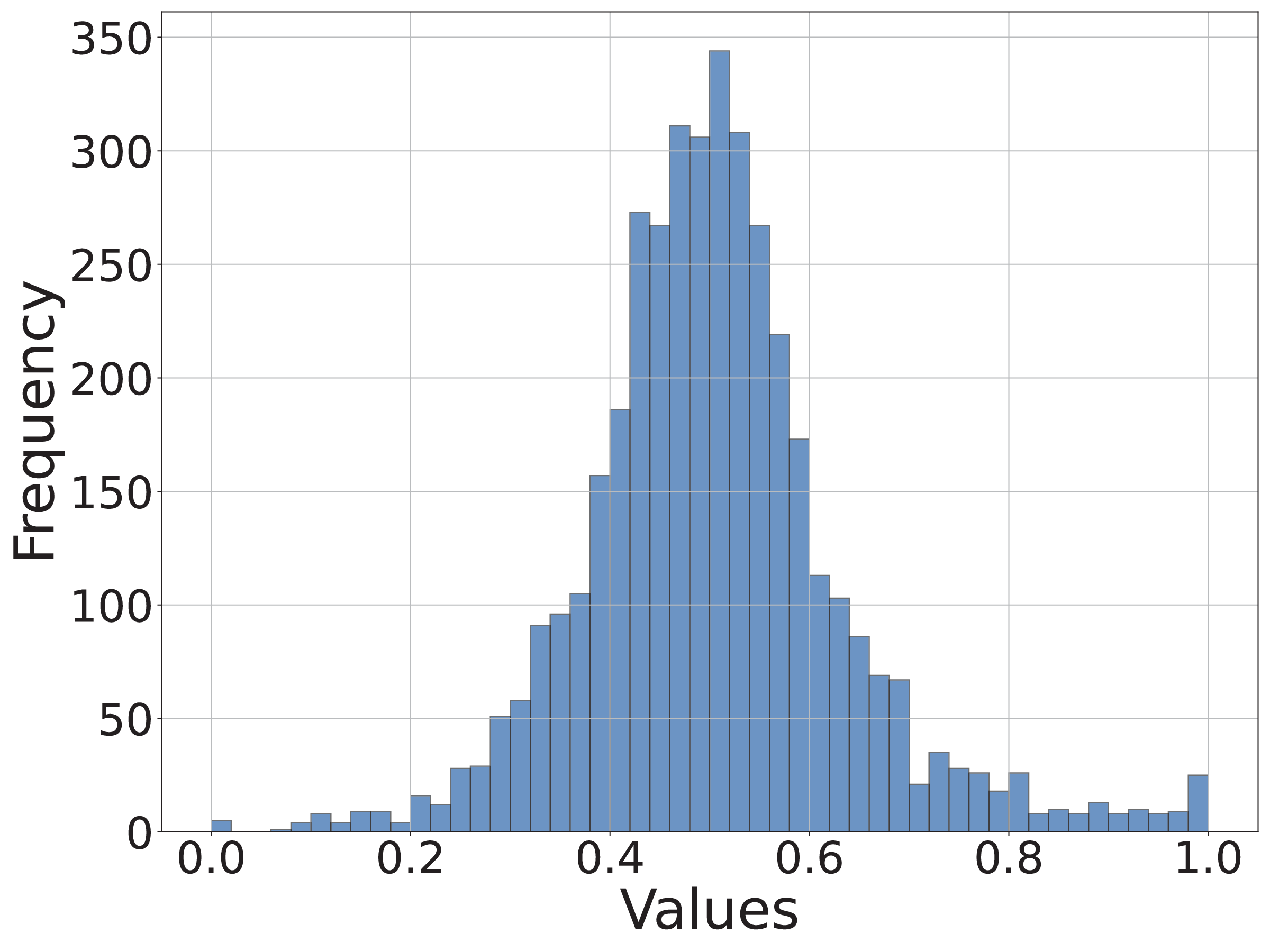}
        \caption{Electricity}
    \end{subfigure}
    \caption{Histogram plots of the value distribution in high frequency component $\alpha=0.5$ }
    \label{fig:alpha 0.5}
\end{figure}

\begin{figure}[h!]
    \centering
    \begin{subfigure}[b]{0.24\textwidth}
        \centering
        \includegraphics[width=0.95\textwidth]{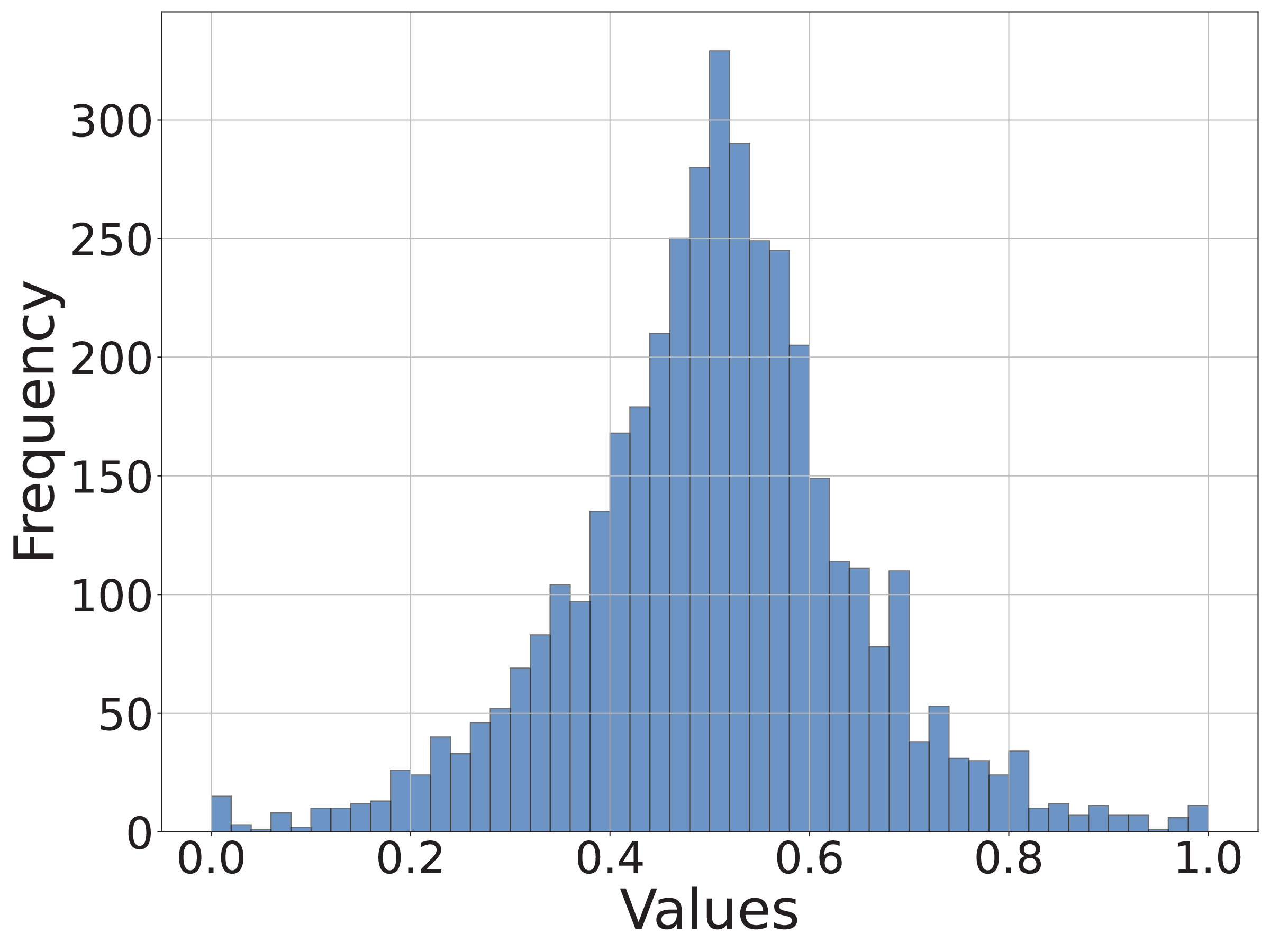}
        \caption{ETTh1}
    \end{subfigure}
    \begin{subfigure}[b]{0.24\textwidth}
        \centering
        \includegraphics[width=0.95\textwidth]{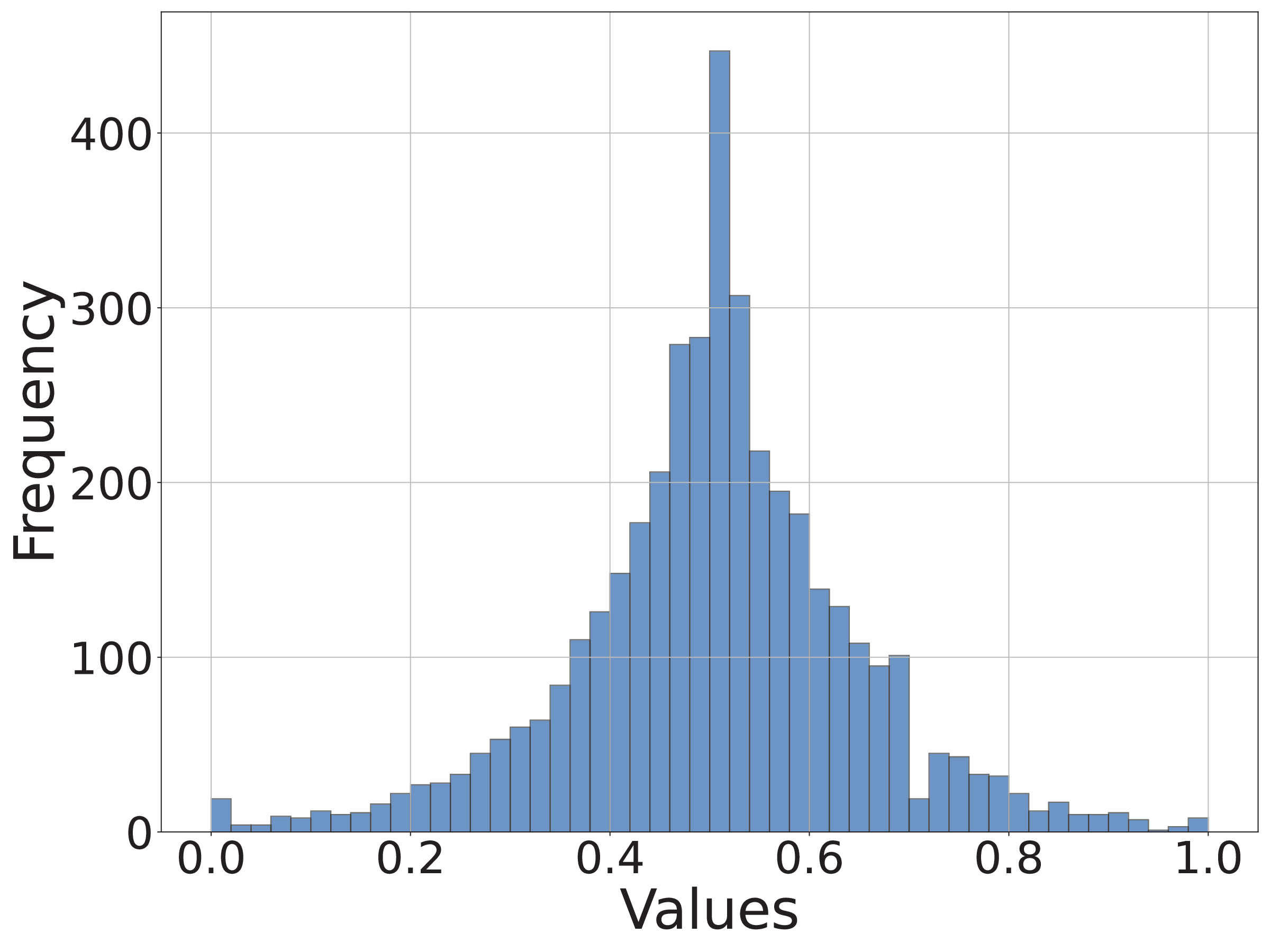}
        \caption{ETTm1}
    \end{subfigure}
    \begin{subfigure}[b]{0.24\textwidth}
        \centering
        \includegraphics[width=0.95\textwidth]{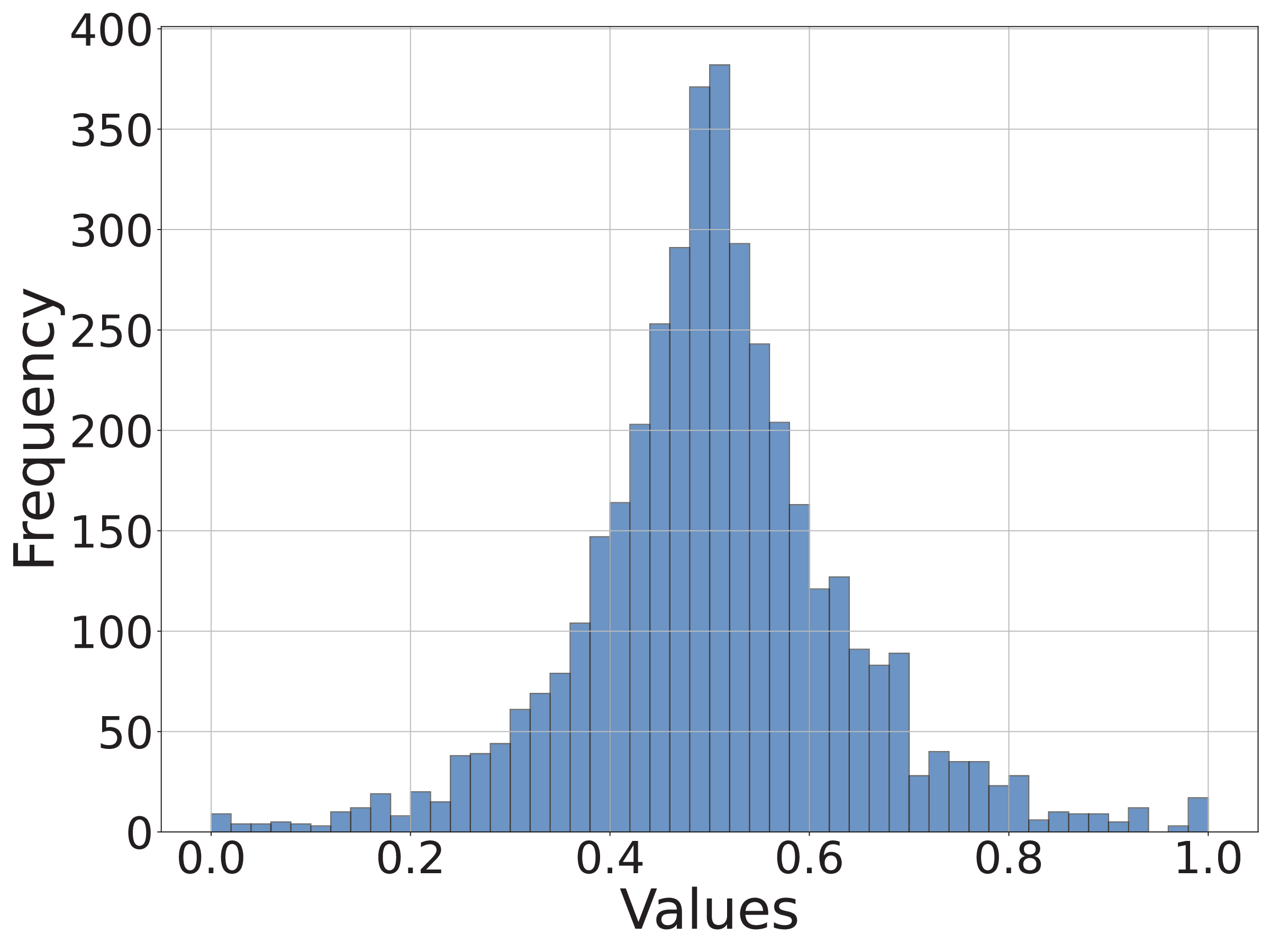}
        \caption{Weather}
    \end{subfigure}
    \begin{subfigure}[b]{0.24\textwidth}
        \centering
        \includegraphics[width=0.95\textwidth]{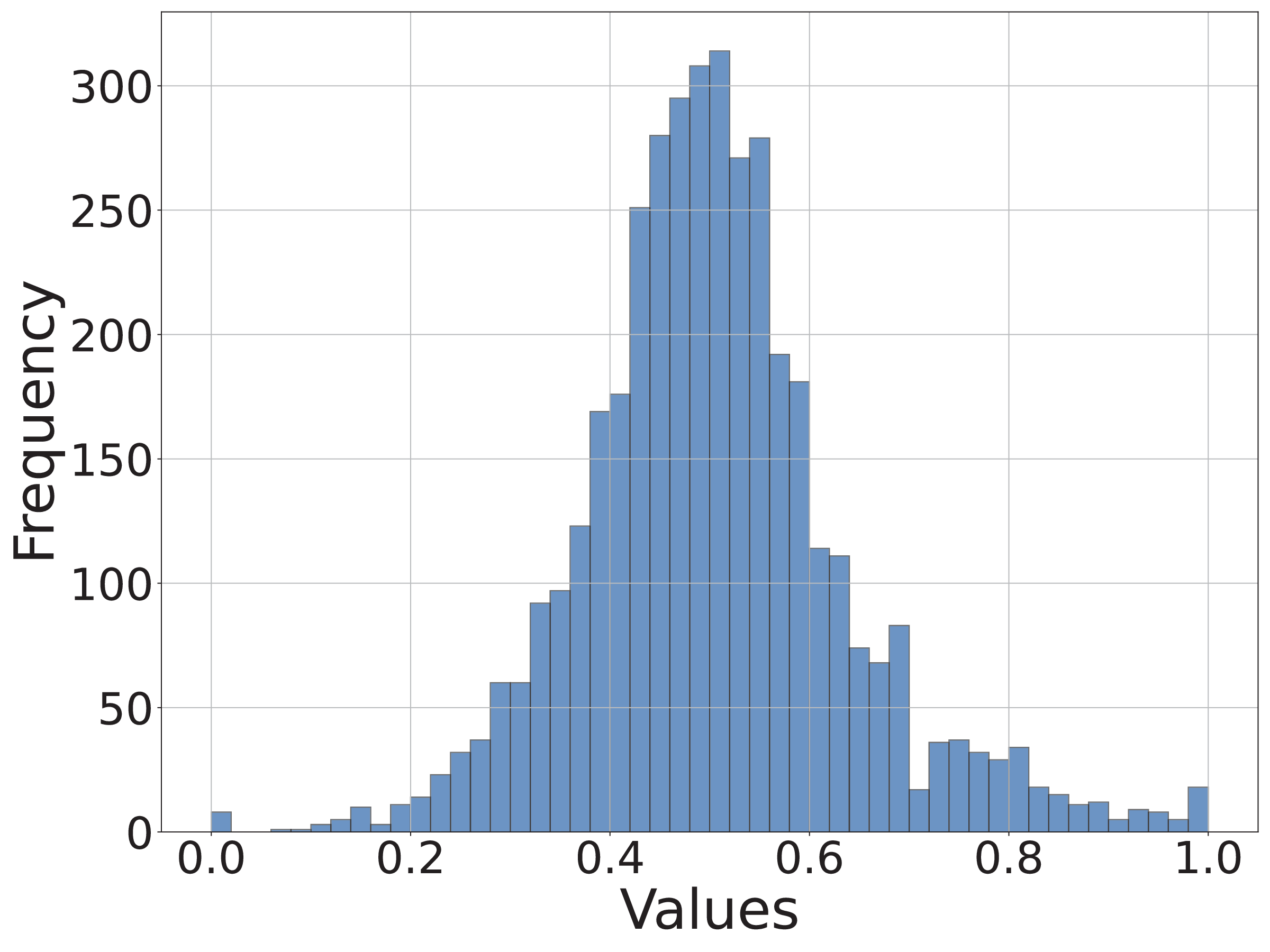}
        \caption{Electricity}
    \end{subfigure}
    \caption{Histogram plots of the value distribution in high frequency component $\alpha=0.7$ }
    \label{fig:alpha 0.7}
\end{figure}

 In Fig.~\ref{fig:alpha 0.5} and Fig.~\ref{fig:alpha 0.7} shows the high frequency value component of for $\alpha=0.5$ and $\alpha=0.7$ respectively for the first feature from each dataset. We notice that for different $\alpha$ values, the values show a Gaussian distribution. 

\section{Evaluation details}\label{append-sec:eval}


\subsection{LLM models used in \solution{}}

\paragraph{Llama-2 7B:}
Llama-2-7B, a 7-billion-parameter model developed by Meta, was trained on 2 trillion tokens. Its relatively smaller size makes it suitable for deployment in resource-constrained environments with limited computational capacity. However, as an earlier model, it supports a maximum context length of only 4,096 tokens. 

\paragraph{Llama-3.2 3B:}
Llama-3.2-3B is a multilingual model developed by Meta with 3 billion parameters. It supports a context length of up to 128K tokens, enabling the processing of larger text inputs—an important feature for multivariate prediction tasks. Since this model has a relatively small number of parameters, it is well-suited for deployment in low-resource environments as well. 

\paragraph{DeepSeek-base 7B:}
DeepSeek-7B-Base is a bilingual model (English and Chinese) trained on 2 trillion tokens. It supports a context length of only 4,096 tokens. With a relatively small parameter count, it is efficient for deployment in low-resource environments. However, its limited context length restricts its capability for long-term predictions.

\paragraph{GPT-4o Mini:}
GPT-4o Mini is a lightweight, cost-effective variant of GPT-4o, featuring a 128K context window and a 16K output limit. Its extended context makes it suitable for long-term and multivariate predictions, while its lightweight nature enables faster and more efficient inference.

\subsection{LLM model parameters used}

\paragraph{Llama-7B, Llama-3B, Deepseek-7B}

We use the standard tokenizers provided with each LLM, configured with \texttt{return\_tensors=``pt''} to convert the tokenized output into \texttt{PyTorch} tensors. In the LLM generator model, we set \texttt{max\_tokens} equal to the number of input tokens, as \solution{} consider the prediction length is the same as the historical time-series length. To enhance the creativity and randomness of the predictions, taking insights from~\cite{gruver2023large} and empirical analysis, we enable \texttt{do\_sample} and configure the sampling parameters as: \texttt{temperature} = 1.0, \texttt{top\_p} = 0.9, and \texttt{renormalize\_logits} = False. All other generation parameters are kept at their default settings.

\paragraph{GPT-4o-mini}
For GPT-4o-mini, we use the standard APIs provided by OpenAI and follow the default message format (e.g., \texttt{[``role'': ``user'', ``content'': \textbf{user prompt}]}). Based on this, we create a \texttt{client} wrapper that calls the \texttt{chat.completions.create()} method, where \texttt{max\_tokens} is set to match the number of input tokens and \texttt{temperature} is set to 1.0. All other parameters are left at their default values.

\subsection{Computing resources}\label{append-subsec:compute resources}

The compute resources used for our LLM models vary depending on whether the task involves univariate or multivariate generation. For univariate modeling, which involves a lower number of tokens compared to multivariate generation, we utilize systems equipped with either an NVIDIA A40 GPU (48~GB) or an NVIDIA GeForce RTX 4090 GPU (24~GB), both supported by 16~GB of main memory and 6 CPU cores.

In contrast, multivariate generation demands greater computational and memory resources. For these tasks, we conduct experiments using an NVIDIA A40 GPU (48~GB) with 24~GB of main memory and 8 CPU cores, as well as an NVIDIA A100 GPU (40~GB) paired with 24~GB of main memory and 8 CPU cores.

\subsection{Evaluation metrics used}

\paragraph{MSE} Mean Squared Error is a widely used regression metric that measures the average of the squares of the errors between predicted and true values. It is defined as:
\[
\text{MSE} = \frac{1}{n} \sum_{i=1}^{n} (y_i - \hat{y}_i)^2
\]
where $y_i$ and $\hat{y}_i$ are the ground truth and predicted values, respectively. MSE penalizes larger errors more heavily, making it sensitive to outliers.

\paragraph{MAE}
{Mean Absolute Error}: MAE quantifies the average absolute difference between predicted and actual values:
\[
\text{MAE} = \frac{1}{n} \sum_{i=1}^{n} |y_i - \hat{y}_i|
\]
It is more robust to outliers compared to MSE and provides a straightforward interpretation in terms of average prediction error.

\paragraph{Kolmogorov-Smirnov (KS)-statistic}
The KS statistic measures the maximum distance between the empirical cumulative distribution functions (CDFs) of the predicted and actual values. Formally:
\[
\text{KS} = \sup_x |F_1(x) - F_2(x)|
\]
where $F_1$ and $F_2$ are the empirical CDFs of the predicted and actual data, respectively. It is used to assess the similarity between two distributions, often applied to evaluate the distributional alignment of time series predictions.

\subsection{Benchmarks}\label{append:subsec-benchmarks}

\paragraph{Informer~\cite{zhou2021informer}}
This model uses a transformer-based approach for long-term time series forecasting, addressing limitations such as quadratic time complexity, high memory usage, and inherent constraints of the encoder-decoder architecture. It introduces a self-attention mechanism called ProbSparse, which reduces complexity and memory usage by focusing on dominant attention patterns and halving the inputs to cascading layers. Additionally, it proposes a generative-style decoder that predicts long-term time series sequences in a single forward pass, thereby improving inference speed.

\paragraph{Autoformer~\cite{wu2021autoformer}}

This study also addresses long-term forecasting by leveraging transformers and proposing a novel time-series decomposition architecture with an auto-correlation mechanism. To enable progressive decomposition of sequences, the decomposition module is embedded into deep learning models, allowing for dependency discovery and representation aggregation at the sub-series level.

\paragraph{Fedformer~\cite{zhou2022fedformer} }


In this study, the authors address the issues of high computational cost and the inability to capture the global view of time series. They incorporate a seasonal-trend decomposition method alongside the transformer architecture, where the decomposition module captures global patterns and the transformer focuses on detailed structural patterns. Furthermore, they enhance the transformer model by making it aware of the frequency distribution of the time series.

\paragraph{Autotimes~\cite{liu2024autotimes}}
This mechanism models time-series forecasting as a language modeling task by projecting time series into the embedding space of LLM understandable tokens and generating future values autoregressively.  AutoTimes uses {in-context forecasting} by treating past time series as prompts, effectively extending the prediction context. Additionally, it embeds textual timestamps to incorporate chronological patterns to the prediction.

\paragraph{LLMTimes~\cite{gruver2023large}}
This method treats time series forecasting as a next-token prediction task by converting time series into sequences of digits. It leverages pre-trained large language models such as GPT-3 and LLaMA-2 to make future predictions. The novelty of this approach lies in how it tokenizes time series data and maps token outputs back to continuous values. It also demonstrates that LLMs can handle missing data and answer questions to help explain their predictions.

\section{Further results}

\subsection{Univariate analysis}\label{append-subsec:uni analysis}
\begin{figure}[h!]
    \centering
    \begin{subfigure}[b]{0.48\textwidth}
        \centering
        \includegraphics[width=\textwidth]{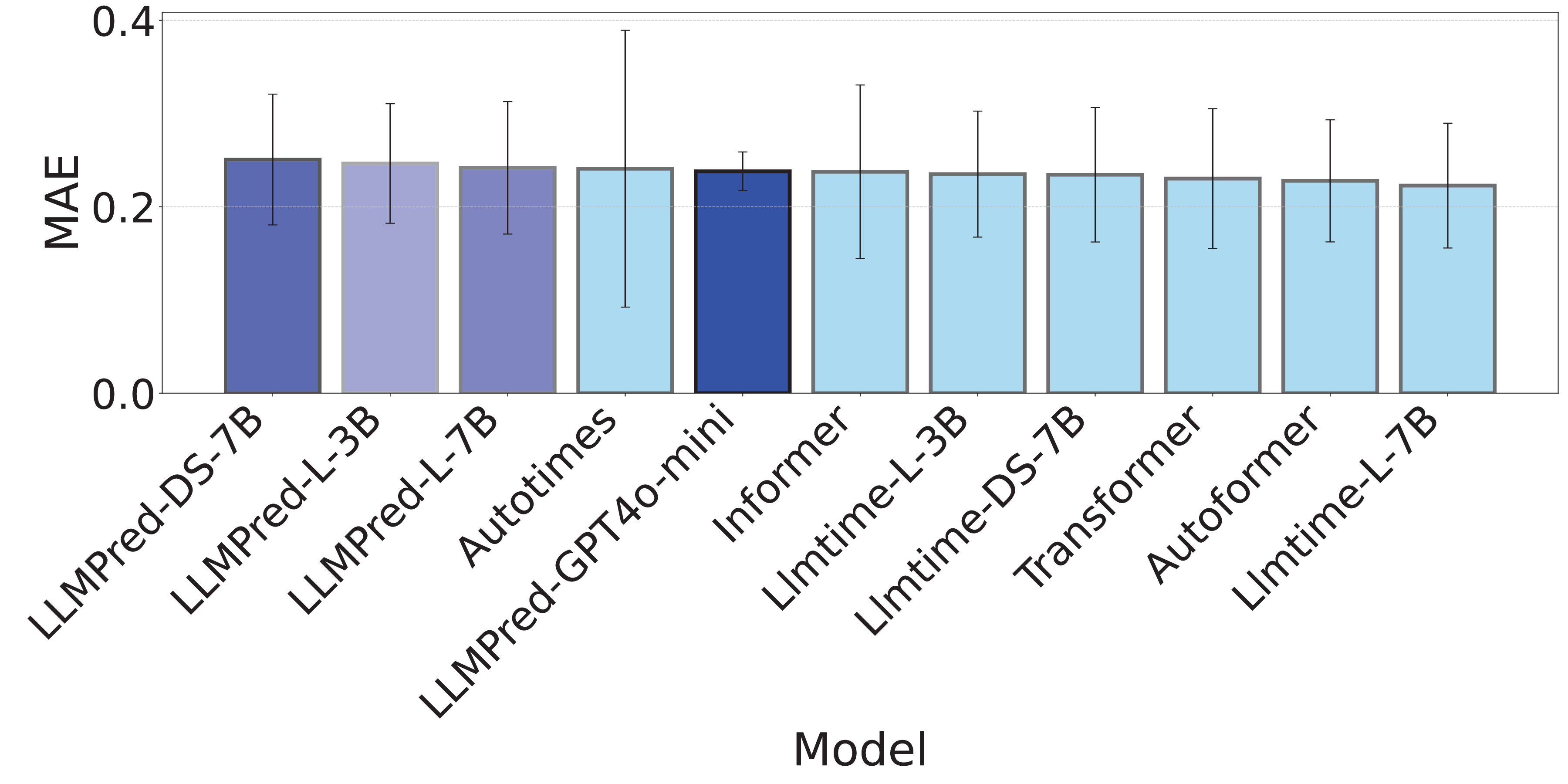}
        \caption{Univariate - MAE}
        \label{fig:uni-mae-all}
    \end{subfigure}
    \begin{subfigure}[b]{0.48\textwidth}
        \centering
        \includegraphics[width=\textwidth]{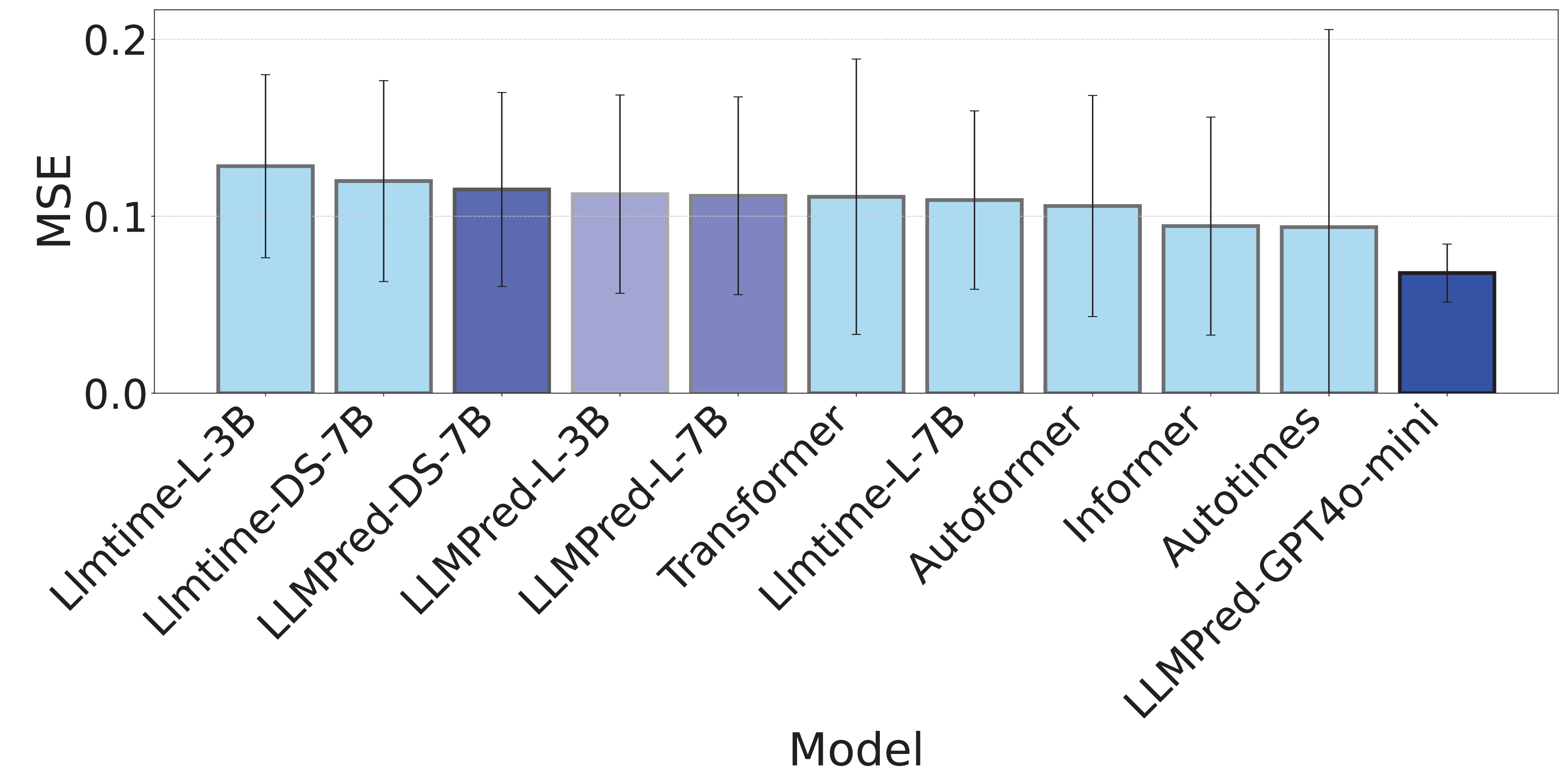}
        \caption{Univariate - MSE prediction length 48}
        \label{fig:uni-mse-48}
    \end{subfigure}
    \begin{subfigure}[b]{0.48\textwidth}
        \centering
        \includegraphics[width=\textwidth]{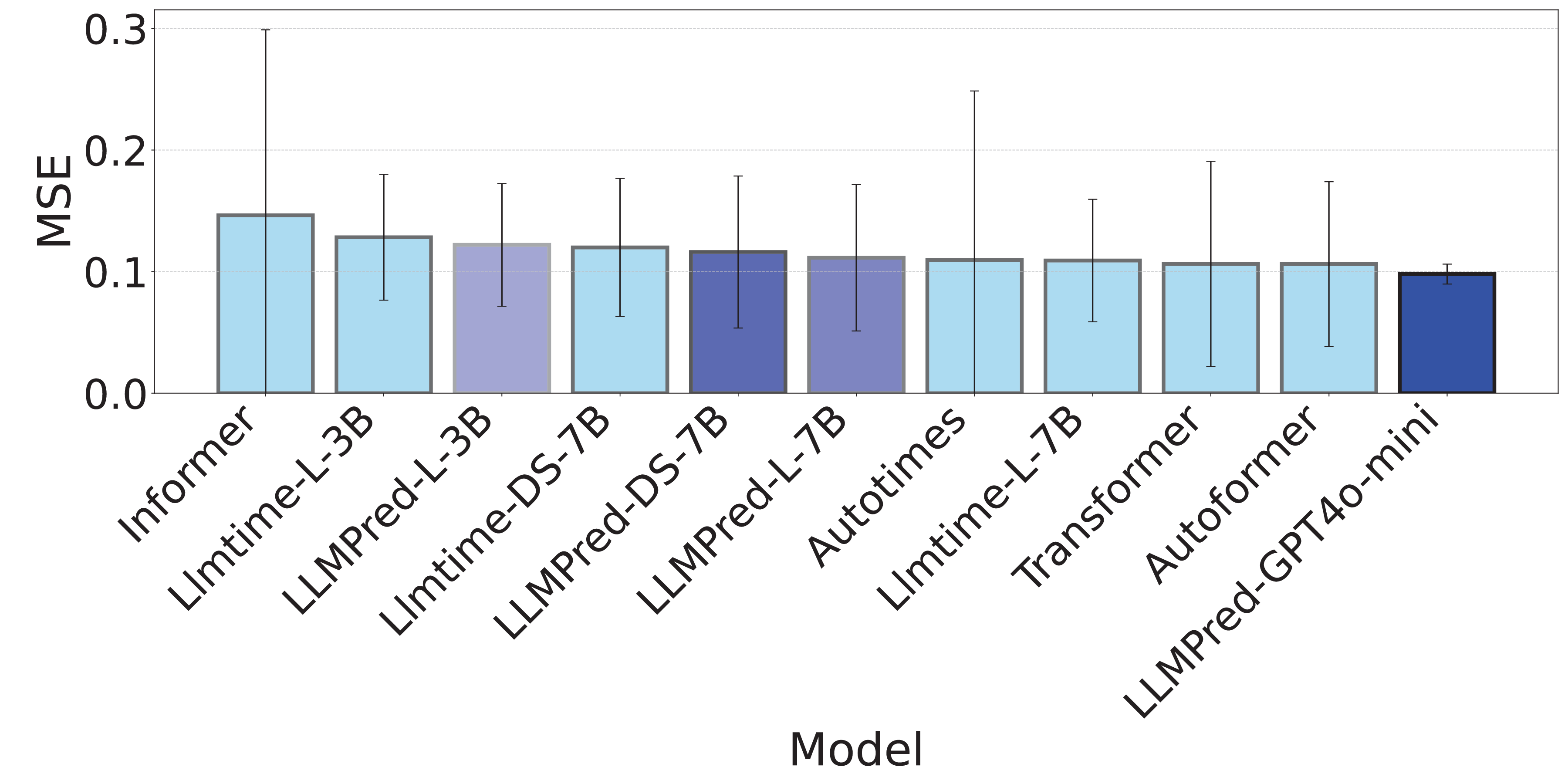}
        \caption{Univariate - MSE prediction length 96}
        \label{fig:uni-mse-96}
    \end{subfigure}
    \caption{Comparison with benchmarks - Univariate prediction}
    \label{fig:uni-further}
\end{figure}

 As shown in Fig.~\ref{fig:uni-mae-all}, \solution{} exhibits relatively lower performance in terms of MAE. This outcome can be attributed to the nature of MAE, which penalizes frequent small errors equally. While \solution{} effectively suppresses large outliers—reflected by its strong MSE performance—small, recurring deviations persist, particularly due to the high-frequency component ($X_h^c$) in the signal. Since $X_h^c$ captures the noisier part of the distribution, the absolute differences between predicted and ground truth values at individual time points remain minor but frequent. We view this as a promising direction for future work, where further refinement of \solution{} could target these residual high-frequency errors to improve MAE without compromising its robustness to outliers.

Fig.~\ref{fig:uni-mse-48} and Fig.~\ref{fig:uni-mse-96} show the MSE for the prediction window length 48 and 96 respectively. As we observed in $\S$~\ref{subsec:uni-analysis}, for both prediction lengths \solution{}-GPT4o-mini shows the best performance with the lowest variance. 

\subsection{Multivariate analysis}\label{append:subsec:multi}

\begin{figure}[h!]
    \centering
    \begin{subfigure}[b]{0.32\textwidth}
        \centering
        \includegraphics[width=\textwidth]{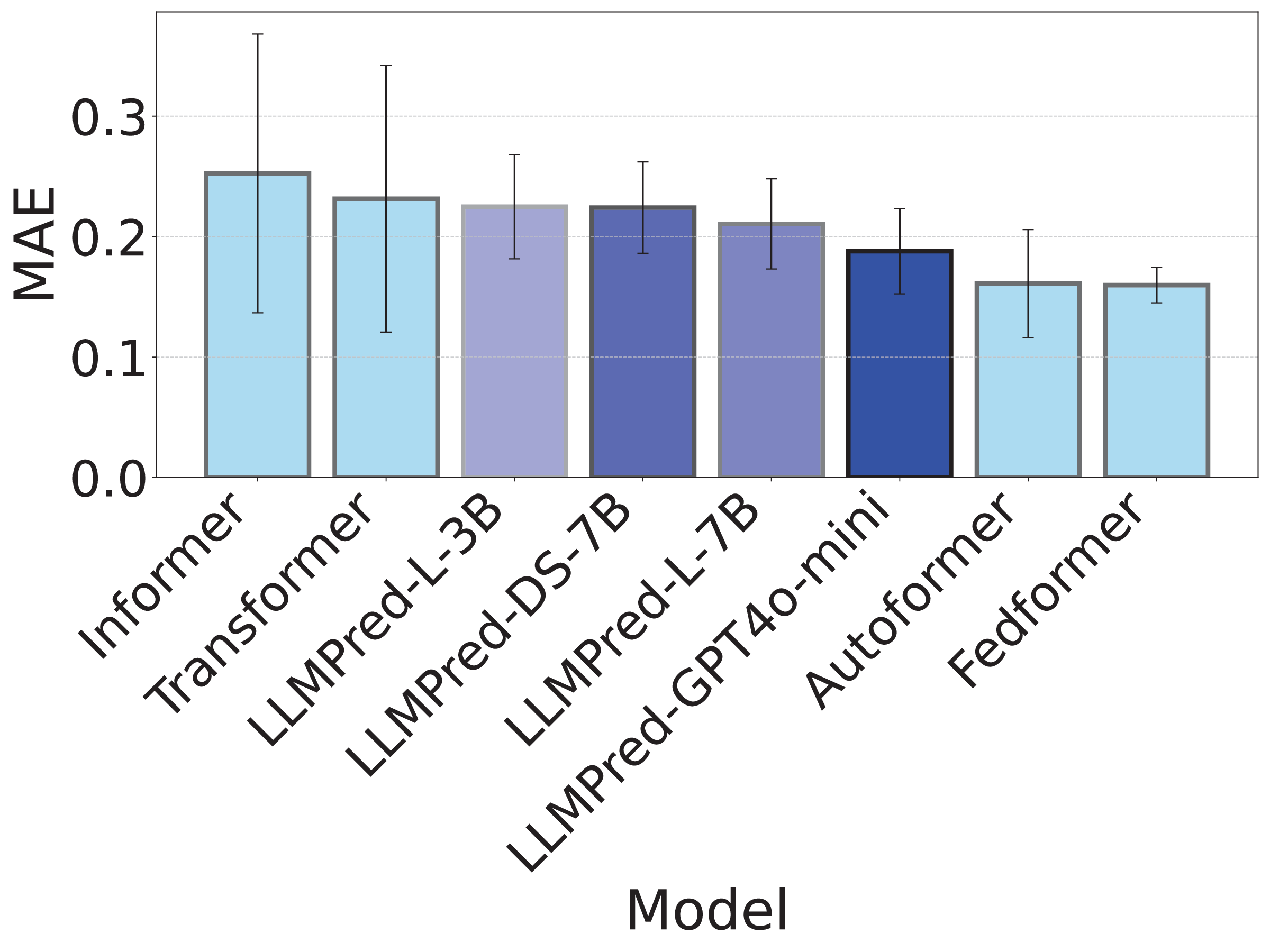}
        \caption{Multivariate - MAE for prediction lengths 48 and 96}
        \label{fig:multi-mae-all}
    \end{subfigure}
    \begin{subfigure}[b]{0.32\textwidth}
        \centering
        \includegraphics[width=\textwidth]{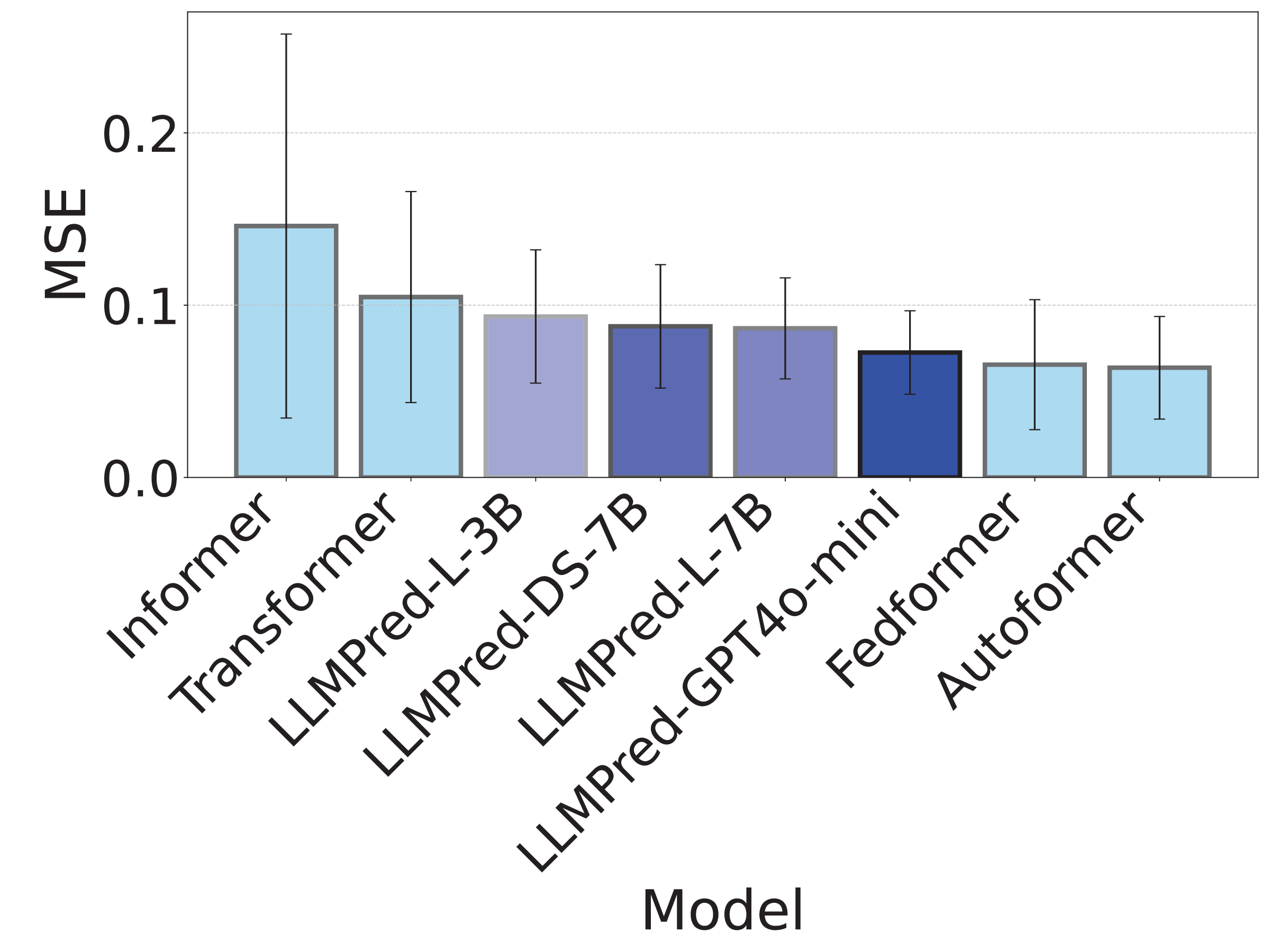}
        \caption{Multivariate - MSE prediction length 48}
        \label{fig:multi-mse-48}
    \end{subfigure}
    \begin{subfigure}[b]{0.32\textwidth}
        \centering
        \includegraphics[width=\textwidth]{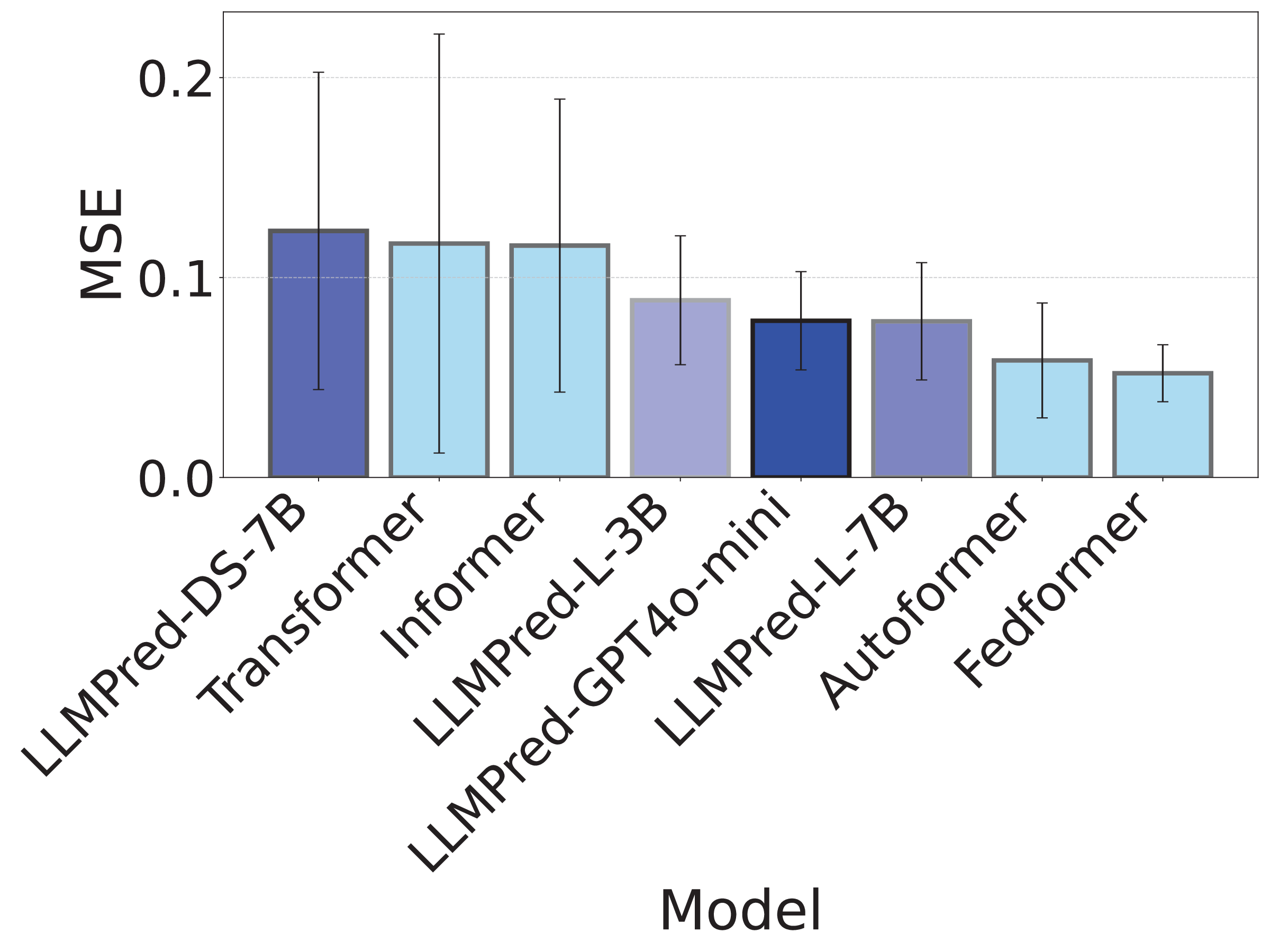}
        \caption{Multivariate - MSE prediction length 96}
        \label{fig:multi-mse-96}
    \end{subfigure}
    \caption{Comparison with benchmarks - Multivariate prediction}
    \label{fig:multi-further}
\end{figure}

Fig.~\ref{fig:multi-further} presents a comparison of MAE across both 48- and 96-step prediction lengths (Fig.~\ref{fig:multi-mae-all}), and MSE separately for 48 (Fig.~\ref{fig:multi-mse-48}) and 96 (Fig.~\ref{fig:multi-mse-96}) steps. Consistent with our observations in $\S$\ref{subsec:multi}, \solution{} ranks third in overall performance and demonstrates strong robustness across datasets, evidenced by a lower average standard deviation compared to other benchmarks. For instance, \solution{} variants yield an average MAE standard deviation of 0.039, while benchmarks exhibit a higher deviation of 0.057. This indicates that \solution{}’s multivariate generation maintains more consistent performance across diverse datasets.

\subsection{Further results on Ablation}

\subsubsection{Cut-off frequency ($f_{cut}$) vs $\alpha$}\label{append-subsubsec:cut-off-dist}

\begin{figure}[h!]
    \centering
    \begin{subfigure}[b]{0.75\textwidth}
        \centering
        \includegraphics[width=0.99\textwidth]{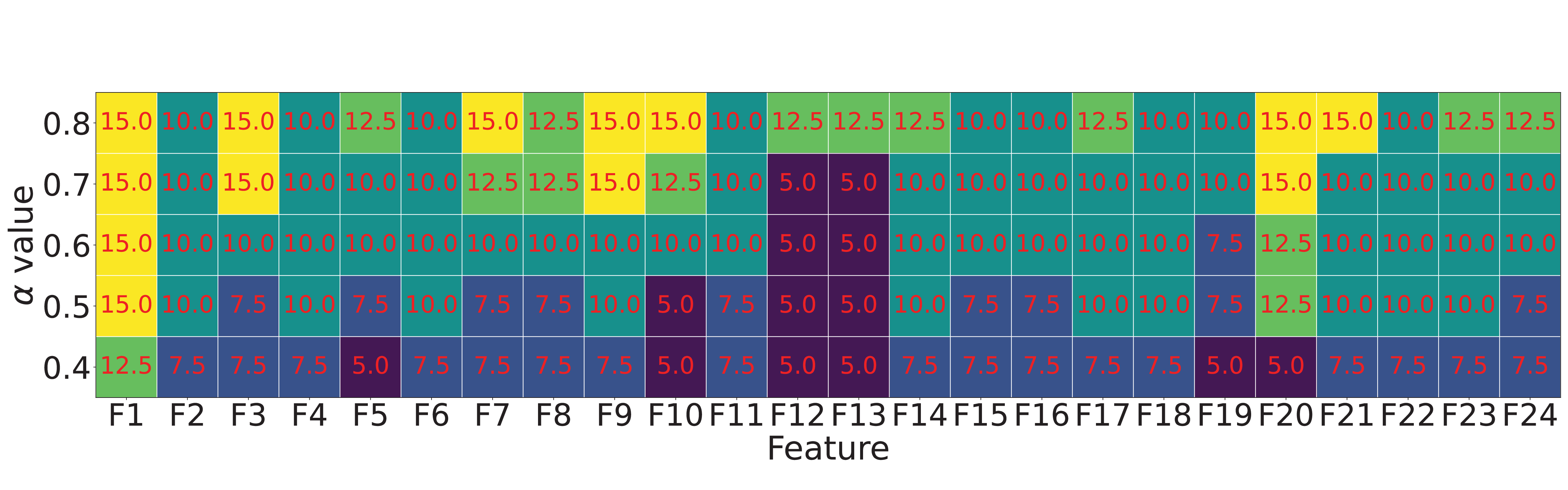}
        \caption{Electricity}
        \label{fig:f_{cut} vs alpha electricty}
    \end{subfigure}
    \begin{subfigure}[b]{0.75\textwidth}
        \centering
        \includegraphics[width=0.99\textwidth]{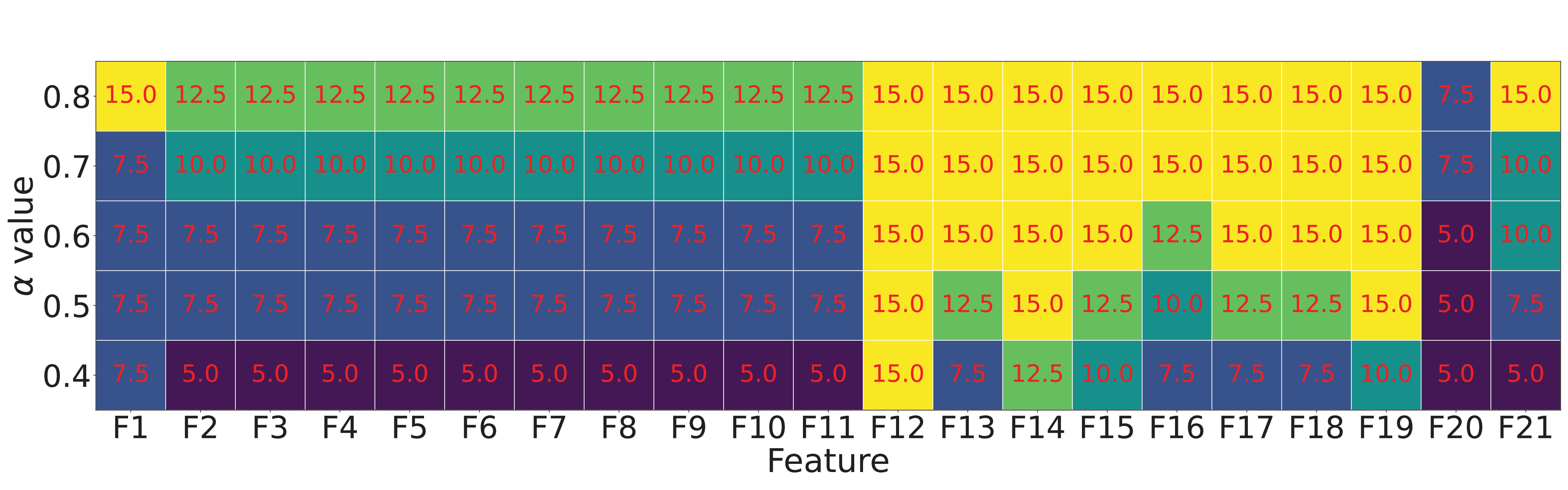}
        \caption{Weather}
    \end{subfigure}
    \begin{subfigure}[b]{0.26\textwidth}
        \centering
        \includegraphics[width=0.99\textwidth]{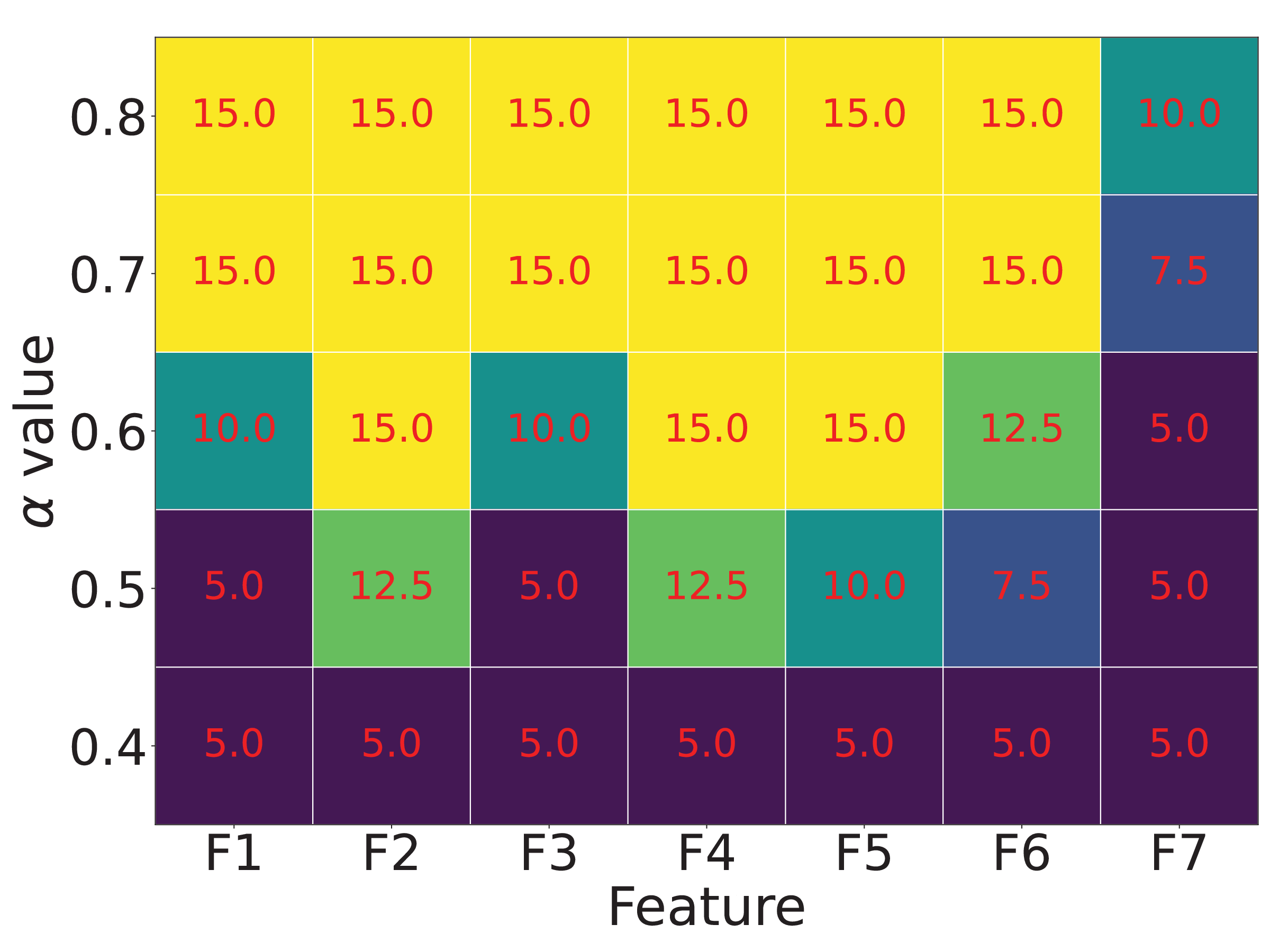}
        \caption{ETTh2}
    \end{subfigure}
    \begin{subfigure}[b]{0.26\textwidth}
        \centering
        \includegraphics[width=0.99\textwidth]{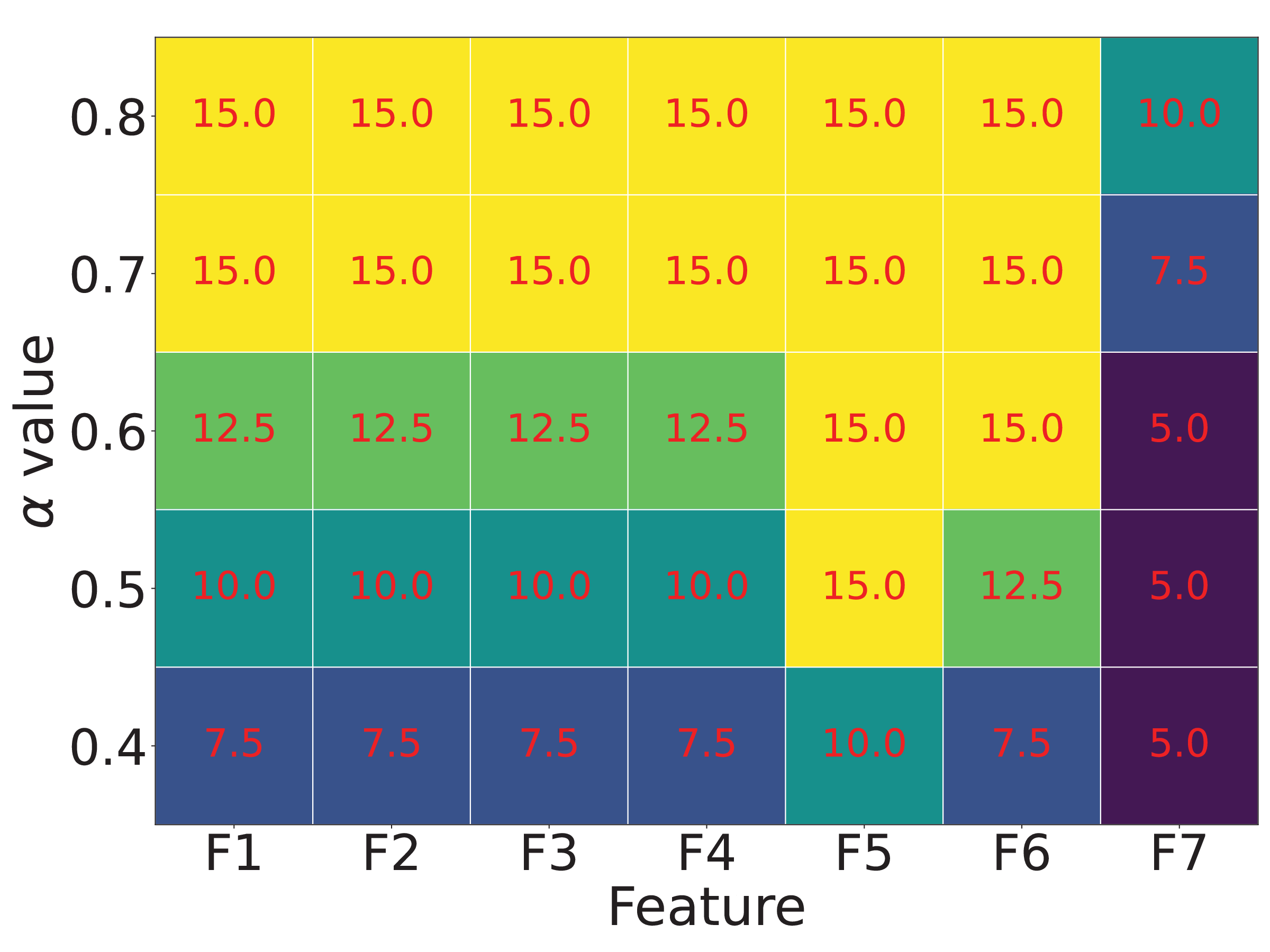}
        \caption{ETTm2}
        \label{fig:f_{cut} vs alpha ettm2}
    \end{subfigure}
    \begin{subfigure}[b]{0.23\textwidth}
        \centering
        \includegraphics[width=0.99\textwidth]{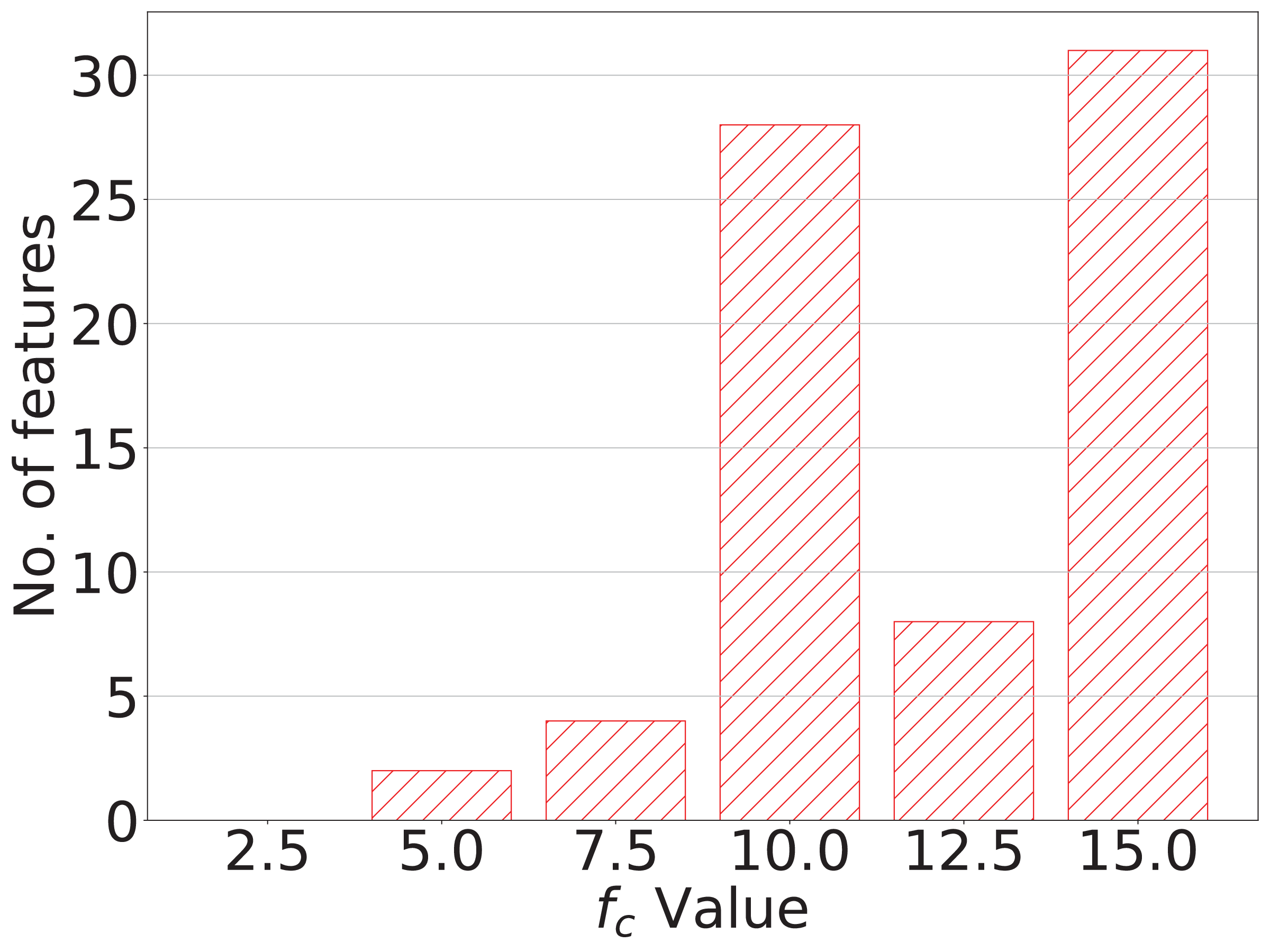}
        \caption{No. of features with different $f_{cut}$ at $\alpha=0.7$}
        \label{fig:no of feat and fc}
    \end{subfigure}
    \caption{Variations in selection of cut-off frequencies for different $\alpha$ values. From Fig.~\ref{fig:f_{cut} vs alpha electricty} to Fig.~\ref{fig:f_{cut} vs alpha ettm2} higher the $f_{cut}$ value brighter the color in the cell.}
    \label{fig:append-fc-vs-alpha}
\end{figure}

Fig.~\ref{fig:append-fc-vs-alpha} presents the distribution of $f_{cut}$ values across different $\alpha$ settings for various datasets. Consistent with our observations in \S~\ref{subsubsec:cut-off vs alpha}, we observe that increasing $\alpha$ generally leads to higher selected $f_{cut}$ values across datasets such as ETTh2, ETTm1, and Weather. Notably, for the Weather dataset, features 12 to 19 predominantly select 15.0 Hz as the $f_{cut}$ value across a wide range of $\alpha$ values. Fig.~\ref{fig:no of feat and fc} illustrates the number of features, aggregated over all datasets, that are decomposed using each $f_{cut}$ value when $\alpha=0.7$. As discussed in \S~\ref{subsubsec:cut-off vs alpha}, at this $\alpha$ level, the majority of features select $f_{cut}$ values in the 10.0--15.0 Hz range. For instance, Fig.~\ref{fig:no of feat and fc} shows that nearly 30 features across all datasets select either 10.0 or 15.0 Hz as their $f_{cut}$.

\subsubsection{Impact of post processing}\label{append-subsubsec:impact-post-process}

We present additional results on the impact of post-processing by evaluating the MAE of the low-frequency component $X_l^c$, as well as the MSE and MAE of the high-frequency component $X_h^c$, averaged across multiple time segments. Table~\ref{table:mae-high-post-processing} reports the MAE distribution of $X_l^c$ before and after post-processing. While post-processing improved MSE, we observe a negative gain in MAE of approximately -5.5\%; however, this still corresponds to an positive reduction of 0.005 on average across all the scenarios in Table~\ref{table:mae-high-post-processing}. As discussed in Appendix~\ref{append-subsec:uni analysis}, this lower gain in MAE is due to small but frequent absolute differences between predictions and ground truth values, which MAE is sensitive to. Notably, MLP-based post-processing reduces the standard deviation of MAE by 74\%, enhancing the robustness of the model.

\begin{table}[htbp]
\centering
\caption{Comparison of model performance before and after the proposed method for $X^c_l$. Mean and standard deviation (sd) \textbf{MAE} values are shown. Values are rounded to two decimal positions due to the space limitation.}
\renewcommand{\arraystretch}{1.2}
\setlength{\tabcolsep}{5pt}
\label{table:mae-high-post-processing}
\scriptsize
\begin{tabular}{|l|c|c|c|c|c|c|c|c|}
\hline
\multirow{2}{*}{\textbf{Dataset}} & \multicolumn{2}{c|}{\textbf{Llama-3B}} & \multicolumn{2}{c|}{\textbf{Llama2-7B}} & \multicolumn{2}{c|}{\textbf{Deepseek-7B}} & \multicolumn{2}{c|}{\textbf{GPT-4o-mini}} \\
\cline{2-9}
 & Before & After & Before & After & Before & After & Before & After \\
\hline
ETTh1 & 0.32 (0.21) & 0.27 (0.07) & 0.37 (0.27) & 0.30 (0.08) & 0.32 (0.28) & 0.28 (0.07) & 0.29 (0.19) & 0.30 (0.10) \\
\hline
ETTh2 & 0.20 (0.17) & 0.17 (0.04) & 0.21 (0.19) & 0.20 (0.02) & 0.19 (0.19) & 0.21 (0.04) & 0.21 (0.23) & 0.28 (0.07) \\
\hline
ETTm1 & 0.32 (0.24) & 0.27 (0.08) & 0.60 (1.00) & 0.32 (0.09) & 0.30 (0.23) & 0.31 (0.08) & 0.33 (0.24) & 0.28 (0.08) \\
\hline
ETTm2 & 0.17 (0.14) & 0.18 (0.02) & 0.18 (0.15) & 0.23 (0.03) & 0.18 (0.15) & 0.20 (0.02) & 0.22 (0.17) & 0.32 (0.05) \\
\hline
Electricity & 0.18 (0.13) & 0.15 (0.01) & 0.20 (0.16) & 0.17 (0.03) & 0.17 (0.11) & 0.16 (0.01) & 0.24 (0.26) & 0.17 (0.01) \\
\hline
Weather & 0.15 (0.16) & 0.24 (0.09) & 0.18 (0.21) & 0.25 (0.06) & 0.14 (0.14) & 0.26 (0.09) & 0.29 (0.35) & 0.29 (0.09) \\
\hline
\end{tabular}
\end{table}
              
Tables~\ref{table:mse-multi-high} and~\ref{table:mae-multi-high} present the MSE and MAE comparisons of the high-frequency component ($X_h^c$) before and after applying the Gaussian value conversion post-processing step. In terms of raw metrics, we observe an overall negative gain of $-11.0\%$ in MSE and $-3.9\%$ in MAE. However, the post-processing step generally improves the stability of results across most scenarios, with an average MSE gain of $11.9\%$ (excluding two anomalous cases: Llama-3B on ETTh2 and Deepseek-7B on Electricity, which show large negative gains of $-611.9\%$ and $-233.9\%$, respectively) and an average MAE gain of $1.5\%$.
As discussed in $\S$~\ref{subsubsec:impact post-processing}, due to the inherently random and noisy nature of $X_h^c$, we rely on distribution matching via the Kolmogorov--Smirnov (KS) statistic rather than direct feature-wise comparison across time steps. Therefore, conventional metrics like MSE and MAE may not fully capture the improvements brought by the KS-statistic-based value transformation, as also observed in $\S$~\ref{subsubsec:impact post-processing}.

\begin{table}[htbp]
\centering
\caption{MSE for high frequency component. Comparison of model performance before and after the proposed method. Mean and standard deviation (sd) values are shown.}
\renewcommand{\arraystretch}{1.2}
\setlength{\tabcolsep}{5pt}
\label{table:mse-multi-high}
\scriptsize
\begin{tabular}{|l|c|c|c|c|c|c|c|c|}
\hline
\multirow{2}{*}{\textbf{Dataset}} & \multicolumn{2}{c|}{\textbf{Llama-3B}} & \multicolumn{2}{c|}{\textbf{Llama2-7B}} & \multicolumn{2}{c|}{\textbf{Deepseek-7B}} & \multicolumn{2}{c|}{\textbf{GPT-4o-mini}} \\
\cline{2-9}
 & Before & After & Before & After & Before & After & Before & After \\
\hline
ETTh1 & 0.22 (0.06) & 0.22 (0.01) & 0.21 (0.06) & 0.22 (0.01) & 0.22 (0.05) & 0.23 (0.01) & 0.21 (0.02) & 0.22 (0.02) \\
\hline
ETTh2 & 0.16 (0.00) & 0.21 (0.01) & 0.17 (0.01) & 0.21 (0.01) & 0.17 (0.01) & 0.21 (0.01) & 0.21 (0.00) & 0.21 (0.00) \\
\hline
ETTm1 & 0.16 (0.04) & 0.19 (0.02) & 0.16 (0.01) & 0.19 (0.02) & 0.18 (0.02) & 0.19 (0.02) & 0.18 (0.02) & 0.18 (0.02) \\
\hline
ETTm2 & 0.18 (0.01) & 0.22 (0.01) & 0.21 (0.02) & 0.23 (0.01) & 0.21 (0.03) & 0.22 (0.01) & 0.21 (0.02) & 0.23 (0.02) \\
\hline
Electricity & 0.22 (0.02) & 0.19 (0.05) & 0.18 (0.03) & 0.20 (0.05) & 0.20 (0.02) & 0.20 (0.05) & 0.20 (0.04) & 0.21 (0.06) \\
\hline
Weather & 0.16 (0.02) & 0.18 (0.00) & 0.13 (0.02) & 0.18 (0.00) & 0.16 (0.04) & 0.18 (0.00) & 0.15 (0.02) & 0.17 (0.00) \\
\hline
\end{tabular}
\end{table}

\begin{table}[htbp]
\centering
\caption{MAE for high frequency distribution. Comparison of model performance before and after the proposed method. Mean and standard deviation (sd) values are shown. }
\renewcommand{\arraystretch}{1.2}
\setlength{\tabcolsep}{5pt}
\label{table:mae-multi-high}
\scriptsize
\begin{tabular}{|l|c|c|c|c|c|c|c|c|}
\hline
\multirow{2}{*}{\textbf{Dataset}} & \multicolumn{2}{c|}{\textbf{Llama-3B}} & \multicolumn{2}{c|}{\textbf{Llama2-7B}} & \multicolumn{2}{c|}{\textbf{Deepseek-7B}} & \multicolumn{2}{c|}{\textbf{GPT-4o-mini}} \\
\cline{2-9}
 & Before & After & Before & After & Before & After & Before & After \\
\hline
ETTh1 & 0.36 (0.05) & 0.36 (0.01) & 0.36 (0.05) & 0.36 (0.01) & 0.37 (0.04) & 0.37 (0.01) & 0.36 (0.02) & 0.36 (0.02) \\
\hline
ETTh2 & 0.30 (0.00) & 0.35 (0.00) & 0.32 (0.01) & 0.36 (0.01) & 0.33 (0.01) & 0.35 (0.00) & 0.36 (0.01) & 0.36 (0.01) \\
\hline
ETTm1 & 0.30 (0.04) & 0.33 (0.02) & 0.30 (0.02) & 0.33 (0.02) & 0.31 (0.02) & 0.33 (0.02) & 0.32 (0.02) & 0.31 (0.03) \\
\hline
ETTm2 & 0.33 (0.01) & 0.36 (0.01) & 0.36 (0.01) & 0.37 (0.02) & 0.36 (0.03) & 0.36 (0.02) & 0.36 (0.03) & 0.37 (0.03) \\
\hline
Electricity & 0.36 (0.03) & 0.33 (0.05) & 0.33 (0.04) & 0.34 (0.07) & 0.35 (0.02) & 0.33 (0.06) & 0.33 (0.05) & 0.33 (0.07) \\
\hline
Weather & 0.29 (0.02) & 0.31 (0.01) & 0.28 (0.03) & 0.32 (0.00) & 0.30 (0.04) & 0.31 (0.00) & 0.29 (0.02) & 0.31 (0.01) \\
\hline
\end{tabular}
\end{table}

\subsection{Further analysis on multivariate prediction}
                
\subsubsection{Sample outputs of multivariate prediction}\label{append-subsec:sample-output-prediction-len}

\begin{figure}[h!]
    \centering
    \begin{subfigure}[b]{0.7\textwidth}
        \centering
        \includegraphics[width=\textwidth]{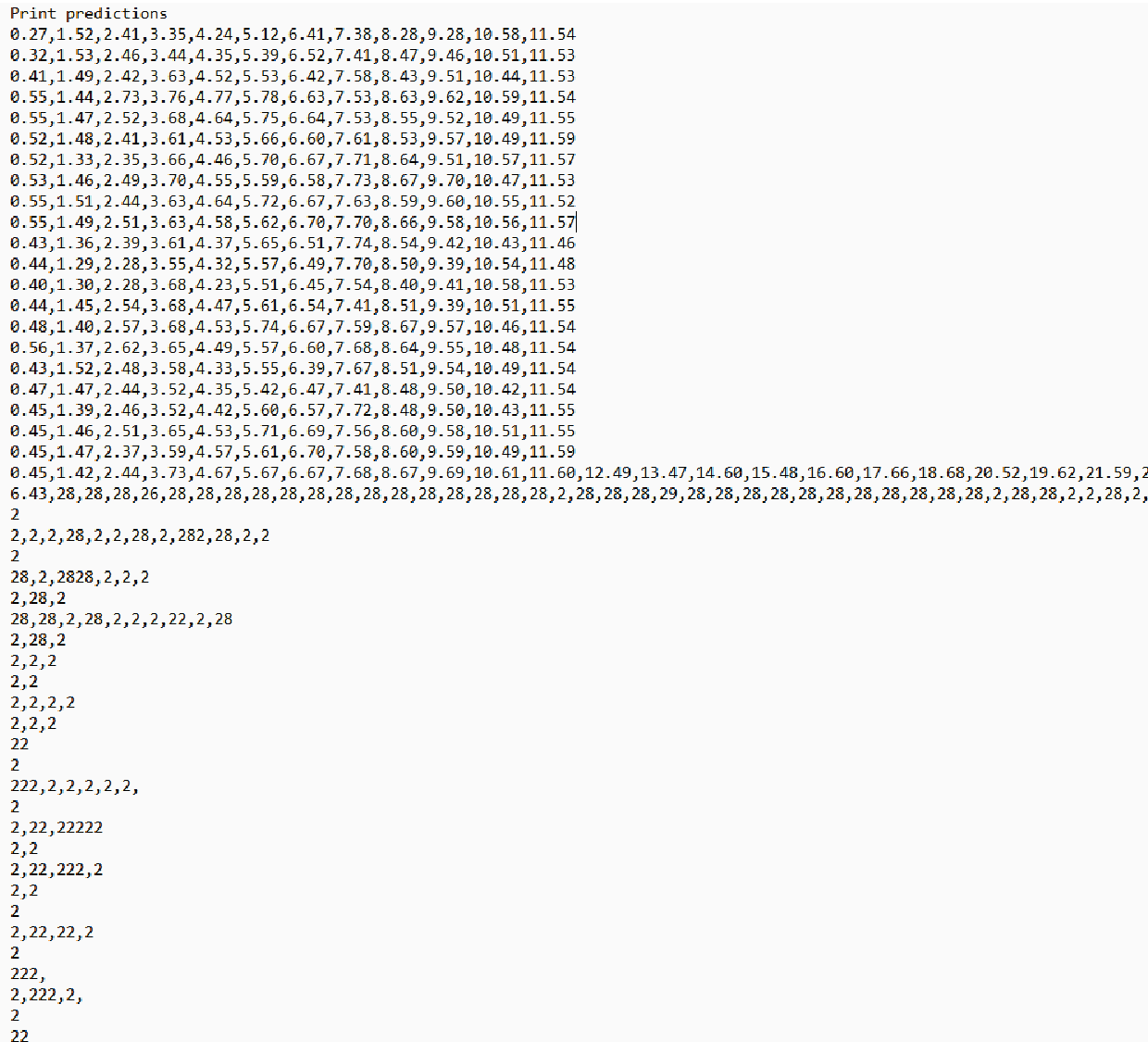}
    \end{subfigure}
    \caption{Deepseek 7B 48 prediction length with 12 features}
    \label{fig:elec-deepseek-48-12}
\end{figure}

\begin{figure}[h!]
    \centering
    \begin{subfigure}[b]{0.75\textwidth}
        \centering
        \includegraphics[width=\textwidth]{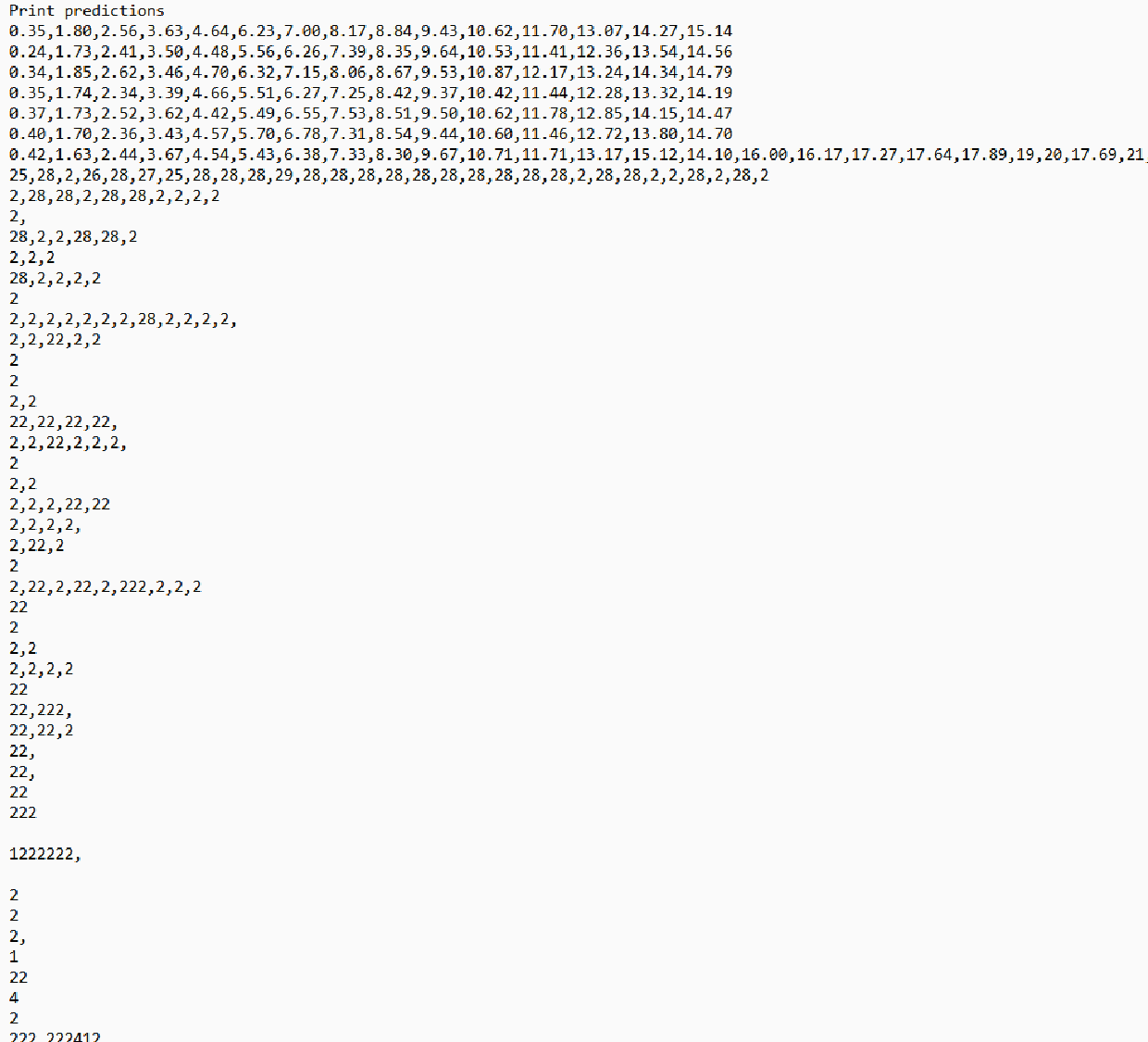}
    \end{subfigure}
    \caption{Deepseek 7B 48 prediction length with 15 features}
    \label{fig:elec-deepseek-48-15}
\end{figure}

\begin{figure}[h!]
    \centering
    \begin{subfigure}[b]{0.75\textwidth}
        \centering
        \includegraphics[width=\textwidth]{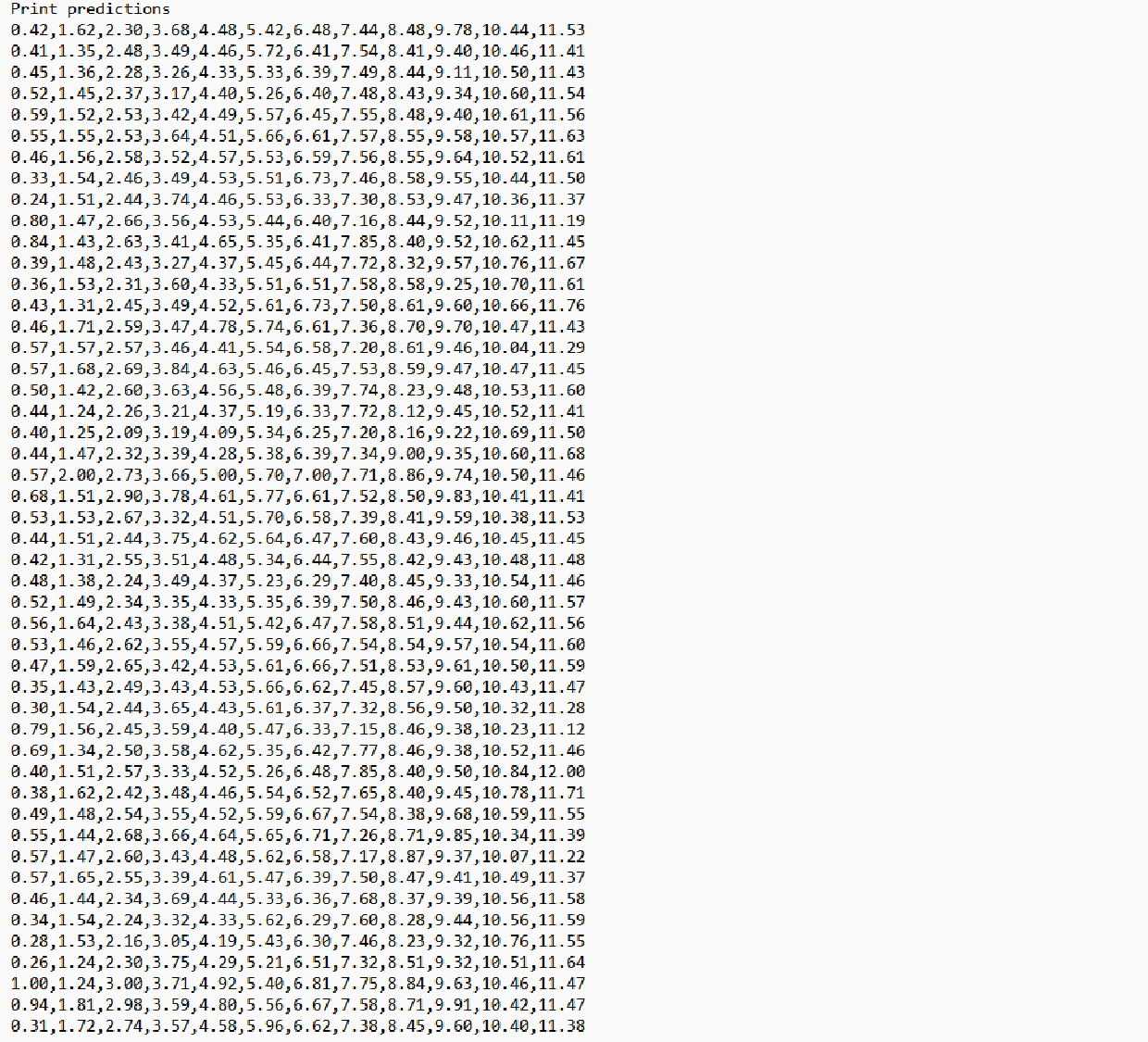}
    \end{subfigure}
    \caption{GPT-4o-mini 48 prediction length with 12 features}
    \label{fig:elec-got-4o-mini-48-12}
\end{figure}

\begin{figure}[t]
    \centering
    \begin{subfigure}[b]{0.70\textwidth}
        \centering
        \includegraphics[width=\textwidth]{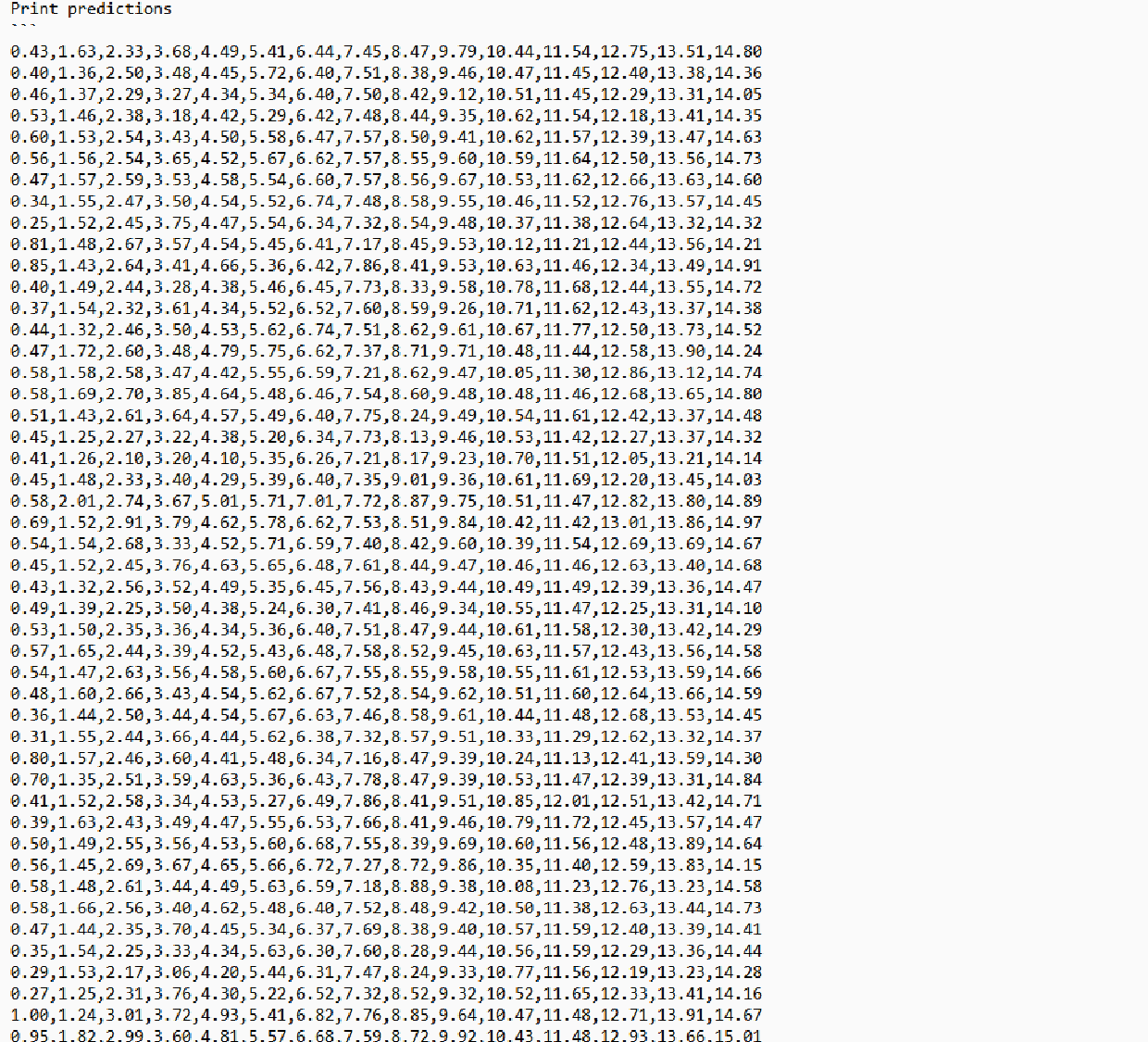}
    \end{subfigure}
    \caption{GPT-4o-mini 48 prediction length with 15 features}
    \label{fig:elec-gpt-4o-mini-48-15}
\end{figure}

Fig.~\ref{fig:elec-deepseek-48-12}, Fig.~\ref{fig:elec-deepseek-48-15}, Fig.~\ref{fig:elec-got-4o-mini-48-12}, and Fig.~\ref{fig:elec-gpt-4o-mini-48-15} present sample outputs for predictions by Deepseek 7B and GPT-4o-mini at a prediction length of 48, using 12 and 15 input features. As discussed in $\S~\ref{subsec:critical-eval}$, when the number of features exceeds 9, Deepseek 7B (Fig.~\ref{fig:elec-deepseek-48-12} and Fig.~\ref{fig:elec-deepseek-48-15}) produces noisy outputs with repetitive patterns or numerical values. A similar trend is observed for Llama 7B. In contrast, GPT-4o-mini (Fig.~\ref{fig:elec-got-4o-mini-48-12} and Fig.~\ref{fig:elec-gpt-4o-mini-48-15}) generates predictions as expected, indicating that it maintains sufficient context length to handle a higher number of features.


\subsection{Sample traces}

\subsubsection{Predictions for different datasets}
\begin{figure}[h!]
    \centering
    \begin{subfigure}[b]{0.24\textwidth}
        \centering
        \includegraphics[width=\textwidth]{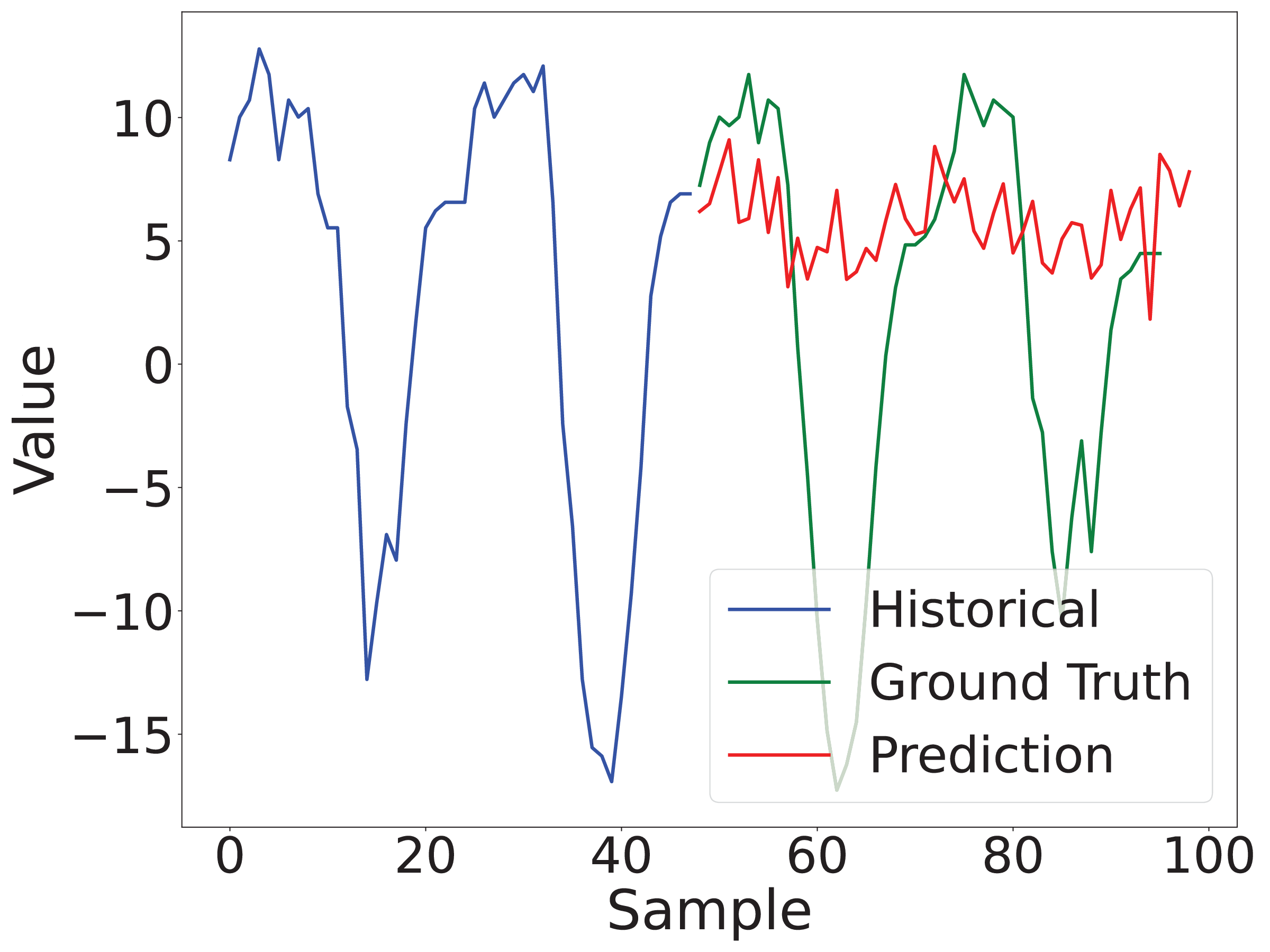}
        \caption{ETTh1-48}
        \label{fig:}
    \end{subfigure}
    \begin{subfigure}[b]{0.24\textwidth}
        \centering
        \includegraphics[width=\textwidth]{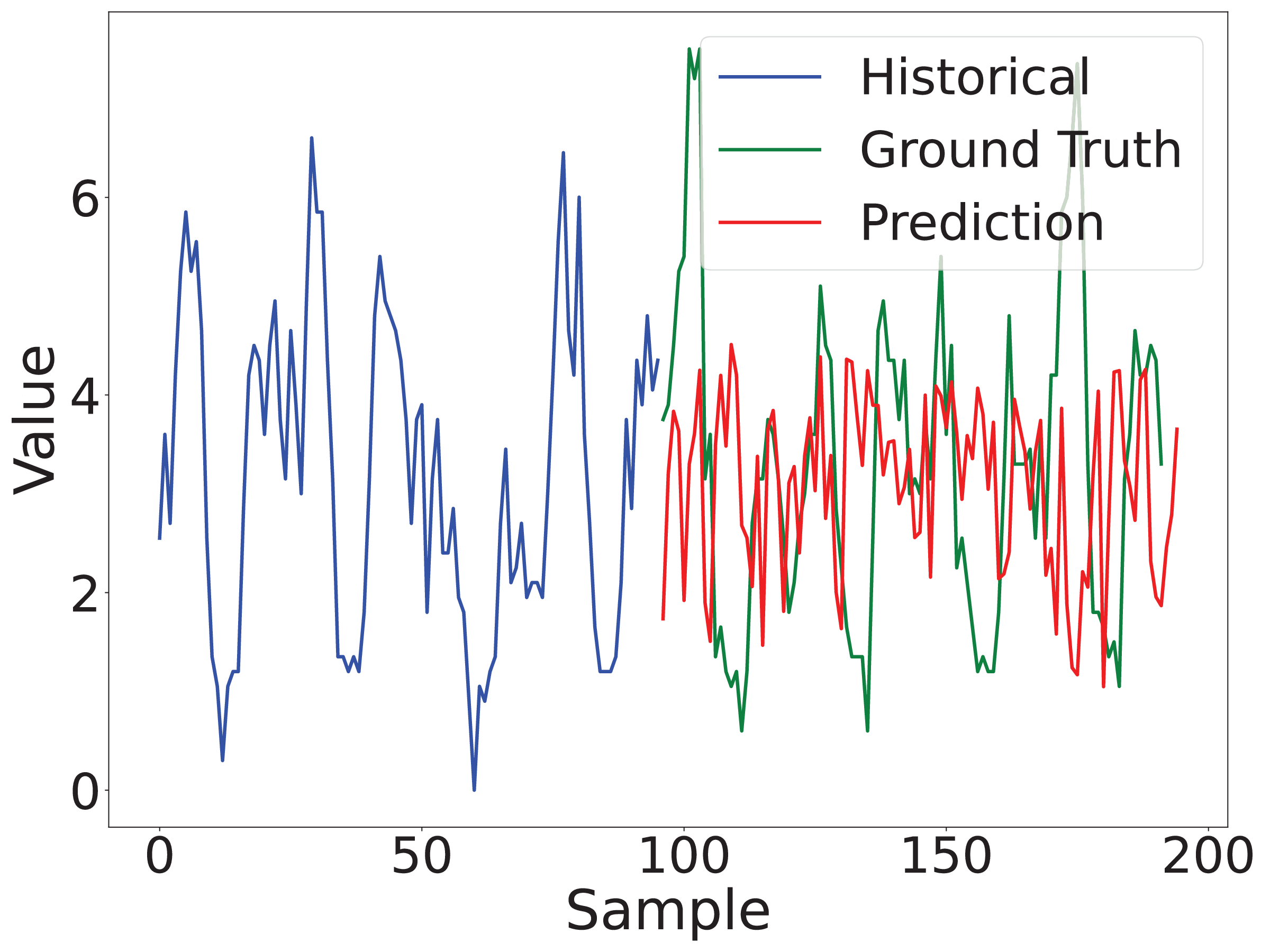}
        \caption{ETTh1-96}
        \label{fig:}
    \end{subfigure}
    \begin{subfigure}[b]{0.24\textwidth}
        \centering
        \includegraphics[width=\textwidth]{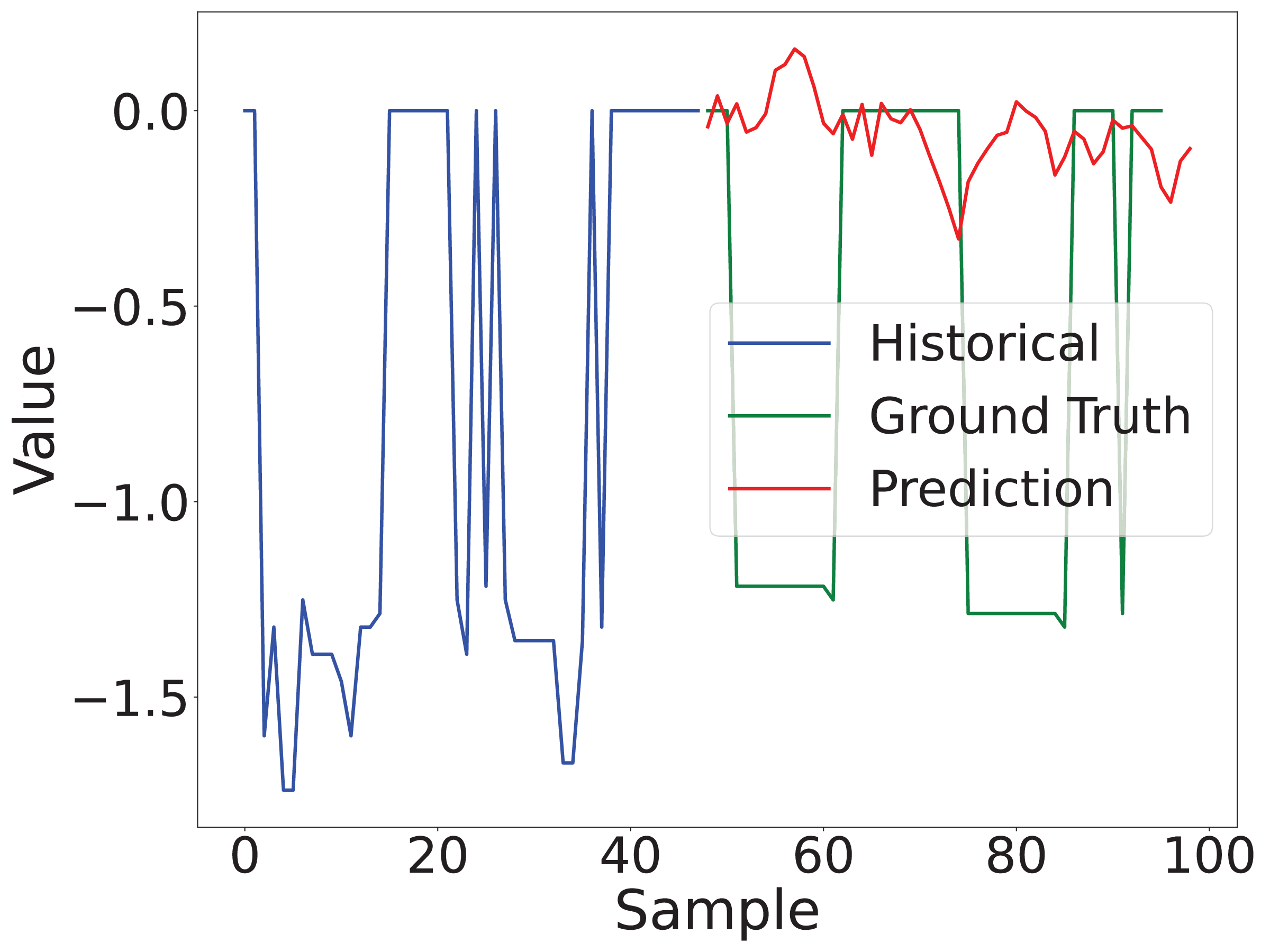}
        \caption{ETTh2-48}
        \label{fig:}
    \end{subfigure}
    \begin{subfigure}[b]{0.24\textwidth}
        \centering
        \includegraphics[width=\textwidth]{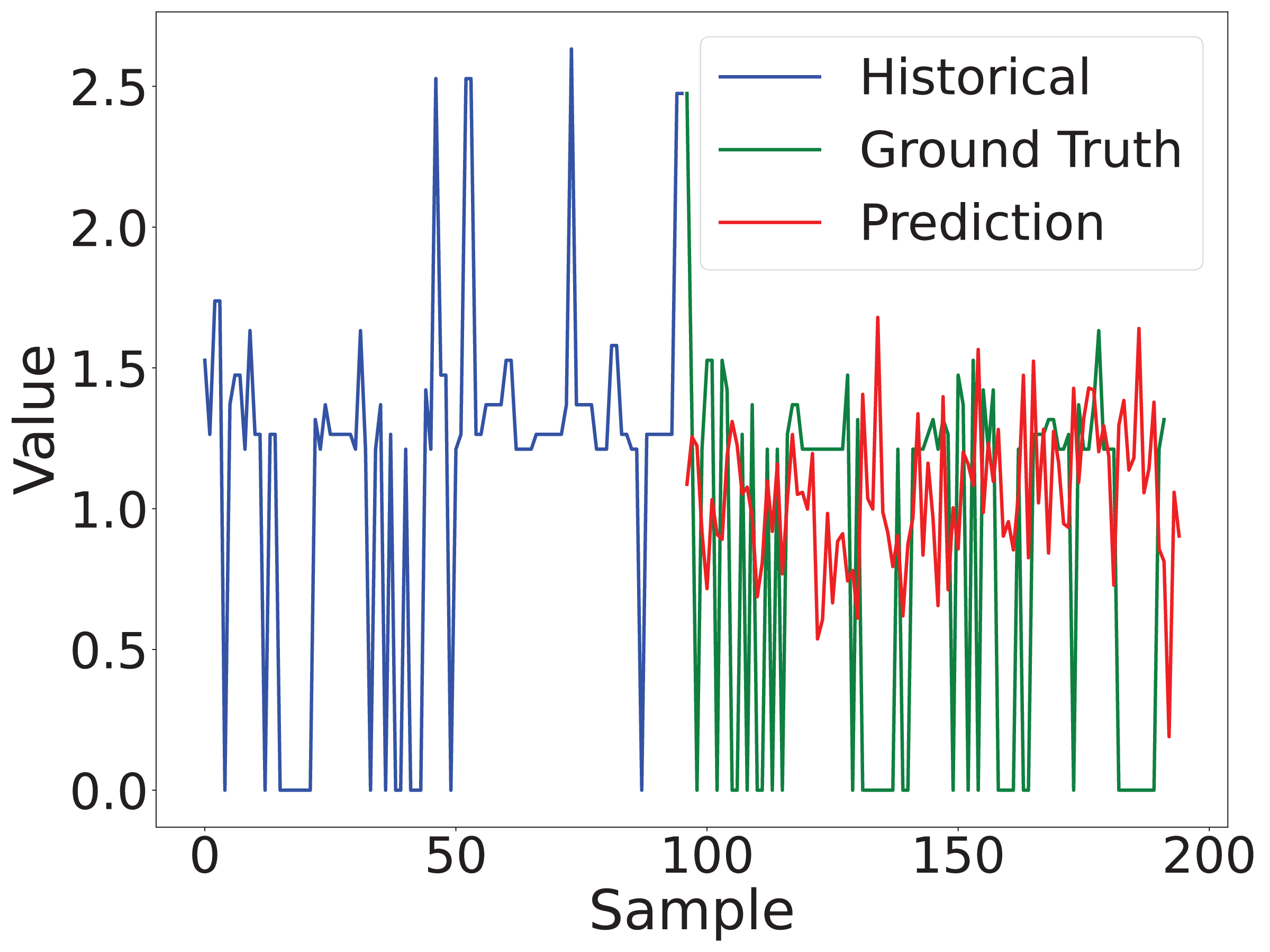}
        \caption{ETTh2-96}
        \label{fig:}
    \end{subfigure}
    \begin{subfigure}[b]{0.24\textwidth}
        \centering
        \includegraphics[width=\textwidth]{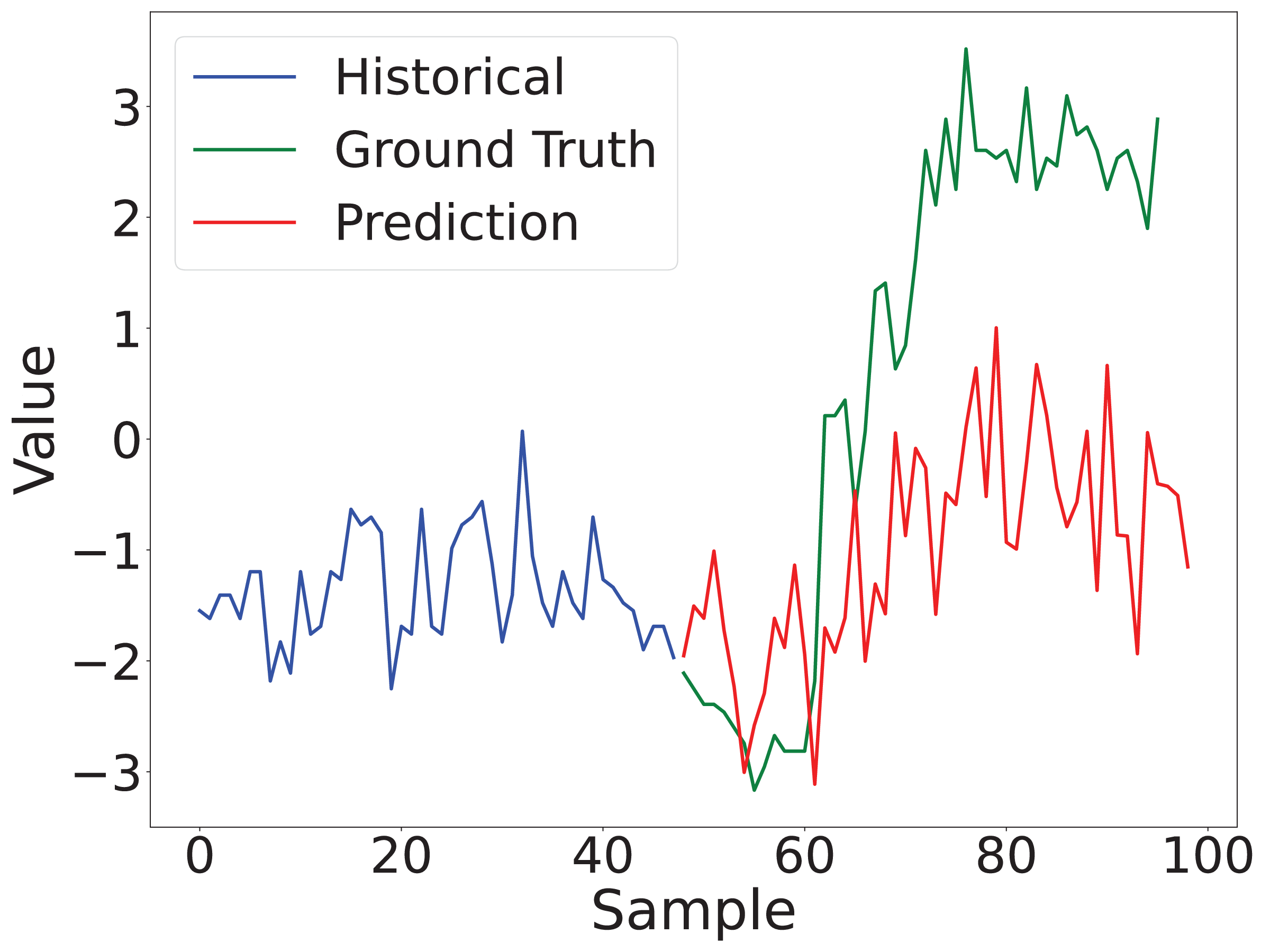}
        \caption{ETTm1-48}
        \label{fig:}
    \end{subfigure}
    \begin{subfigure}[b]{0.24\textwidth}
        \centering
        \includegraphics[width=\textwidth]{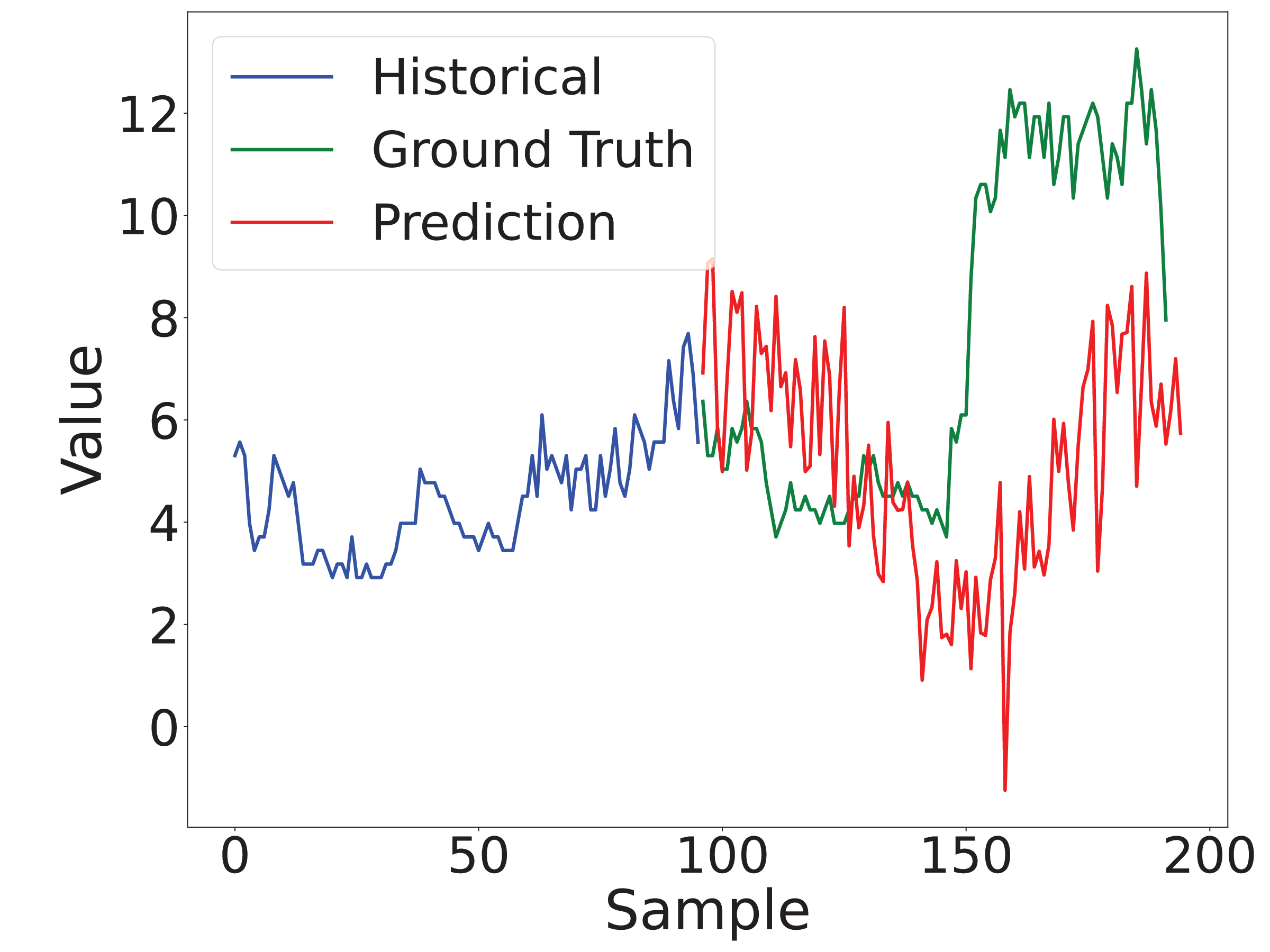}
        \caption{ETTm1-96}
        \label{fig:}
    \end{subfigure}
    \begin{subfigure}[b]{0.24\textwidth}
        \centering
        \includegraphics[width=\textwidth]{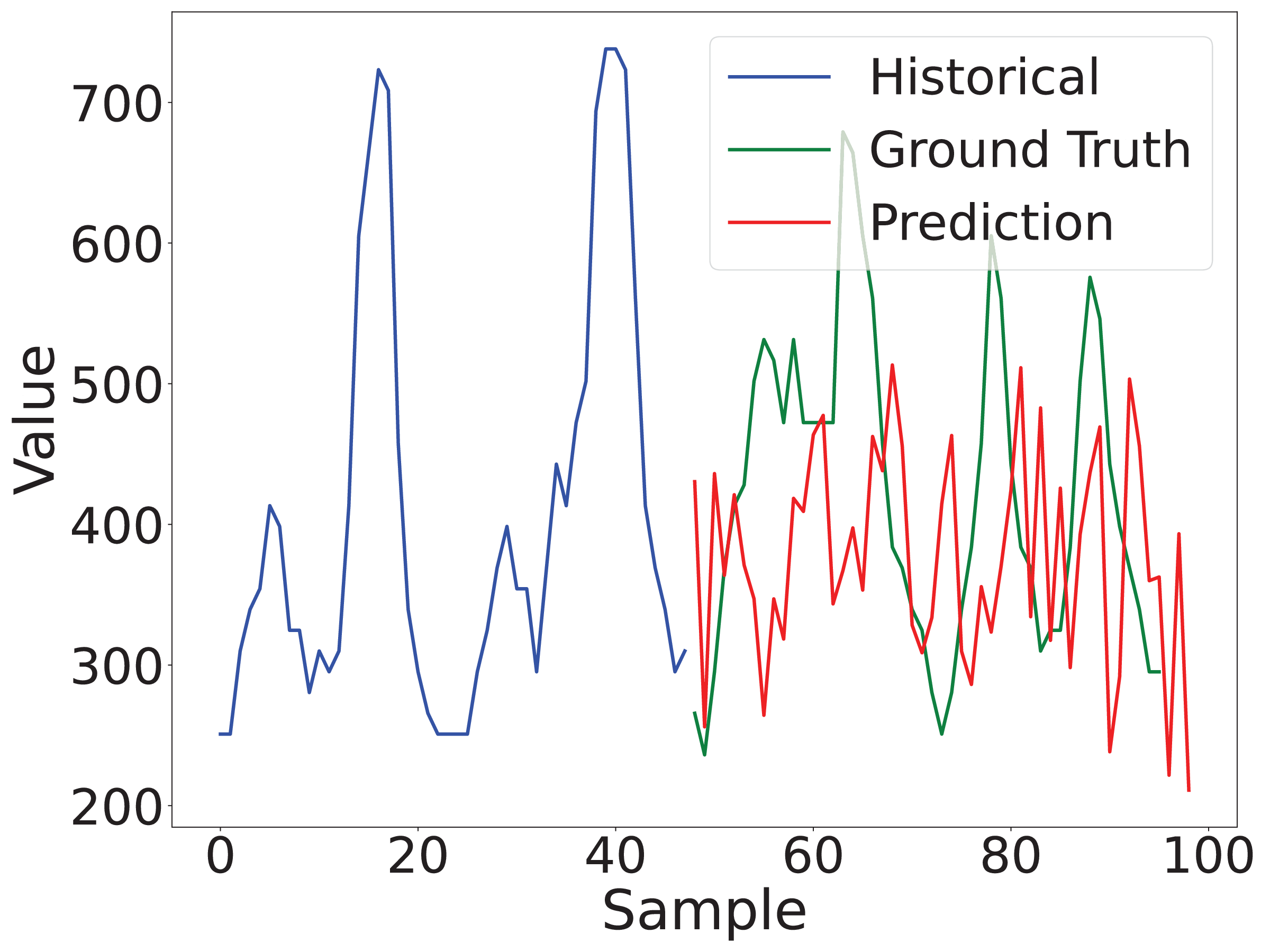}
        \caption{electricity-48}
        \label{fig:}
    \end{subfigure}
    \begin{subfigure}[b]{0.24\textwidth}
        \centering
        \includegraphics[width=\textwidth]{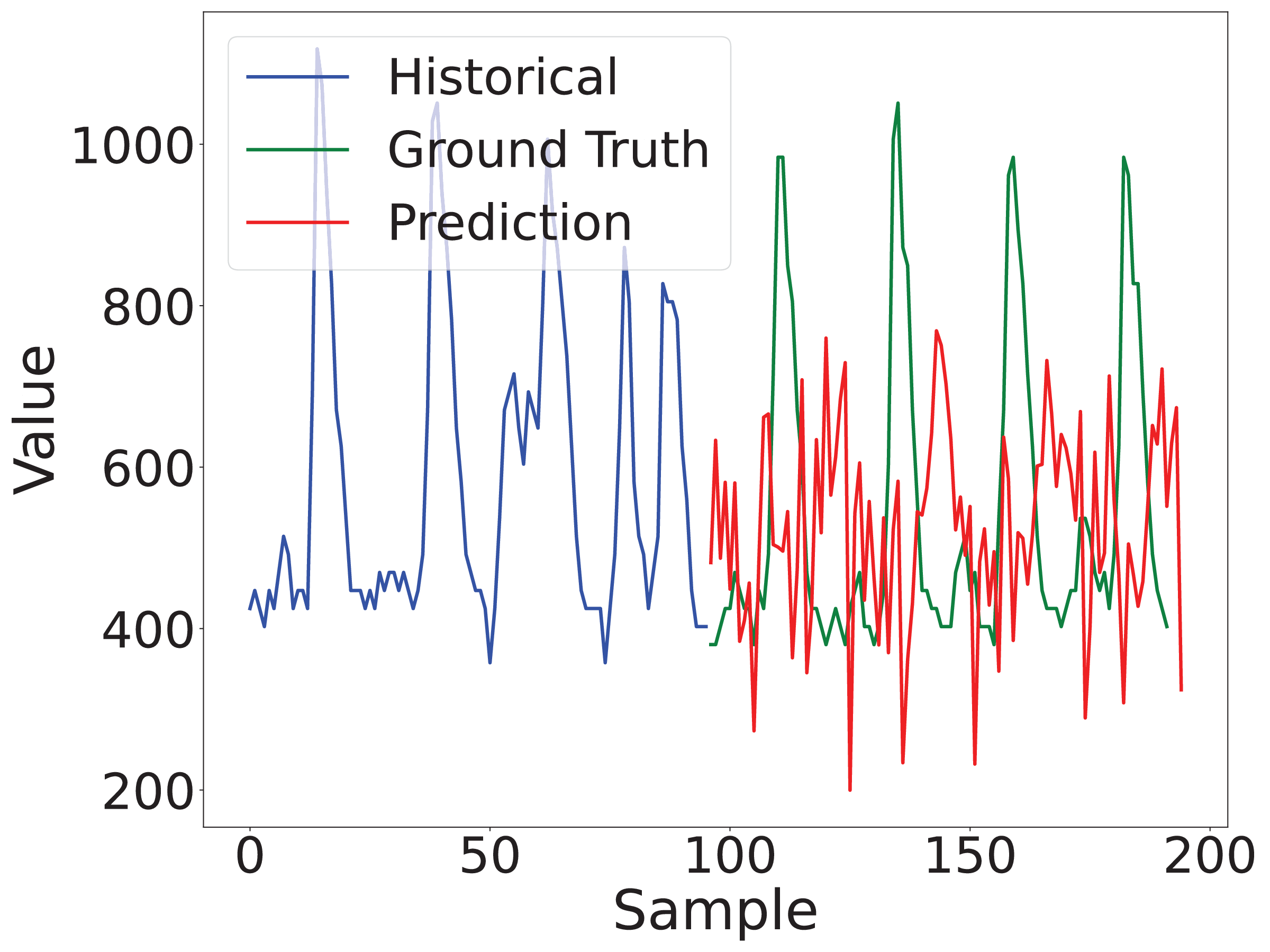}
        \caption{electricity-96}
        \label{fig:}
    \end{subfigure}
    \caption{Sample traces of \solution{} prediction using GPT-4o-mini model in univariate generation}
    \label{fig:uni-gpt-4o-mini}
\end{figure}

\begin{figure}[h!]
    \centering
    \begin{subfigure}[b]{0.24\textwidth}
        \centering
        \includegraphics[width=\textwidth]{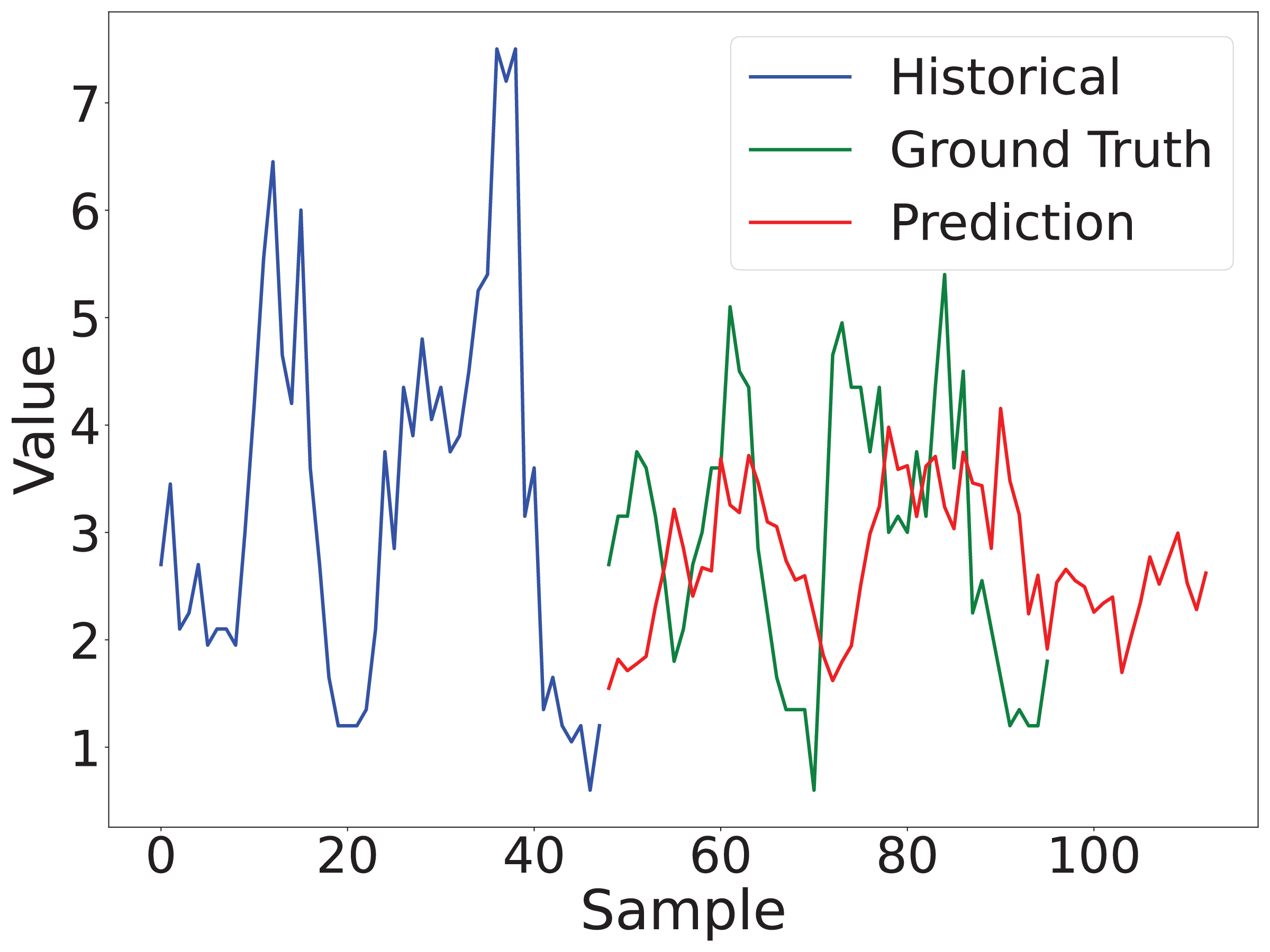}
        \caption{ETTh1-48}
        \label{fig:}
    \end{subfigure}
    \begin{subfigure}[b]{0.24\textwidth}
        \centering
        \includegraphics[width=\textwidth]{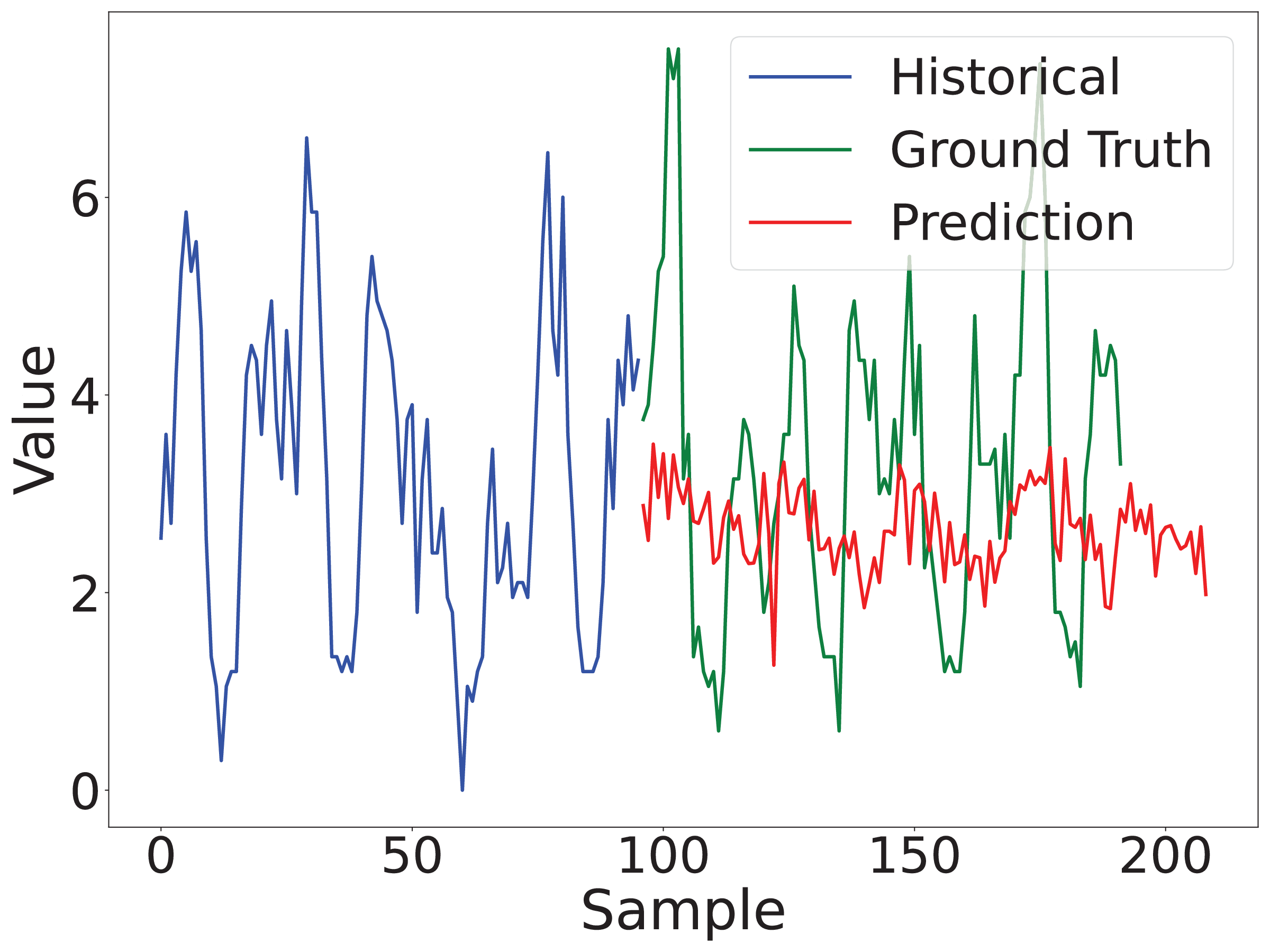}
        \caption{ETTh1-96}
        \label{fig:}
    \end{subfigure}
    \begin{subfigure}[b]{0.24\textwidth}
        \centering
        \includegraphics[width=\textwidth]{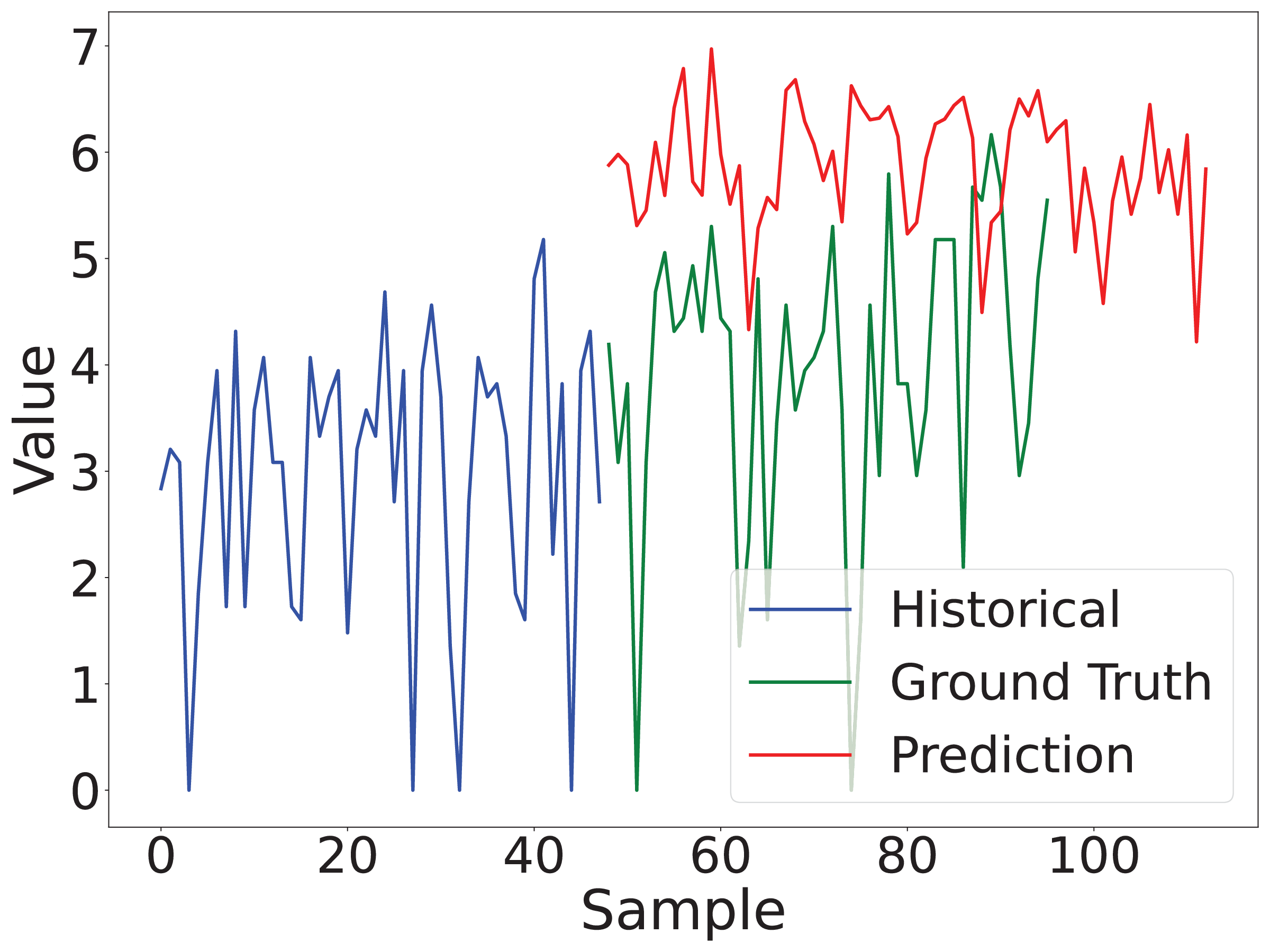}
        \caption{ETTh2-48}
        \label{fig:}
    \end{subfigure}
    \begin{subfigure}[b]{0.24\textwidth}
        \centering
        \includegraphics[width=\textwidth]{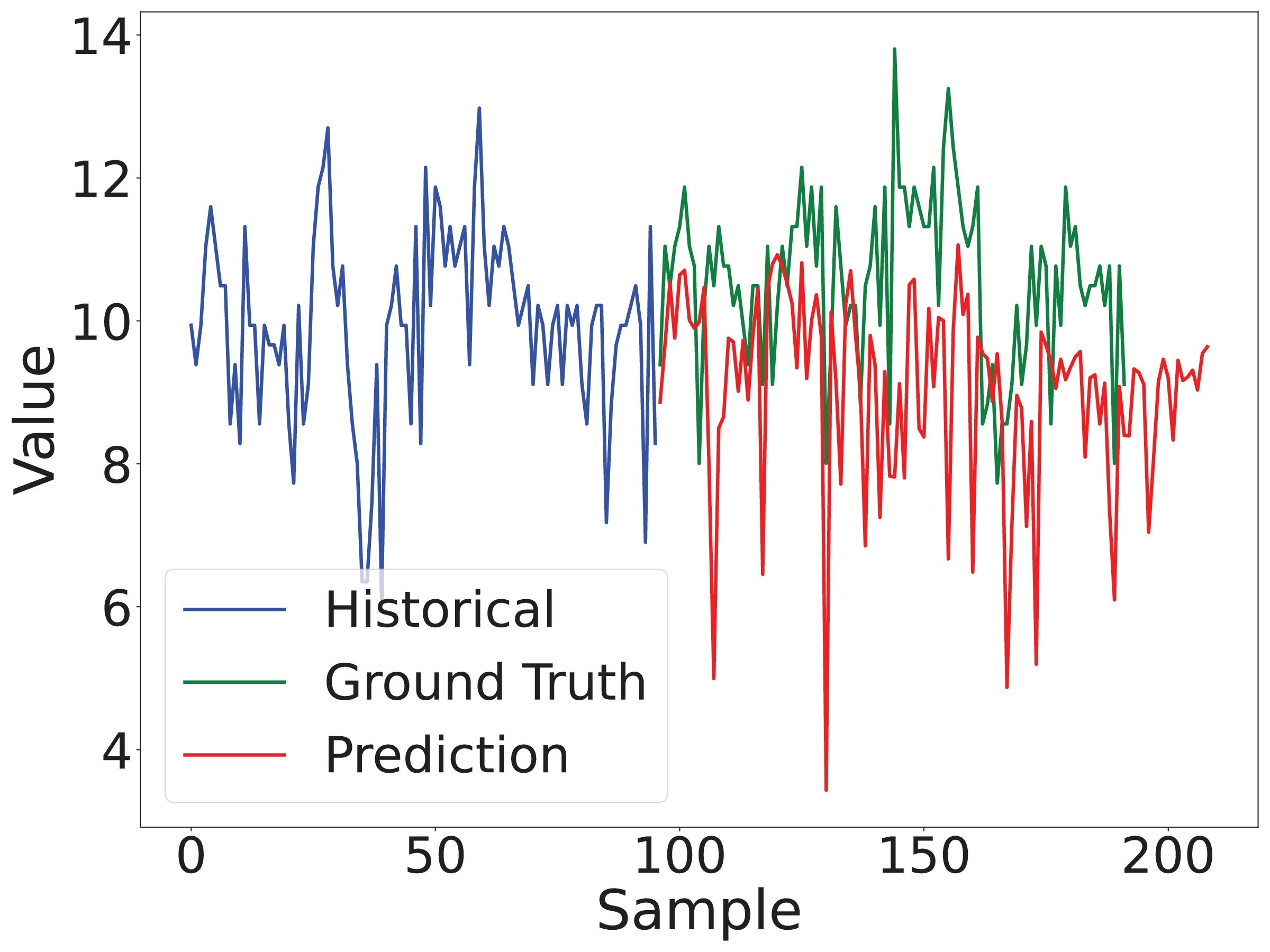}
        \caption{ETTh2-96}
        \label{fig:}
    \end{subfigure}
    \begin{subfigure}[b]{0.24\textwidth}
        \centering
        \includegraphics[width=\textwidth]{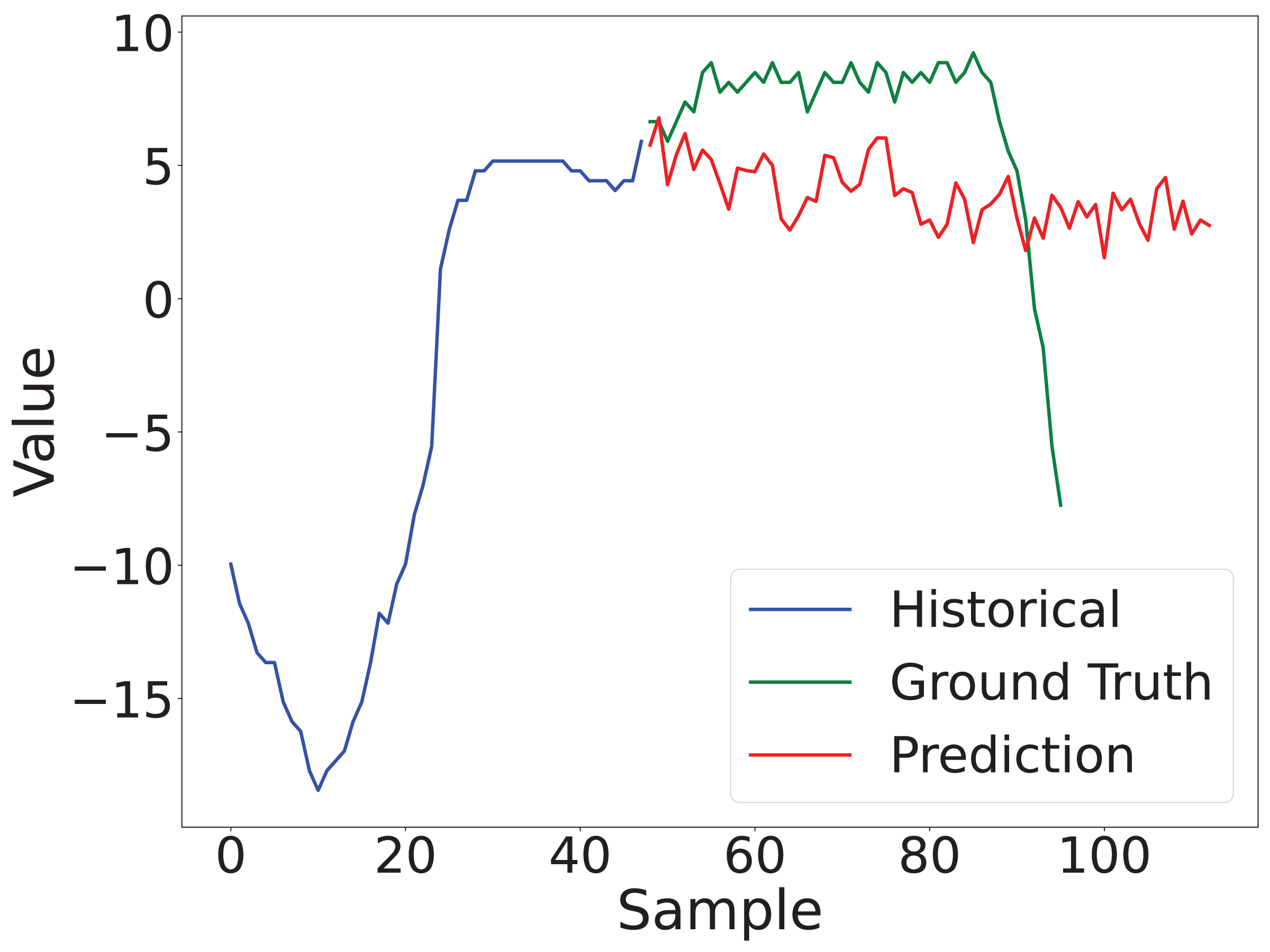}
        \caption{ETTm1-48}
        \label{fig:}
    \end{subfigure}
    \begin{subfigure}[b]{0.24\textwidth}
        \centering
        \includegraphics[width=\textwidth]{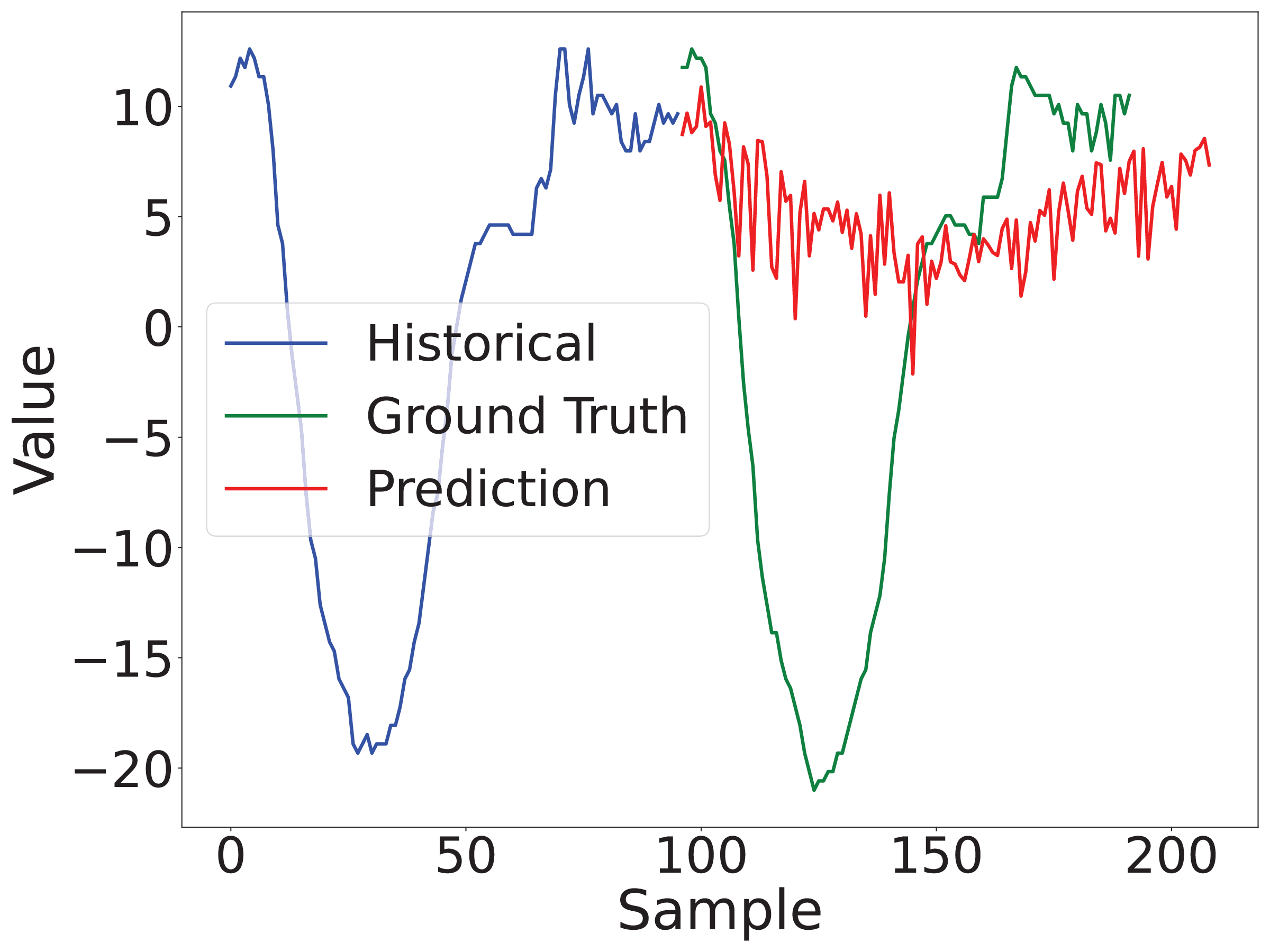}
        \caption{ETTm1-96}
        \label{fig:}
    \end{subfigure}
    \begin{subfigure}[b]{0.24\textwidth}
        \centering
        \includegraphics[width=\textwidth]{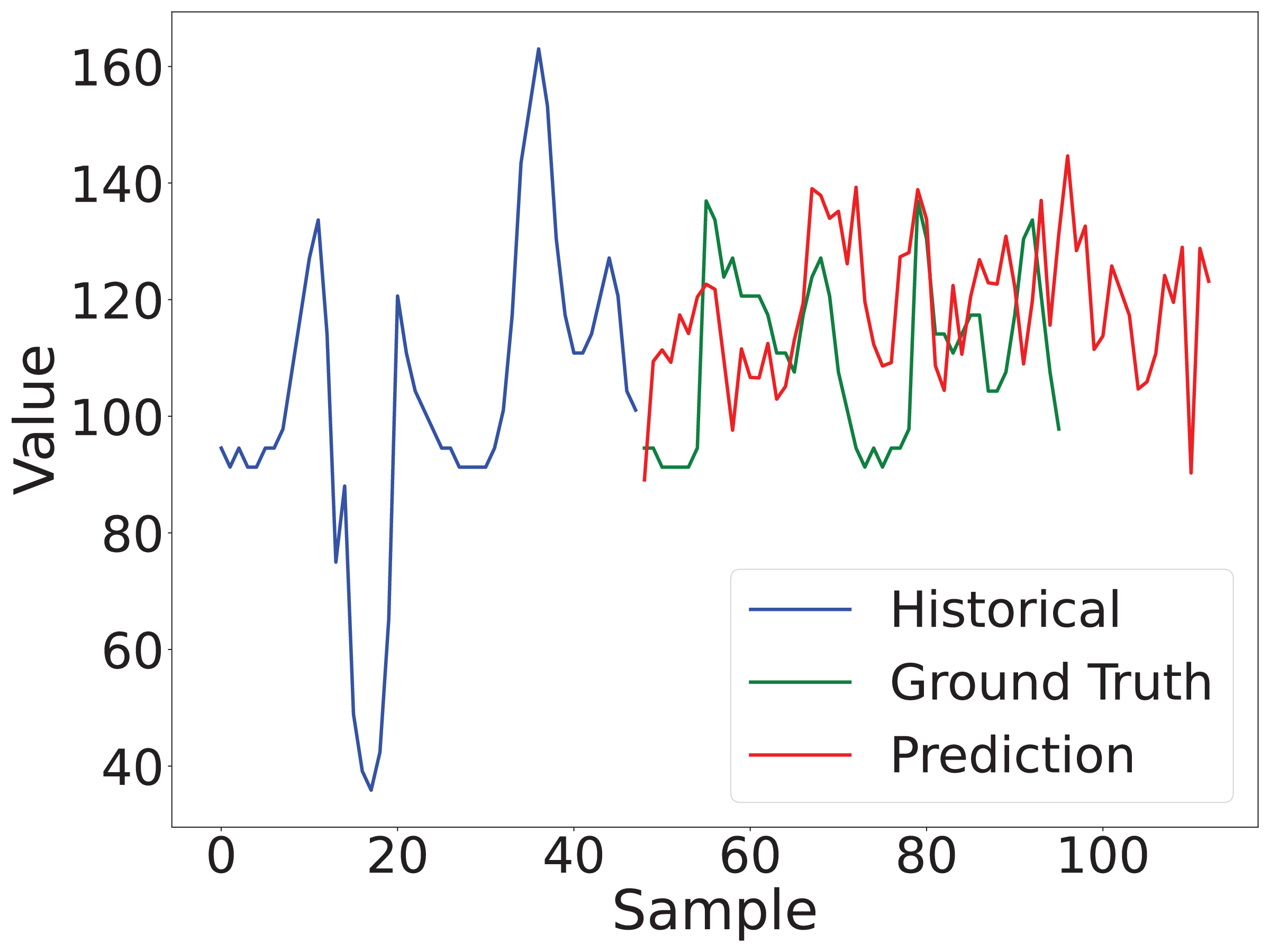}
        \caption{electricity-48}
        \label{fig:}
    \end{subfigure}
    \begin{subfigure}[b]{0.24\textwidth}
        \centering
        \includegraphics[width=\textwidth]{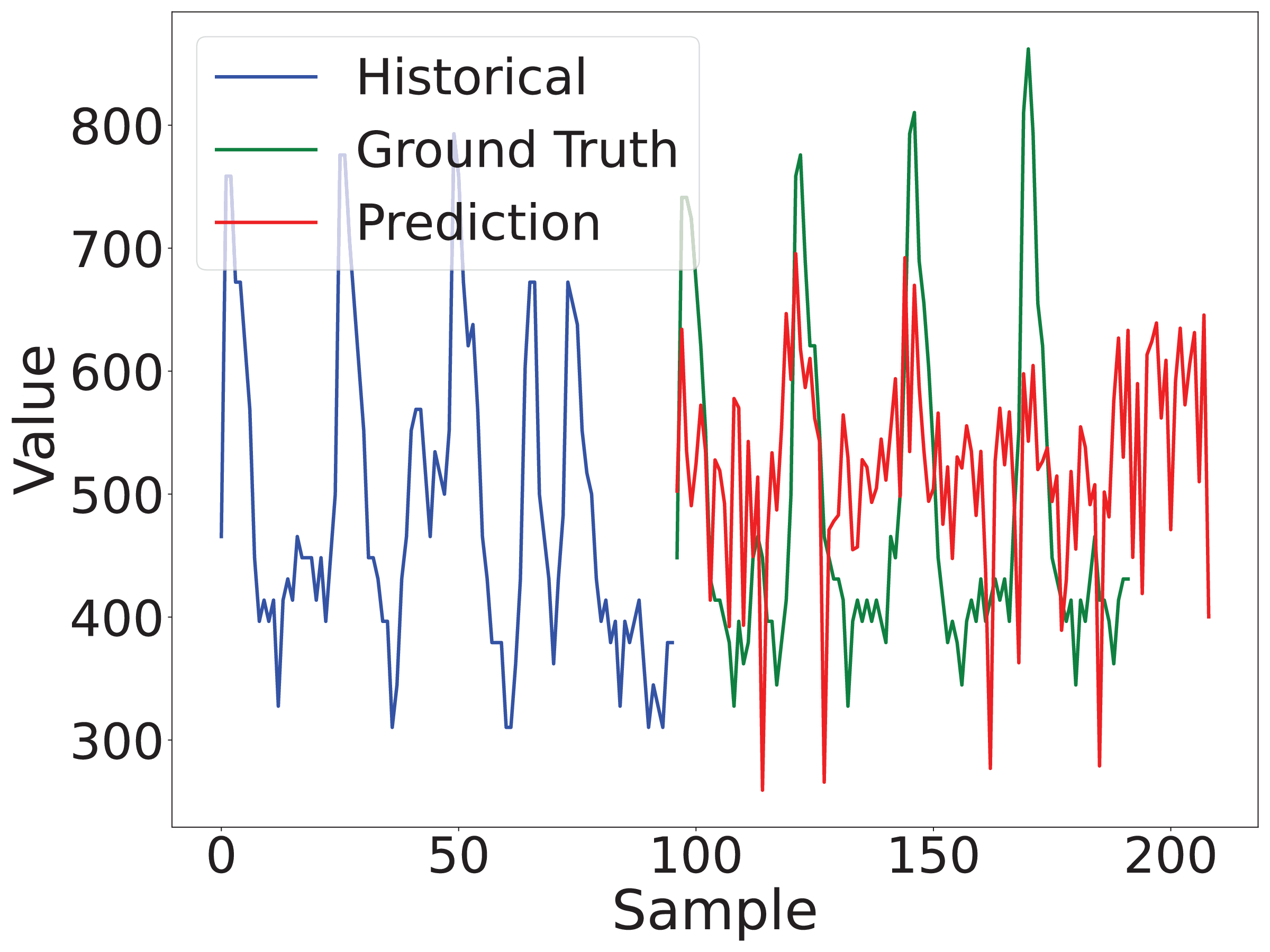}
        \caption{electricity-96}
        \label{fig:}
    \end{subfigure}
    \caption{Sample traces of \solution{} prediction using Llama-7B in univariate generation}
    \label{fig:uni-llama-7b}
\end{figure}

\begin{figure}[h!]
    \centering
    \begin{subfigure}[b]{0.24\textwidth}
        \centering
        \includegraphics[width=\textwidth]{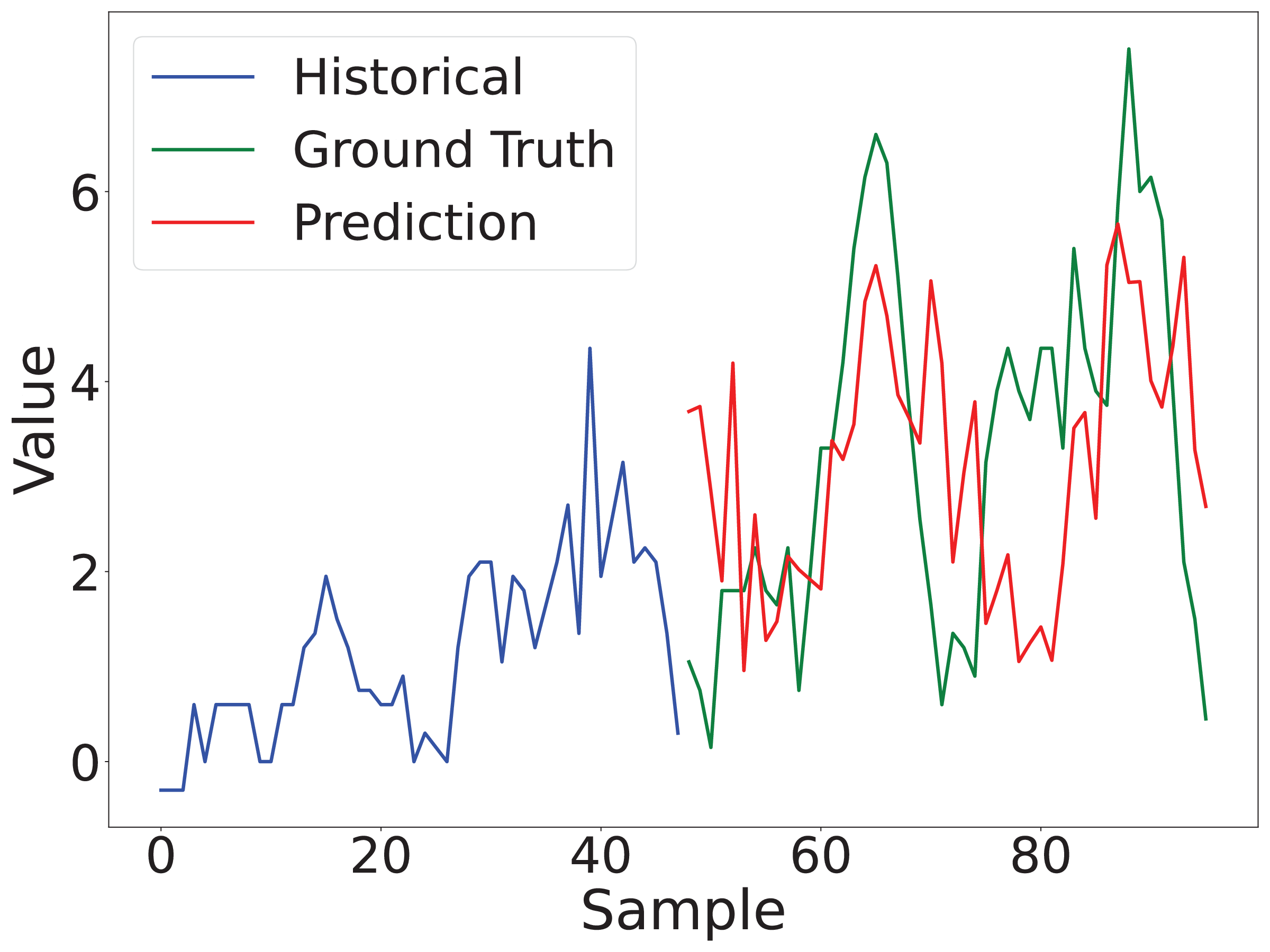}
        \caption{ETTh1-48}
        \label{fig:good impact high freq 1}
    \end{subfigure}
    \begin{subfigure}[b]{0.24\textwidth}
        \centering
        \includegraphics[width=\textwidth]{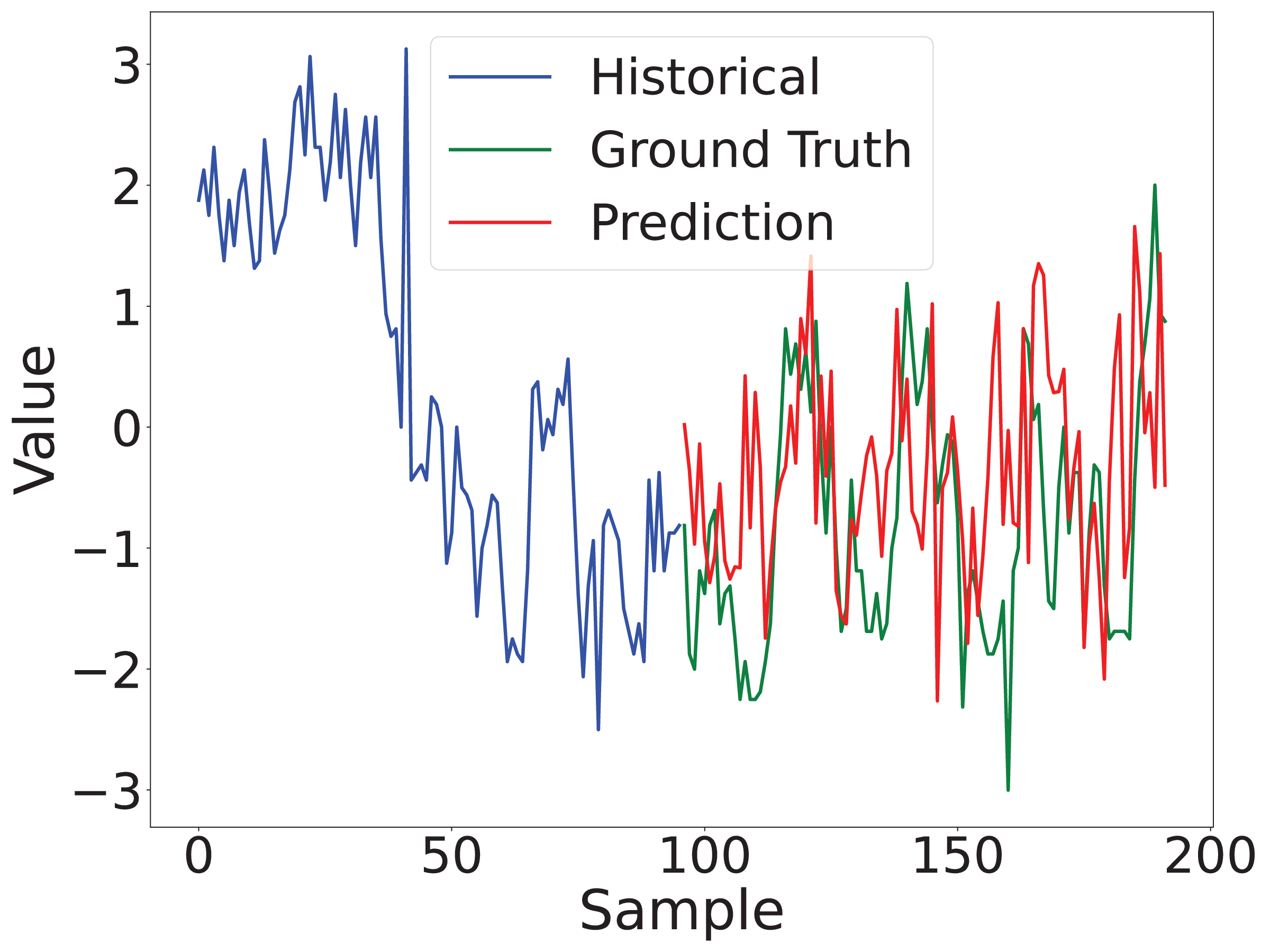}
        \caption{ETTh1-96}
        \label{fig:good impact high freq 2}
    \end{subfigure}
    \begin{subfigure}[b]{0.24\textwidth}
        \centering
        \includegraphics[width=\textwidth]{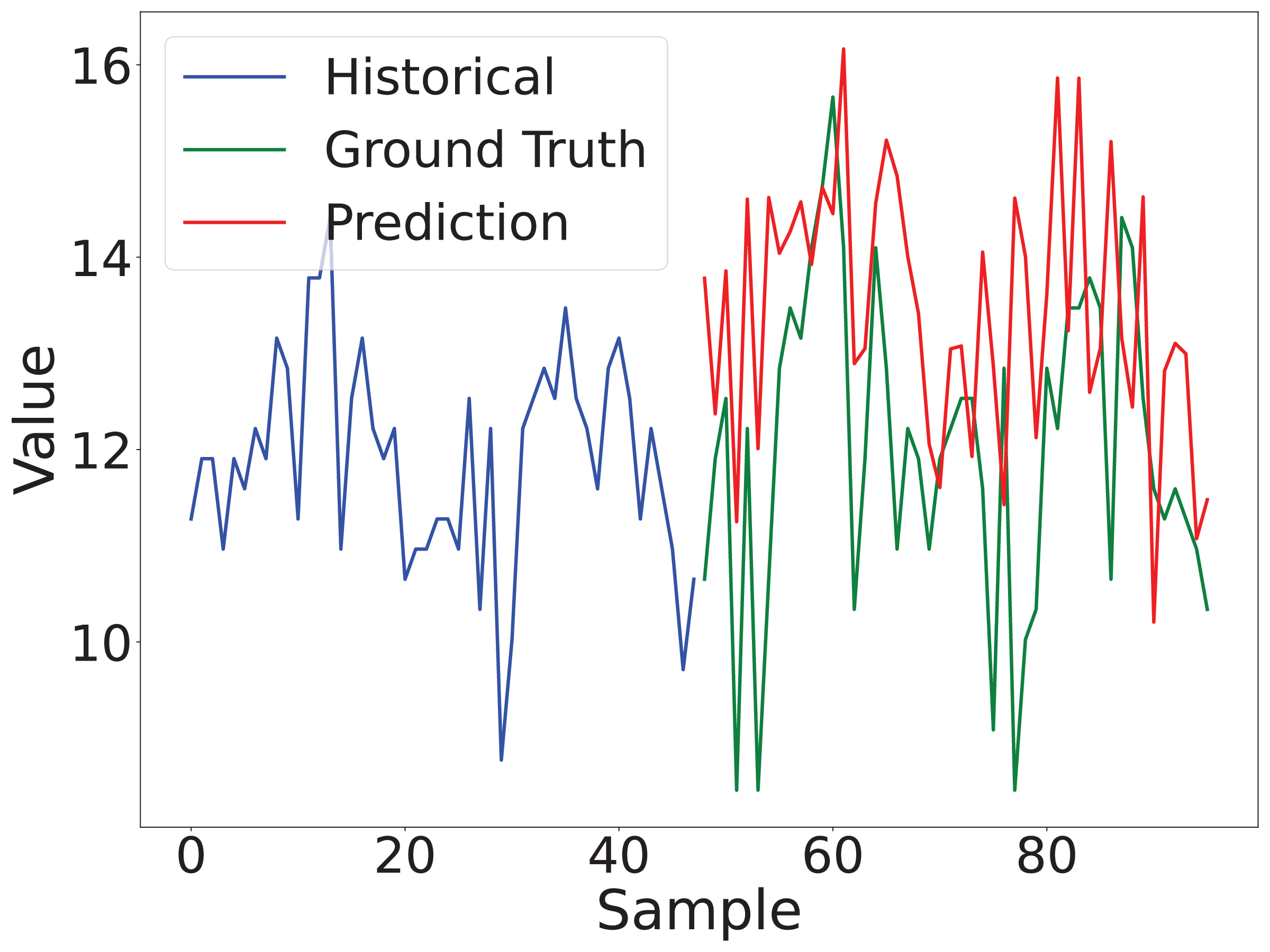}
        \caption{ETTh2-48}
        \label{fig:}
    \end{subfigure}
    \begin{subfigure}[b]{0.24\textwidth}
        \centering
        \includegraphics[width=\textwidth]{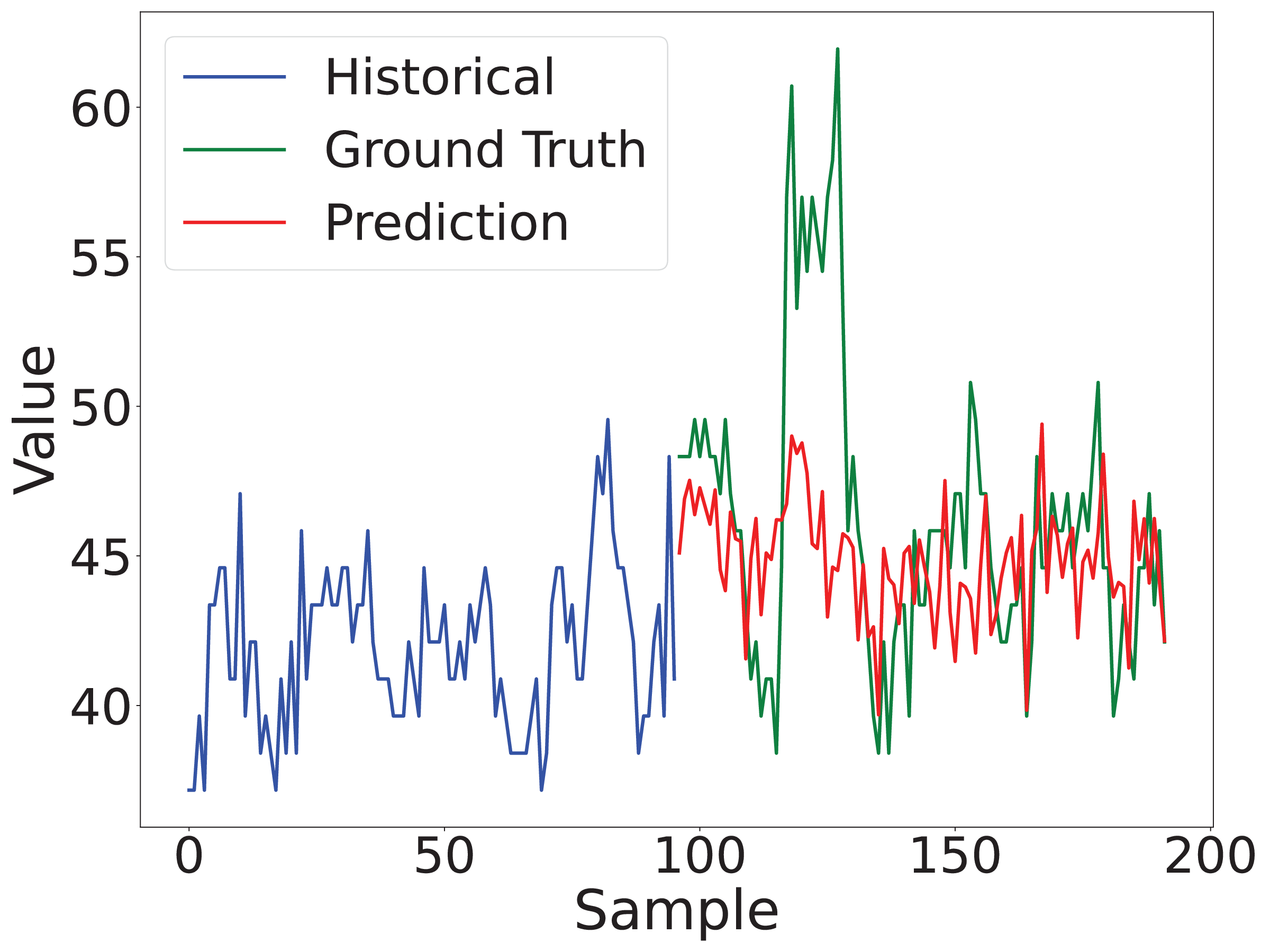}
        \caption{ETTh2-96}
        \label{fig:}
    \end{subfigure}
    \begin{subfigure}[b]{0.24\textwidth}
        \centering
        \includegraphics[width=\textwidth]{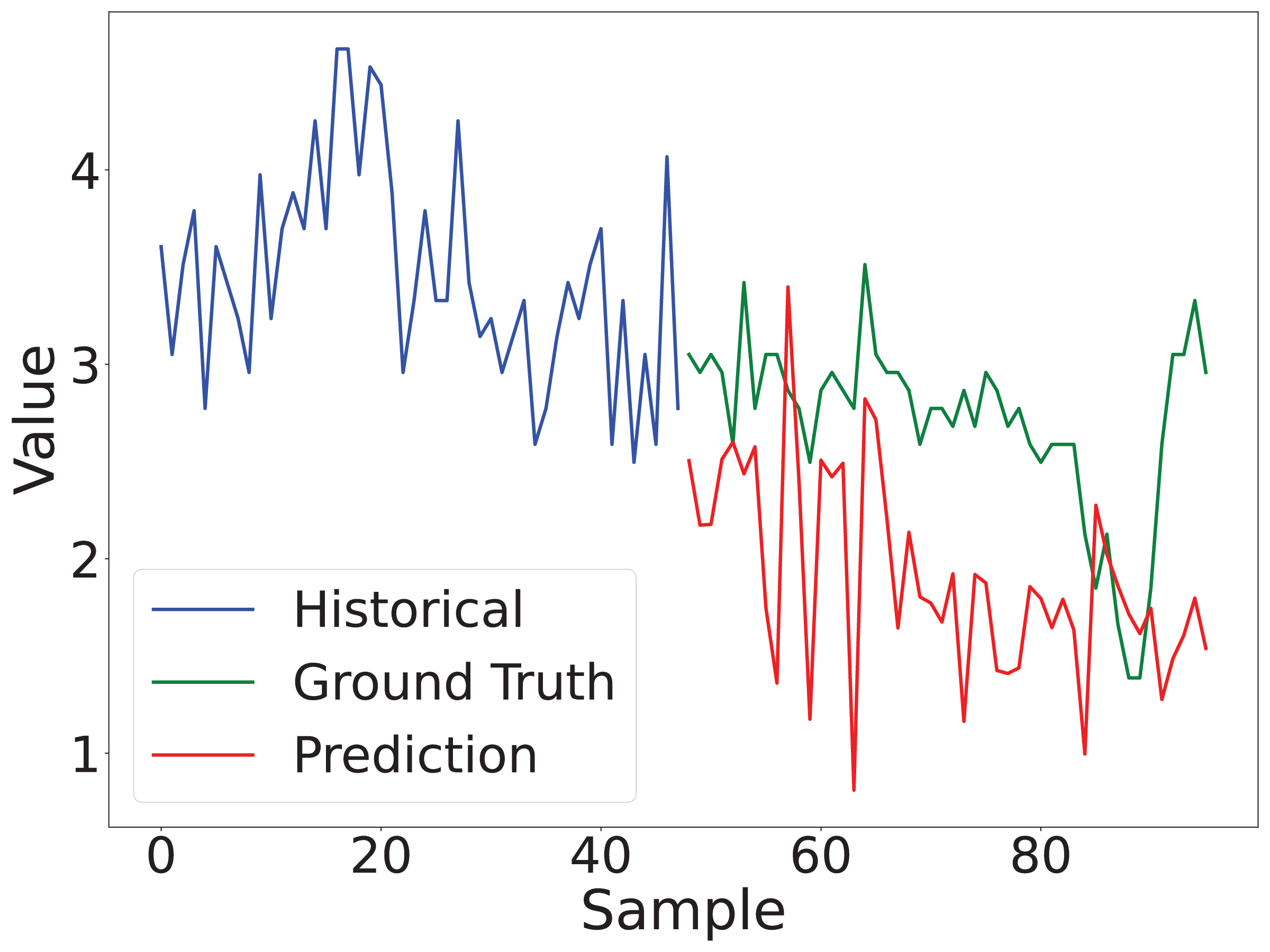}
        \caption{ETTm1-48}
        \label{fig:}
    \end{subfigure}
    \begin{subfigure}[b]{0.24\textwidth}
        \centering
        \includegraphics[width=\textwidth]{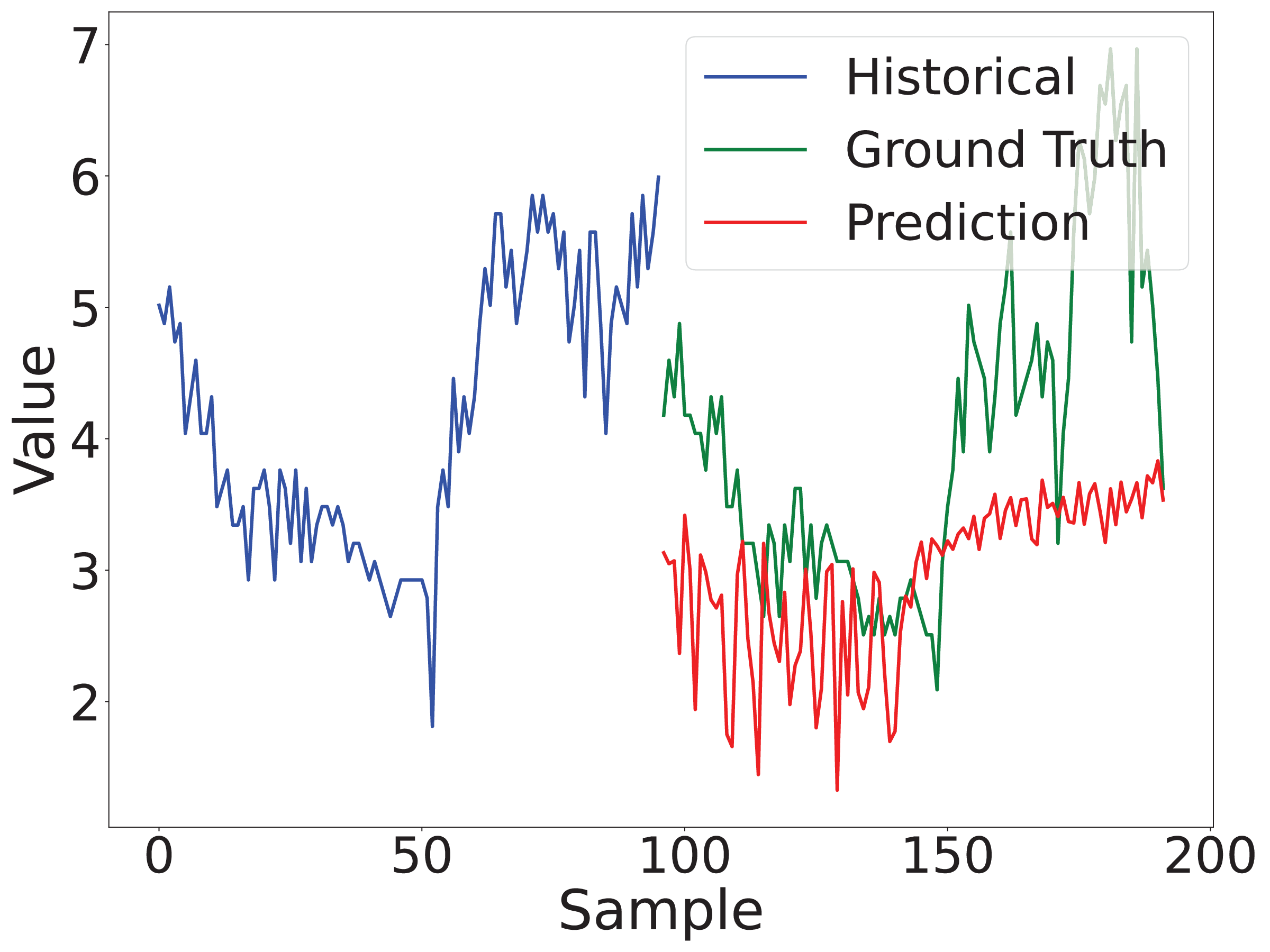}
        \caption{ETTm1-96}
        \label{fig:}
    \end{subfigure}
    \begin{subfigure}[b]{0.24\textwidth}
        \centering
        \includegraphics[width=\textwidth]{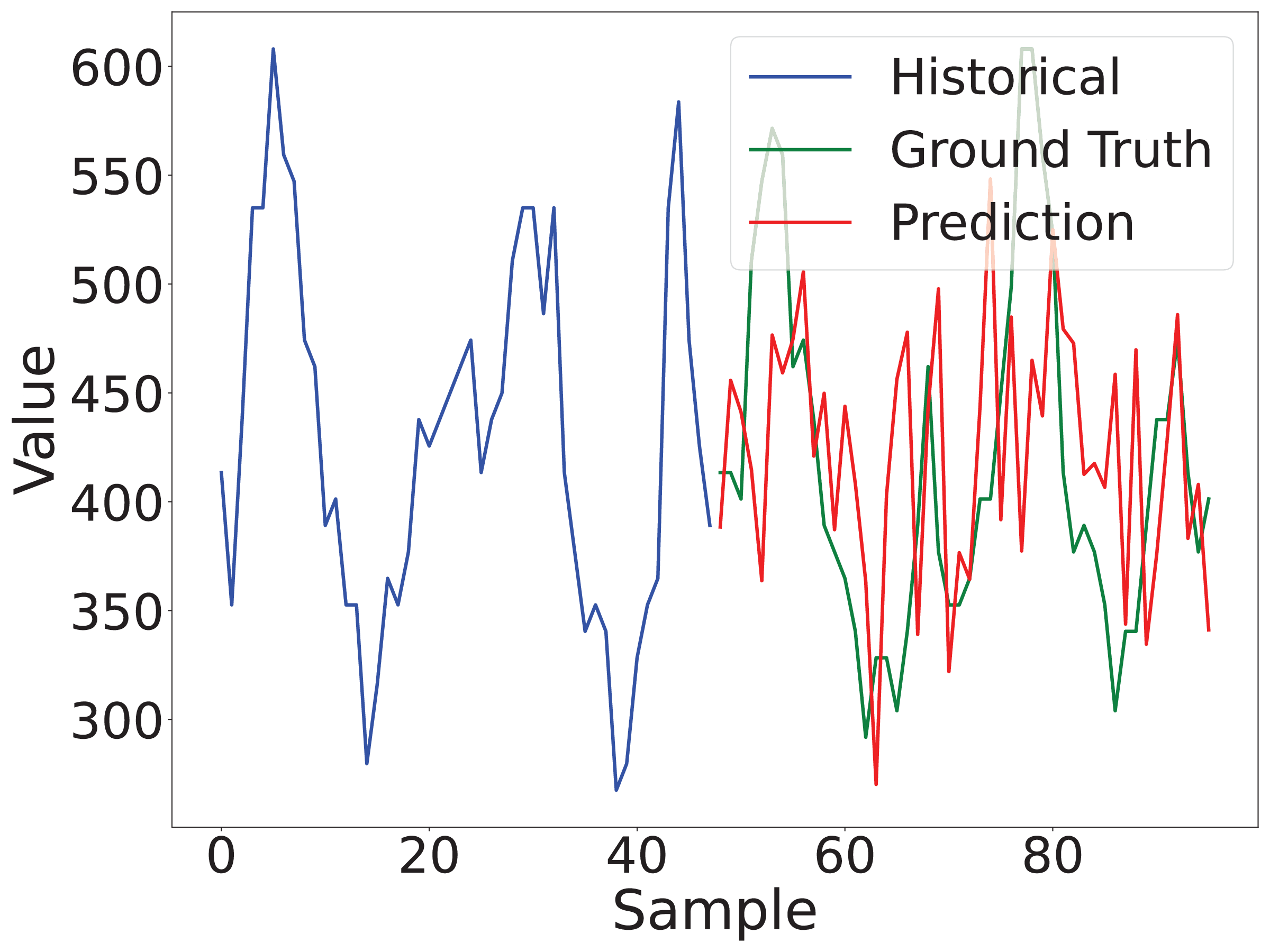}
        \caption{electricity-48}
        \label{fig:}
    \end{subfigure}
    \begin{subfigure}[b]{0.24\textwidth}
        \centering
        \includegraphics[width=\textwidth]{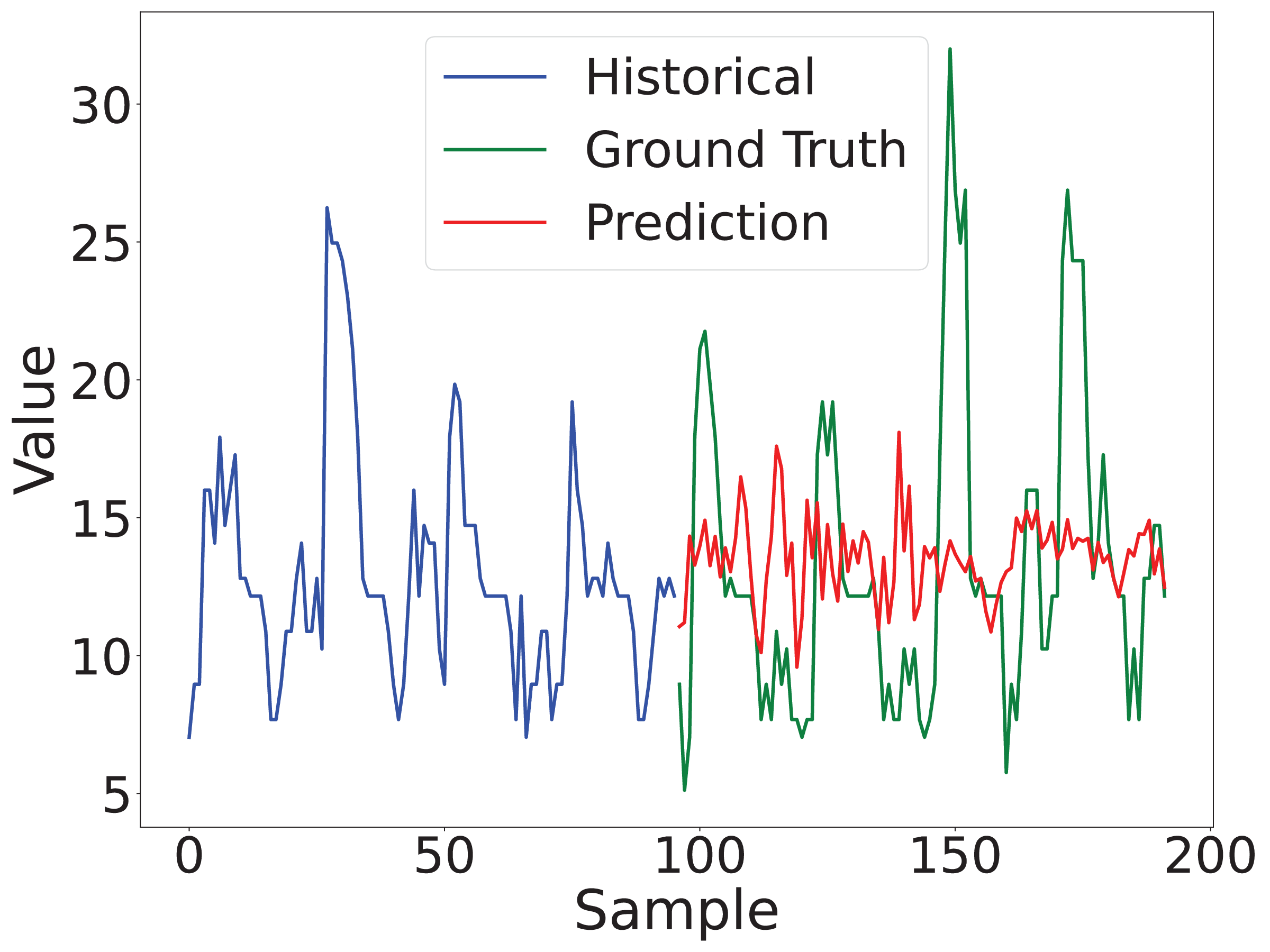}
        \caption{electricity-96}
        \label{fig:}
    \end{subfigure}
    \caption{Sample traces of \solution{} prediction using GPT-4o-mini in multivariate generation}
    \label{fig:multi-gpt-4o-mini}
\end{figure}

In Fig.~\ref{fig:uni-gpt-4o-mini}, Fig.~\ref{fig:uni-llama-7b} and Fig.~\ref{fig:multi-gpt-4o-mini} show the sample prediction traces for different datasets for both 48 and 96 prediction length scenarios. The predictions are from GPT-4o-mini (Fig.~\ref{fig:uni-gpt-4o-mini}, Fig.~\ref{fig:multi-gpt-4o-mini}) and Llama-7B (Fig.~\ref{fig:uni-llama-7b}) models. 

\subsubsection{Limitations in predictions to further improve}\label{append:subsubsec:limitations through samples}

\begin{figure}[h!]
    \centering
    \begin{subfigure}[b]{0.24\textwidth}
        \centering
        \includegraphics[width=\textwidth]{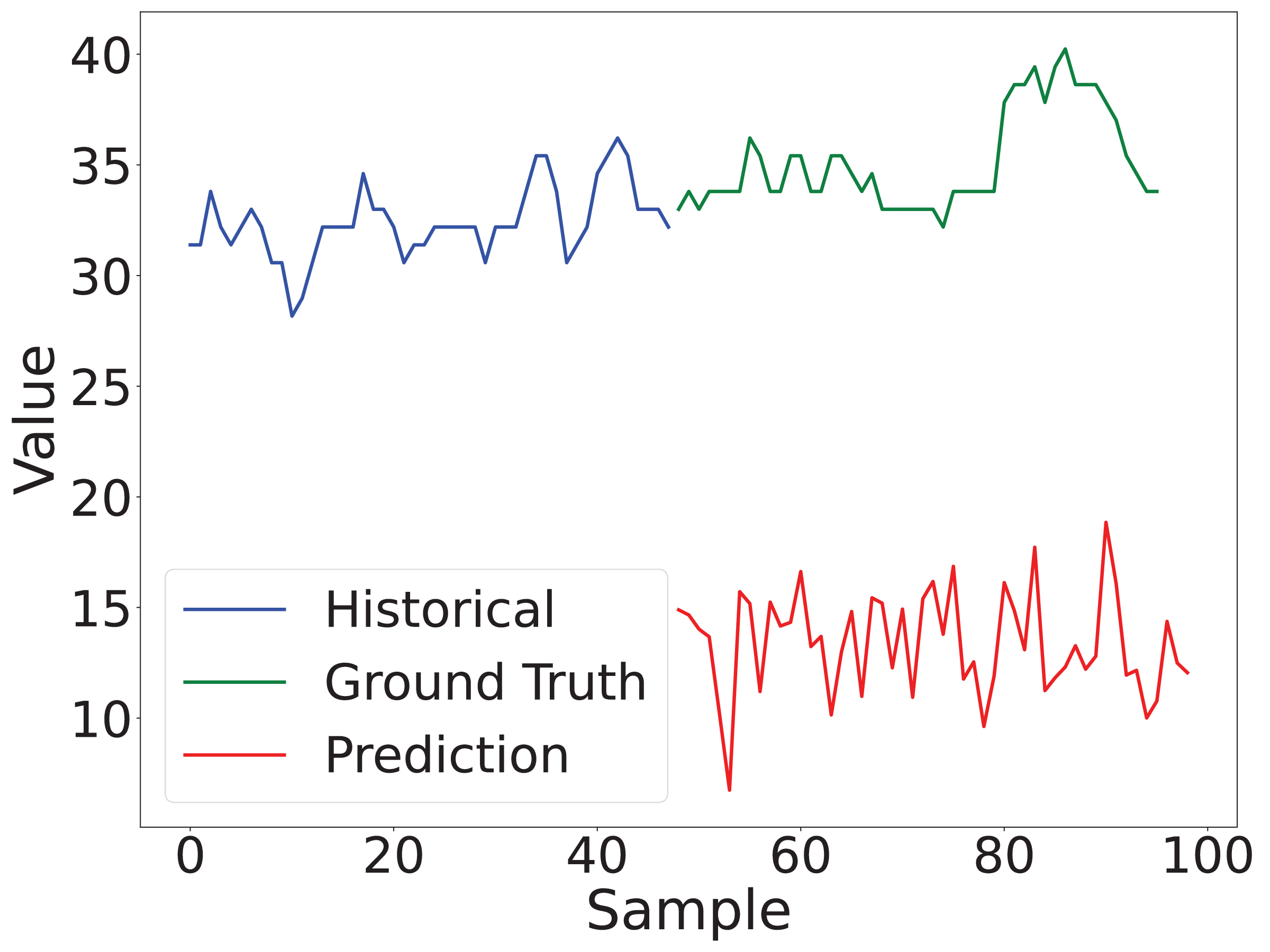}
        \caption{Higher offset 1}
        \label{fig:high offset}
    \end{subfigure}
    \begin{subfigure}[b]{0.24\textwidth}
        \centering
        \includegraphics[width=\textwidth]{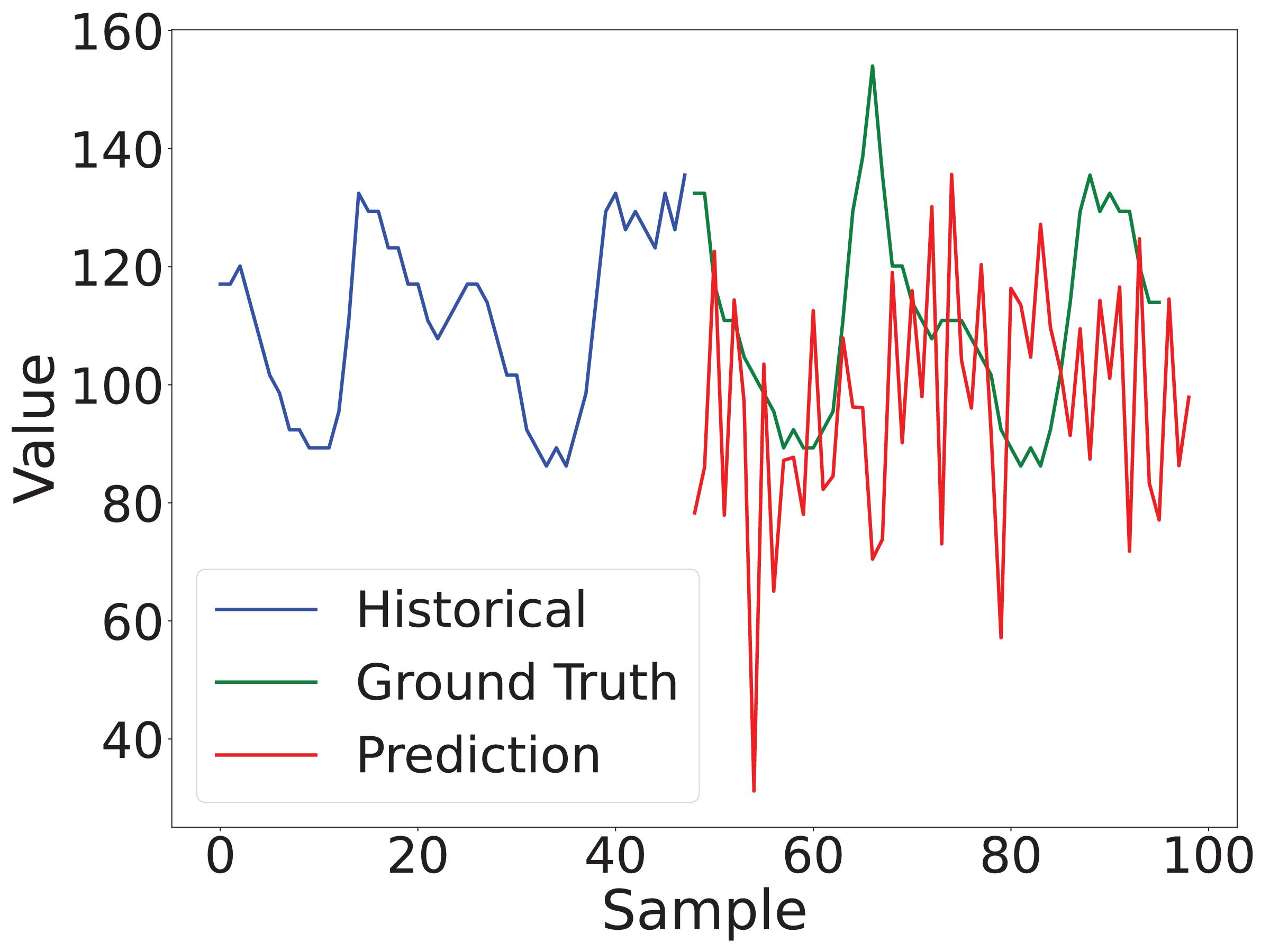}
        \caption{Excess noise from $X^c_h$}
        \label{fig:high-freq-impact}
    \end{subfigure}
    \begin{subfigure}[b]{0.24\textwidth}
        \centering
        \includegraphics[width=\textwidth]{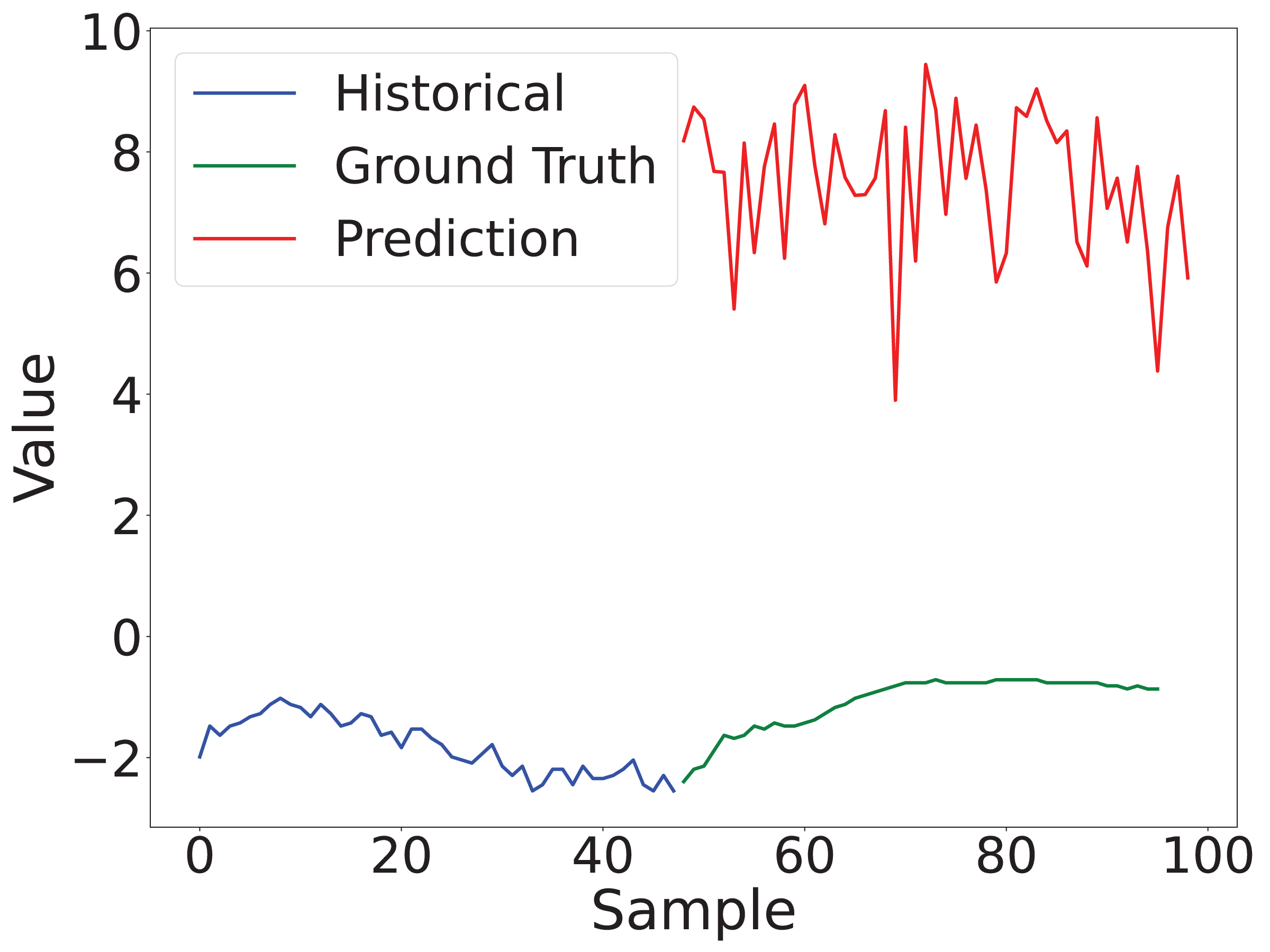}
        \caption{Combinatorial impact 1} 
        \label{fig:comb impact 1}
    \end{subfigure}
      \begin{subfigure}[b]{0.24\textwidth}
        \centering
        \includegraphics[width=\textwidth]{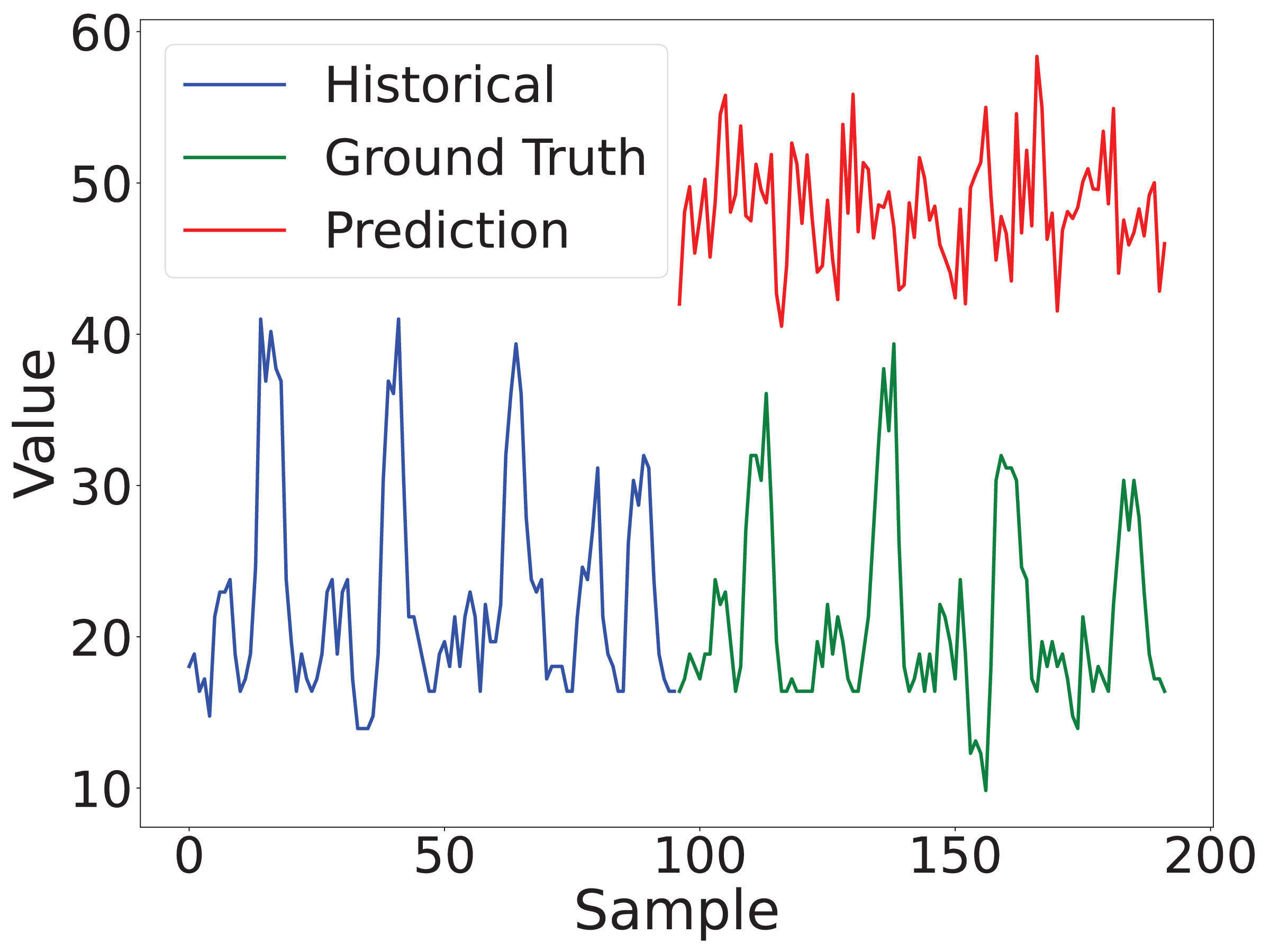}
        \caption{Combinatorial impact 2}
        \label{fig:comb impact 2}
    \end{subfigure}
    \caption{Sample traces of \solution{} predictions highlighting areas for potential improvement}
    \label{fig:limitations}
\end{figure}

 Figure~\ref{fig:limitations} presents sample traces highlighting areas where \solution{} could be further improved. In Fig.~\ref{fig:high offset}, we observe a vertical offset between the predicted and ground truth sequences. This inconsistency stems from the max-normalization used during data preprocessing. Small deviations in the normalized space can translate into significant differences in the original time-series scale after de-normalization. As future work, we plan to investigate alternative normalization techniques, such as standard normalization and min-max normalization, to potentially enhance \solution{}'s performance.
Another observed limitation (Fig.~\ref{fig:high-freq-impact}) involves the influence of $X^c_h$ (the high-frequency component), which introduces undesired fluctuations on top of $X^c_l$ (the low-frequency component). This effect is especially noticeable when the original signal is smooth, leading to an unnaturally noisy output. However, in cases where the input sequence naturally contains notable $X^c_h$ components—such as in Fig.~\ref{fig:good impact high freq 1} and Fig.~\ref{fig:good impact high freq 2}—\solution{} successfully preserves the high-frequency patterns with high fidelity. To address this, we plan to extend \solution{} with an adaptive mechanism that selectively includes the $X^c_h$ component based on the smoothness of the historical sequence, assessed through power spectral analysis.
Finally, Fig.~\ref{fig:comb impact 1} and Fig.~\ref{fig:comb impact 2} illustrate the combined effects of the aforementioned limitations.

\section{Societal impacts}\label{append-sec:societal impact}

The development of \solution{} offers several positive societal impacts, particularly in time-series forecasting applications such as finance, weather prediction, and traffic management. Accurate forecasting in these domains enables better decision-making that can enhance quality of life in both the short and long term. For example, using \solution{} for weather forecasting can support effective agricultural planning, while its application in long-term financial forecasting can assist governments in making informed decisions to strengthen the economy. However, \solution{} may also raise certain ethical and societal concerns. Since it relies on real-world data as input, there are potential privacy issues—especially when used in sensitive sectors like healthcare. Furthermore, as \solution{} is built on Large Language Models (LLMs), it inherits common challenges associated with LLMs, such as the risk of misinformation, bias across datasets, and over-reliance on automated decisions. To mitigate these risks, it is important to use LLMs that undergo proper auditing and debiasing, incorporate post-verification mechanisms, and maintain human oversight in critical decision-making processes.

\end{document}